\documentclass{article} %
\usepackage{iclr2027_conference,times}

\usepackage{url}

\usepackage{amsmath,amsthm,amssymb}
\usepackage{graphicx,booktabs,array,enumitem,float}
\usepackage{pgfplots,placeins}
\usepackage{xurl} %
\usepackage{hyperref} %
\pgfplotsset{compat=1.18}

\floatstyle{ruled}\newfloat{algorithm}{tbp}{loa}\floatname{algorithm}{Algorithm}
\floatstyle{plain}\newfloat{algorithmgroup}{tbp}{loa}\floatstyle{ruled}
\newtheorem{theorem}{Theorem}
\newtheorem{lemma}{Lemma}
\newtheorem{corollary}{Corollary}
\newtheorem{proposition}{Proposition}
\newtheorem{assumption}{Assumption}
\theoremstyle{definition}\newtheorem{definition}{Definition}
\theoremstyle{remark}\newtheorem{remark}{Remark}
\newcommand{\E}{\mathbb E}\newcommand{\Pp}{\mathbb P}
\newcommand{\one}{\mathbf 1}\newcommand{\cC}{\mathcal C}
\newcommand{\kl}{\operatorname{kl}}\newcommand{\KL}{\operatorname{KL}}

\DeclareMathOperator{\Var}{Var}
\newcommand{\Reg}{\mathcal R}\newcommand{\history}{\mathcal H}
\newcommand{\rhist}{\mathcal G}\newcommand{\harm}{\mathsf h}
\newcolumntype{L}[1]{>{\raggedright\arraybackslash}p{#1}}

\title{The Price of Hard Coverage Constraints in Online Conformal Prediction}

\author{Mingrui Liu\\Department of Computer Science\\George Mason University\\\texttt{mingruil@gmu.edu}}

\hypersetup{
 pdftitle={The Price of Hard Coverage Constraints in Online Conformal Prediction},pdfauthor={Mingrui Liu},
 pdfsubject={Frontier geometry, learned calibration, and efficiency throughout deployment}}

\iclrfinalcopy
\begin{document}
\maketitle

\vspace*{-0.2in}
\begin{abstract}

Conformal prediction provides distribution-free uncertainty quantification, but in sequential deployment, standard coverage guarantees can still allow substantial fluctuations in the number of miscoverages over time.
We study an online setting in which prediction sets are produced one round at a time and the cumulative number of miscoverages must stay within a prescribed budget throughout deployment.
Such a hard coverage guarantee can be enforced by enlarging prediction sets, but how much additional prediction-set size is fundamentally necessary remains unclear.
We characterize this unavoidable efficiency cost over $T$ prediction rounds.
Remarkably, the cost persists even with i.i.d.\ data, a known score distribution, and perfect knowledge of the population-optimal prediction set.
Under suitable regularity conditions relating coverage and prediction-set size, we determine the minimum cumulative excess prediction-set size required by any online method satisfying the hard coverage constraint.
This minimum cost is $\Theta(\log T)$ for expected cost at a fixed time known in advance, $\Theta(T^{1/3})$ for the largest expected cost over all fixed times up to $T$, and $\Theta(\sqrt{T})$ for the expected worst realized cost when the evaluation time is chosen in hindsight.
We further show that these rates can be attained using past scores, without knowing the score distribution.
Thus, even perfect statistical knowledge does not make sequential reliability free.

\end{abstract}

\section{Introduction}\label{sec:intro}

Conformal prediction provides prediction sets with rigorous coverage guarantees while making minimal assumptions on the underlying predictive model \citep{shafer-vovk,angelopoulos-bates}. Its online variants have been motivated by sequential prediction problems where uncertainty estimates are repeatedly used over time, including epidemic forecasting and electricity-demand prediction \citep{gibbs21,conformal-pid,areces}. In such settings, reliability during deployment can matter beyond guarantees that hold only marginally or over a long horizon. Similar concerns arise more broadly in sequential decision making, where conservative online-learning methods require cumulative performance to remain above a prescribed baseline throughout operation \citep{wu2016conservative}. Consider an online deployment over rounds $t=1,2,\ldots$ with target miscoverage level $\alpha\in(0,1)$. At round $t$, the method selects a prediction-set rule $C_t$ using past observations, then observes the feature $X_t$ and outputs $C_t(X_t)$ before observing the outcome $Y_t$. Even if each prediction has miscoverage probability exactly $\alpha$, random fluctuations can cause the miscoverage count to temporarily exceed its nominal level $\alpha t$. A natural response is to enlarge prediction sets when the miscoverage quota becomes tight, reducing the chance of another miscoverage at the cost of less informative predictions.

Document retrieval makes this tradeoff concrete. For each incoming query, a fixed retriever returns a shortlist intended to contain a target document in the collection \citep{intrator2024retrieval}. A miscoverage occurs when that document is omitted. Larger shortlists reduce omissions but require more downstream processing. Returning the entire collection guarantees coverage, but eliminates the efficiency benefit of retrieval. Suppose the target document is revealed after each query and the cumulative fraction of omissions must remain at or below a prescribed limit throughout deployment. This \emph{hard coverage constraint} must hold after every query, so good performance later cannot undo an earlier violation. Even a retriever with the correct miscoverage probability can violate it through random fluctuations. Preventing such violations can therefore require larger shortlists. The minimum cumulative excess prediction-set size quantifies how much retrieval efficiency must be sacrificed to enforce the constraint. Limited correction resources can make such a constraint necessary. \mbox{In inspection,} longer lists of possible faults require more tool preparation. Inspection reveals the true fault before the next prediction, and omissions need immediate correction. Suppose the correction budget starts at zero and gains one unit before processing parts $10,20,30,\ldots$. Unused budget carries over, but future budget cannot be used early. Then at most $\lfloor t/10\rfloor$ omissions are affordable after $t$ parts. A final-time or long-run guarantee could allow an omission before resources are available to correct it. A high-probability guarantee still permits violations on some runs (Appendix~\ref{app:ext-workcell}).\ This motivates our central question: \emph{how~much additional prediction-set size is fundamentally necessary to enforce hard coverage constraints throughout online deployment?}

A growing literature studies how online conformal methods adapt prediction sets using past miscoverages, scores, or distributional changes \citep{gibbs21,gibbs24,bhatnagar,conformal-pid,betting,gradient-equilibrium,areces,wu2025error}. Recent work has also made prediction-set efficiency an explicit objective \citep{yang2024bellman,srinivas,vaze}. These methods provide sophisticated ways to learn, adapt, or control online coverage, but they do not characterize the minimum efficiency cost imposed solely by a constraint on cumulative miscoverage. To isolate this intrinsic cost, we first analyze an oracle stationary setting with i.i.d.\ data, a fixed nonconformity score, and complete knowledge of the data distribution. Let $M_t=\sum_{u=1}^t\mathbf{1}\{Y_u\notin C_u(X_u)\}$ denote the cumulative number of miscoverages by round $t$, and require $M_t\leq\lfloor\alpha t\rfloor$ for every $t$. This hard constraint represents the zero-tolerance endpoint of a broader family of constraints permitting additional miscoverages. In the oracle analysis, the population coverage--efficiency frontier and its optimal prediction sets are known exactly, so estimation and distribution tracking are absent. Nevertheless, enforcing the realized coverage constraint still incurs a nonzero efficiency cost. For a horizon $T$, let $j(\alpha)$ denote the minimum expected prediction-set size at miscoverage $\alpha$ within a fixed family, allowing randomization. With $\mu$ measuring set size, define the constraint price as $\mathcal R_T=\E[\sum_{t=1}^T\mu(C_t(X_t))]-Tj(\alpha)$, the cumulative excess prediction-set size relative to the unconstrained population oracle.

Characterizing this minimum price raises three main challenges. First, an oracle with the correct miscoverage probability on every round can still exceed the miscoverage quota by chance. We must quantify the additional prediction-set size required to convert population-level coverage into pathwise control. Second, the cost depends critically on when efficiency is evaluated. A known evaluation horizon allows planning the remaining miscoverage quota. Controlling expected efficiency at every prefix requires preserving slack for future miscoverages, changing the optimal rate even when the final horizon is known. Our construction also works without knowing the horizon. Third, lowering the instantaneous miscoverage probability generally requires larger sets, with the enlargement governed by the local geometry of the population coverage--efficiency frontier. Thus, miscoverage control must be related to prediction-set size. Random fluctuations in realized cost create an additional difficulty when performance is evaluated retrospectively.

We connect control of cumulative miscoverage with the population coverage--efficiency frontier. An exact decomposition separates the cost associated with miscoverage slack, deviations from the frontier's supporting line, and size above the frontier. Conditioning independent Bernoulli$(\alpha)$ indicators on feasible paths yields an order-optimal controller at a fixed evaluation horizon. Barrier-based policies preserve slack while limiting conservatism without knowing the horizon. Martingale arguments characterize fluctuations at the worst realized prefix. Under the stated frontier conditions, these ideas yield matching upper and lower bounds for three temporal regimes. Randomized empirical ranks attain the corresponding optimal orders without knowing the score distribution under the stated regularity conditions. This separates the statistical cost of threshold estimation from the intrinsic cost of enforcing coverage. Our main contributions are summarized as follows.
\begin{itemize}

\item We formulate the \emph{price of hard cumulative coverage} as the minimum cumulative excess prediction-set size, relative to the unconstrained population-optimal predictor, required solely to keep realized miscoverages within a running budget.

\item We sharply characterize this price through matching upper and lower bounds, revealing three distinct temporal regimes: $\Theta(\log T)$ when performance is evaluated at a fixed time known in advance, $\Theta(T^{1/3})$ when efficiency is controlled at every fixed prefix in expectation, and $\Theta(\sqrt T)$ when performance is evaluated at the worst realized prefix in hindsight. We establish these results using feasible-path conditioning, barrier control, and martingale arguments, and show that empirical rank-based procedures attain these orders under the same geometric conditions without knowing the score distribution.
All three criteria impose the same quota at every round. Table~\ref{tab:trichotomy} lists their rates and controllers, and Section~\ref{sec:ext-implications} discusses implications for existing online conformal methods.

\item Experiments support the theoretical analysis of evaluation time and threshold estimation. Numerical prefix-optimal benchmarks and guarded baselines assess finite-horizon performance and horizon-specific tuning. Held-out regression experiments estimate learned-score curvature and empirical frontier gaps, with uncertainty estimates. Paired online experiments compare prediction-set size and compliance with the hard coverage constraint.

\end{itemize}

\begin{table}[!htbp]
\centering\small
\setlength{\tabcolsep}{4pt}
\caption{Optimal prices under the assumptions of Theorem~\ref{thm:optimal-prices}
(adaptive stopping: Appendix~\ref{app:stopping-optimal}). Here $R_t$ is realized
cumulative excess size above $tj(\alpha)$, and $\Reg_t=\E_D R_t$.
A stopping time uses only information available so far. All constructions use
Algorithm~\ref{alg:rank} and enforce the same quota after every prediction.}
\label{tab:trichotomy}\label{tab:algorithmic-upper}
\begin{tabular}{@{}L{0.45\linewidth}L{0.30\linewidth}l@{}}
\toprule
Criterion minimized over feasible policies & Construction & Optimal price\\\midrule
(i) Terminal: $\Reg_T$ & Counting (Algorithm~\ref{alg:count}) & $\Theta(\log T)$\\
(ii) Expected prefixes: $\max_{t\le T}\Reg_t$ & Barrier (Algorithm~\ref{alg:sine}) & $\Theta(T^{1/3})$\\
(ii$'$) Adaptive stopping: $\sup_\tau\E_D R_\tau$ & Barrier (Algorithm~\ref{alg:sine}) & $\Theta(T^{1/3})$\\
(iii) Retrospective: $\E\max_{t\le T}R_t$ & Barrier (Algorithm~\ref{alg:sine}) & $\Theta(\sqrt T)$\\
\bottomrule
\end{tabular}
\end{table}

\section{Related Work}\label{sec:related}

\textbf{Online conformal prediction and adaptation.}
A broad line of work studies conformal prediction in sequential and nonstationary environments, including adaptive quantile updates, strongly adaptive guarantees, control-based calibration, betting methods, decaying-step procedures, online optimization, and online-learning interpretations \citep{gibbs21,zaffran,gibbs24,bhatnagar,conformal-pid,betting,angelopoulos2024decaying,gradient-equilibrium,areces,universal-portfolio,wu2025error,feldman}. These methods develop mechanisms for adapting prediction sets to past miscoverages, scores, or distributional changes. Our oracle analysis deliberately removes this source of difficulty and instead isolates the efficiency cost induced solely by constraining the realized coverage path.

\textbf{Coverage and prediction-set efficiency.}
Prediction-set size is a standard measure of conformal efficiency \citep{lei2014bands,sadinle2019least,dhillon2024expected}, and several recent works study coverage and efficiency jointly in online settings through stochastic control, competitive analysis, dynamic regret, quantile tracking, and limited-feedback selection \citep{yang2024bellman,srinivas,vaze,gollapudi}. This literature establishes that coverage alone is insufficient and that the informativeness of the prediction set must also be accounted for. Our focus is different: we characterize the minimum efficiency loss that \emph{any} policy must incur solely to enforce a hard miscoverage quota, even when the population-optimal prediction rule is already known.

\textbf{Risk control, local validity, and time-uniform guarantees.}
Conformal prediction has been extended beyond marginal miscoverage to general risk control, nonexchangeable settings, multivalid and strongly adaptive guarantees, and validity that holds uniformly over time or under adaptive stopping \citep{bates2021risk,angelopoulos2024crc,barber2023beyond,mvp,bhatnagar,scharfstein2026timeuniform,hultberg2026anytime,dubey2026anytime}. Related statistical tools include time-uniform confidence sequences and sequential quantile estimation \citep{howard2021confidence,howard2022quantiles}, while strongly adaptive and anytime notions are also central in online learning \citep{daniely2015strongly,harvey2023anytime}. These works strengthen the form or temporal scope of validity. In contrast, we impose a pathwise miscoverage quota and study the minimum prediction-set-size price required to enforce it.

\textbf{Online learning under running constraints.}
Our work is related more broadly to safe and conservative sequential learning under performance or resource constraints \citep{mannor2009samplepath,wu2016conservative,kazerouni2017conservative,garcelon2020conservative,du2021conservative,bernasconi2022safe,mcmahan2024anytime,sridharan2025unknown}. This literature studies sample-path constraints, conservative exploration, anytime feasibility, and learning under unknown constraints. We share this fundamental-limits perspective, but our constraint acts on realized miscoverage and its cost is measured by excess prediction-set size. The frontier couples these quantities even without a learning problem. Conservative bandits protect baseline reward, while anytime-constrained MDPs and replenishment models enforce running resource budgets \citep{chen2026replenishment}. Classical queueing control gives an inverse-square excess-cost--buffer relation \citep{berry-gallager}. Our finite-horizon comparison separates the population size comparator, the information available when a map is chosen, and the three temporal criteria (Appendix~\ref{app:control-comparison}). The disjoint-window lower bound covers arbitrary history-dependent policies, and empirical ranks implement the matching constructions.

\par
\section{Notation and Problem Setup}\label{sec:setup}

\textbf{Prediction-set family and online protocol.}
Let $\mathcal X$ and $\mathcal Y$ denote the feature and outcome spaces, and let $\cC$ be the collection of prediction-set rules available to the calibrator. Each $C\in\cC$ is a measurable map $C:\mathcal X\to 2^{\mathcal Y}$, so that $C(x)$ is the prediction set issued at feature $x$. A common special case is a fixed nonconformity score $s$, with threshold sets $C_\tau(x)=\{y:s(x,y)\le\tau\}$ \citep{angelopoulos-bates}. Let $\mu$ measure set size and define $\ell_C(x):=\mu(C(x))$. We assume $B:=\mu(\mathcal Y)<\infty$ and $\ell_C(x)\in[0,B]$. The family $\cC$ contains the full-set and empty-set maps.

\begin{assumption}[Online efficiency protocol]\label{ass:protocol}
The pairs $(X_t,Y_t)$ are i.i.d.\ from a distribution $D$ on $\mathcal X\times\mathcal Y$. The family $\cC$ is fixed independently of these observations, possibly using independent training data. At round $t$, the calibrator selects $C_t\in\cC$ from past observations and fresh independent randomization, then observes $X_t$, issues $C_t(X_t)$, and finally observes $Y_t$.
\end{assumption}

We condition on any independent training used to construct $\cC$. Covariate-dependent thresholding is included by treating the entire rule $x\mapsto C(x)$ as a map selected before the fresh covariate is observed.

\textbf{Hard cumulative coverage.}
Fix a target miscoverage level $\alpha\in(0,1)$. Let $E_t=\one\{Y_t\notin C_t(X_t)\}$ be the miscoverage indicator, $M_t=\sum_{u\le t}E_u$ the cumulative number of miscoverages, and $W_t=\sum_{u\le t}\ell_{C_u}(X_u)$ the cumulative prediction-set size.

\begin{definition}[Hard cumulative coverage]\label{def:valid}
A calibrator is feasible if
$M_t\le b_t:=\lfloor\alpha t\rfloor$
at every round $t$ in its specified horizon, for every data stream and every realization of its internal randomization.
\end{definition}

This pathwise requirement is independent of the i.i.d.\ assumption used to evaluate efficiency. It controls cumulative miscoverage at every prefix, but does not imply $M_t/t\to\alpha$, conditional coverage, or rolling-window coverage. At $\alpha=0.1$, the first nine outputs must be full on every stream. Finite $B$ is therefore essential for finite cost.

\textbf{Population coverage--efficiency frontier.}
For a fixed rule $C\in\cC$, let $m_D(C):=\Pr_{(X,Y)\sim D}\{Y\notin C(X)\}$ denote its population miscoverage probability and $V_D(C):=\E_{(X,Y)\sim D}[\mu(C(X))]$ its expected prediction-set size. To allow randomization, let $\Delta(\cC)$ denote the set of probability distributions over $\cC$. For $\pi\in\Delta(\cC)$, define $m_D(\pi):=\E_{C\sim\pi}[m_D(C)]$ and $V_D(\pi):=\E_{C\sim\pi}[V_D(C)]$.

The population coverage--efficiency frontier is
$j(q)=\inf_{\pi\in\Delta(\cC):\,m_D(\pi)=q}V_D(\pi)$, $q\in[0,1]$.
Thus, $j(q)$ is the minimum expected size within $\cC$ at miscoverage $q$, allowing randomization. Taking $\cC$ to be a fixed score's thresholds gives its score-restricted frontier.

\textbf{Constraint price.}
Let $\mathcal A$ denote an online calibrator following the protocol above. Over a horizon of $T$ rounds, define
$\mathcal R_T(\mathcal A,D):=\E_D W_T-Tj(\alpha)$ and
$\mathcal R_T^\star(D):=\inf_{\mathcal A\ \mathrm{feasible}}\mathcal R_T(\mathcal A,D)$.
Here, $\mathcal R_T(\mathcal A,D)$ is the expected cumulative excess prediction-set size of calibrator $\mathcal A$ relative to the unconstrained population benchmark $Tj(\alpha)$, while $\mathcal R_T^\star(D)$ is the smallest such excess achievable by any feasible online calibrator. Hence $\mathcal R_T^\star(D)$ is the fundamental price of the hard coverage constraint. Policies in the infimum may depend on $D$ and $T$, but not on unseen outcomes. Allowing randomized rules makes $j$ convex and nonincreasing. See Appendix~\ref{app:conformal}.

\textbf{Coverage balance.}
Define the miscoverage slack $G_t=\alpha t-M_t$. Since the balance increment $g_t=\alpha-E_t$ (the pinball-loss subgradient at finite thresholds) satisfies $\sum_{t\le T}g_t=G_T$, gradient equilibrium $G_T/T\to0$ is equivalent to asymptotic coverage balance $M_T/T\to\alpha$~\citep{gradient-equilibrium}. It does not characterize the prediction-set-size cost of enforcing the pathwise constraint.

\vspace{-0.15in}
\section{Algorithms and Theoretical Guarantees}\label{sec:algorithms-theory}

We abbreviate $\Reg_T(\mathcal A,D)$ as $\Reg_T$.
Counting controls terminal cost, while Barrier controls expected-prefix cost.
Empirical ranks implement both without knowing the score distribution.

\subsection{Excess set-size decomposition}\label{sec:framework}
\vspace*{-0.05in}

Let $\varpi_t$ be the independent random seed at round $t$ and let
$\history_t=\sigma(X_u,Y_u,\varpi_u:1\le u\le t)$ denote the observed history.
The conditional miscoverage probability is
$p_t:=\Pp(E_t=1\mid\history_{t-1})=\E[m_D(C_t)\mid\history_{t-1}]$.
This averages over the current seed and fresh observation. Since the
conditional law of $C_t$ is a mixture allowed in the definition of $j$,
$\epsilon_t:=\E[V_D(C_t)\mid\history_{t-1}]-j(p_t)\ge0$ measures how
much its expected prediction-set size exceeds the frontier value $j(p_t)$.
Even if each selected rule attains the frontier at its own miscoverage,
randomizing between different miscoverage levels can give $\epsilon_t>0$.

Choose $\lambda\ge0$ such that
$j(q)\ge j(\alpha)-\lambda(q-\alpha)$ for all $q\in[0,1]$.
Such a $\lambda$ exists because $j$ is convex and nonincreasing.
If $j$ is differentiable at $\alpha$, then $\lambda=-j'(\alpha)$.
Define $\psi(q):=j(q)-j(\alpha)+\lambda(q-\alpha)$ for $q\in[0,1]$.
Thus $\psi(q)\ge0$ is the frontier's gap above this supporting line, and
$\psi(\alpha)=0$.

\begin{theorem}[Excess set-size decomposition]\label{thm:decomposition}\label{prop:decomposition}
Under Assumption~\ref{ass:protocol}, for every policy $\mathcal A$ and
integer $T\ge1$, $\Reg_T(\mathcal A,D)
 =\lambda\E G_T+\E\sum_{t=1}^T\psi(p_t)
                  +\E\sum_{t=1}^T\epsilon_t$.
\end{theorem}
\begin{remark}
The first term measures the cost of expected miscoverage slack,
$\E G_T=\alpha T-\E M_T$. The second measures the frontier's gap above
the supporting line at $p_t$. The third measures size above the frontier.
All three are nonnegative for a feasible policy, so small slack alone
does not ensure a small constraint price.
The full proof is in Appendix~\ref{app:decomposition}.
\end{remark}
\textbf{Proof sketch.}
Substitute $j(p_t)=j(\alpha)-\lambda(p_t-\alpha)+\psi(p_t)$ into
conditional expected size $j(p_t)+\epsilon_t$
(Lemma~\ref{lem:one-round-identity}). Sum over rounds and use
$\E\sum_{t\le T}p_t=\E M_T$ to obtain $\lambda\E G_T$.

\textbf{Example.}
For $Y=Q(U)$ with $U\sim\mathrm{Unif}[0,1]$ and $Q(u)=(3u+u^2)/4$,
intervals $[0,\tau]$ attain the length frontier among all measurable sets:
$j(q)=1-5q/4+q^2/4$ (Appendix~\ref{app:contact}). Therefore
$\lambda=5/4-\alpha/2$ and $\psi(q)=(q-\alpha)^2/4$:
the second term in Theorem~\ref{thm:decomposition} measures squared deviations
of conditional miscoverage from $\alpha$.

\subsection{Calibration algorithms}\label{sec:controllers}
\vspace*{-0.05in}

Fix a nonconformity score $s$ with $C_\tau\in\cC$ for every threshold, and let
$F(u)=\Pp_D\{s(X,Y)\le u\}$. Algorithms~\ref{alg:count} and~\ref{alg:sine}
request a miscoverage probability $q_t\in[0,1]$ using past miscoverage indicators.
For known continuous $F$ and $0<q_t<1$, use the threshold
$F^{-1}(1-q_t)$, where $F^{-1}(u)=\inf\{z:F(z)\ge u\}$.
Requests $q_t=0,1$ use the full-set and empty-set maps, respectively.
When $F$ is unknown, Algorithm~\ref{alg:rank} selects thresholds from past
scores (Section~\ref{sec:learning}). Smaller $q_t$ favors larger sets.

\textbf{How Algorithm~\ref{alg:rank} meets a request.}
With continuous i.i.d.\ scores, the new score's rank is uniform on
$\{1,\ldots,t\}$ given past ranks and rounding randomness, averaging over
past score values. The $J$-th past order statistic covers $J$ ranks.
The empty and full sets cover zero and all $t$ ranks. Randomly rounding
$t(1-q_t)$ to adjacent integers gives miscoverage probability $q_t$ without knowing
$F$ (Theorem~\ref{thm:rank}). Assumption~\ref{ass:contact} controls the
resulting set size.

\begin{algorithmgroup}[t]
\small
\textbf{Shared protocol.} Choose Counting for $t=1,\ldots,T$ or Barrier for $t=1,2,\ldots$.
Start with $M_0=G_0=0$ and no past scores. Write $b_t=\lfloor\alpha t\rfloor$,
$S_t=s(X_t,Y_t)$, and $C_\tau(x)=\{y:s(x,y)\le\tau\}$.
Each round, pass the controller's $q_t$ to Empirical ranks and update
$M_t=M_{t-1}+E_t$ and $G_t=G_{t-1}+\alpha-E_t$.

\par\medskip
\makeatletter
\let\sharedalgorithmstyle\fs@ruled
\def\fs@ruled{\sharedalgorithmstyle
  \def\@fs@pre{}\def\@fs@mid{}\def\@fs@post{}}
\makeatother
\setlength{\intextsep}{0pt}
\setlength{\tabcolsep}{4pt}
\noindent\begin{tabular}{@{}*{2}{L{\dimexpr(\linewidth-4\tabcolsep-2\arrayrulewidth)/3\relax}|}L{\dimexpr(\linewidth-4\tabcolsep-2\arrayrulewidth)/3\relax}@{}}
\noalign{\hrule height .8pt}
\begin{minipage}[t]{\linewidth}\vspace{2pt}
\begin{algorithm}[H]
\caption{Counting}\label{alg:count}
\end{algorithm}
\vspace{2pt}\end{minipage} &
\begin{minipage}[t]{\linewidth}\vspace{2pt}
\begin{algorithm}[H]
\caption{Barrier}\label{alg:sine}
\end{algorithm}
\vspace{2pt}\end{minipage} &
\begin{minipage}[t]{\linewidth}\vspace{2pt}
\begin{algorithm}[H]
\caption{Empirical ranks}\label{alg:rank}
\end{algorithm}
\vspace{2pt}\end{minipage} \\\hline
\begin{minipage}[t]{\linewidth}\vspace{3pt}
\small\raggedright
\textbf{Parameters:} miscoverage level $\alpha\in(0,1)$, horizon $T\in\mathbb N_{\ge1}$.\par
\textbf{Round input:} $t\le T$, cumulative miscoverage count $M_{t-1}$.\par
\textbf{Output:} requested miscoverage probability $q_t\in[0,1]$.
\vspace{3pt}\end{minipage} &
\begin{minipage}[t]{\linewidth}\vspace{3pt}
\small\raggedright
\textbf{Parameter:} miscoverage level $\alpha\in(0,1)$. No horizon is required.\par
\textbf{Round input:} $t\ge1$, miscoverage slack $G_{t-1}$.\par
\textbf{Output:} requested miscoverage probability $q_t\in[0,1]$.
\vspace{3pt}\end{minipage} &
\begin{minipage}[t]{\linewidth}\vspace{3pt}
\small\raggedright
\textbf{Parameter:} fixed nonconformity score $s$.\par
\textbf{Round input:} $t,q_t$, sorted scores $S_{(1)}\le\cdots\le S_{(t-1)}$.\par
\textbf{Output:} prediction set $C_t(X_t)$, miscoverage indicator $E_t$, updated scores.
\vspace{3pt}\end{minipage} \\\hline
\begin{minipage}[t]{\linewidth}\vspace{3pt}
\small\raggedright
\textbf{1. Compute counts once.} Set all $H_t(m)=0$, then $H_T(b_T)=1$.
For $t=T,\ldots,1$ and $m=0,\ldots,b_{t-1}$, set
\[
\begin{aligned}
H_{t-1}(m)&=H_t(m)\\
         &\quad+H_t(m+1).
\end{aligned}
\]
\textbf{2. Read counts.} Set
\[
\begin{aligned}
w_0&=H_t(M_{t-1}),\\
w_1&=H_t(M_{t-1}+1).
\end{aligned}
\]
\textbf{3. Return} $q_t=w_1/(w_0+w_1)$.
\vspace{3pt}\end{minipage} &
\begin{minipage}[t]{\linewidth}\vspace{3pt}
\small\raggedright
\textbf{1. Set the barrier.} Let $L_t=4+\lceil t^{1/3}\rceil$.
Set $\Phi_t(z)=z(L_t-z)$ for $0<z<L_t$, and $0$ otherwise.\par\smallskip
\textbf{2. Compute weights.}
\[
\begin{aligned}
w_0&=(1-\alpha)\Phi_t(G_{t-1}+\alpha),\\
w_1&=\alpha \Phi_t(G_{t-1}+\alpha-1).
\end{aligned}
\]
\textbf{3. Return} $q_t=w_1/(w_0+w_1)$.
\vspace{3pt}\end{minipage} &
\begin{minipage}[t]{\linewidth}\vspace{3pt}
\small\raggedright
\textbf{1.} Set $r_t=\lfloor t(1-q_t)\rfloor$ and $\theta_t=t(1-q_t)-r_t$.\par
\textbf{2.} Draw an independent $\varpi_t\sim\mathrm{Unif}[0,1)$ and set $J_t=r_t+\one\{\varpi_t<\theta_t\}$.\par
\textbf{3.} Set $C_t$ to the empty-set map if $J_t=0$, the full-set map if $J_t=t$, and otherwise $C_{\tau_t}$ with threshold $\tau_t=S_{(J_t)}$.\par
\textbf{4.} Observe $X_t$ and issue $C_t(X_t)$. Observe $Y_t$, store $S_t=s(X_t,Y_t)$, and return $E_t=\one\{Y_t\notin C_t(X_t)\}$.
\vspace{3pt}\end{minipage} \\\hline
\end{tabular}
\end{algorithmgroup}

\textbf{Fixed horizon.}\phantomsection\label{sec:count-main}
Let $\mathcal A_T$ be the set of binary miscoverage paths satisfying every quota and
ending with $M_T=b_T$. After $t$ rounds and $m$ miscoverages, $H_t(m)$ counts
the feasible ways to complete the remaining rounds. At round $t$, Counting sets $q_t$ to
the fraction of feasible completions with $E_t=1$. If only one indicator allows
completion, a full-set or empty-set map forces it.
Under continuous i.i.d.\ scores, each path in $\mathcal A_T$ is equally
likely. This is $P_\alpha(\cdot\mid\mathcal A_T)$, where $P_\alpha$ is
the law of $T$ independent Bernoulli$(\alpha)$ indicators.
Write $\kl(q\Vert\alpha)=q\log(q/\alpha)
+(1-q)\log((1-q)/(1-\alpha))$ for Bernoulli relative entropy,
with $0\log0=0$ and natural logarithms.
Define $\mathcal K_T:=\E\sum_{t\le T}\kl(q_t\Vert\alpha)$.
For Counting, the chain rule gives
$\mathcal K_T=\log[1/P_\alpha(\mathcal A_T)]=O(\log T)$.
Since $\kl(q\Vert\alpha)\ge2(q-\alpha)^2$,
$\E\sum_{t\le T}(q_t-\alpha)^2\le\mathcal K_T/2$.
The count table uses $O(T+\alpha T^2)$ arithmetic operations and stored counts
(Appendix~\ref{app:counting-proof}).

\textbf{Unknown horizon.}\phantomsection\label{sec:anytime}
Algorithm~\ref{alg:sine} weights the two possible next miscoverage slack values
by $\Phi_t$. Near zero it favors larger sets to prevent a quota violation.
Near $L_t$ it favors smaller sets to avoid accumulating unnecessary slack.
A zero weight triggers a full-set or empty-set map.
\phantomsection\label{rem:barrier-potential}
This keeps $0<G_t<L_t$ on every stream, so $\lambda L_t$ bounds
the slack contribution to $\Reg_t$. The potential comparison in Appendix~\ref{app:anytime}
gives $\E\sum_{t\le T}(q_t-\alpha)^2
=O(1+\log L_T+\sum_{t\le T}L_t^{-2})$.
Choosing $L_t\asymp t^{1/3}$ balances the slack bound $L_T$ and
$\sum_{t\le T}L_t^{-2}$, both of order $T^{1/3}$.
Counting reaches $M_T=b_T$ at its specified final round. Barrier instead
keeps slack for future fluctuations and can continue without restarting.
\phantomsection\label{sec:temporal-criterion}

Write $R_t=W_t-tj(\alpha)$ for realized cumulative excess prediction-set size.
Then $\Reg_t=\E_D R_t$ is the constraint price defined in
Section~\ref{sec:setup}. Expected-prefix cost first averages $R_t$ at each
fixed time, then takes the largest value. Retrospective cost first takes
the largest $R_t$ within each run, then averages across runs.
Adaptive stopping uses only observations available when stopping: for example,
stop when $M_t$ first reaches a prescribed level, or at $T$.

\subsection{Empirical-Rank Implementation and Upper Bounds}\label{sec:learning}
\vspace*{-0.05in}

\begin{assumption}[Quadratic bound on excess set size]\label{ass:contact}
There exists $0\le c_+<\infty$ such that every threshold map $C_\tau$,
including the full-set and empty-set maps, satisfies
$0\le V_D(C_\tau)-j(\alpha)+\lambda(m_D(C_\tau)-\alpha)
\le c_+(m_D(C_\tau)-\alpha)^2$.
\end{assumption}
\textbf{Target efficiency and local smoothness.}\phantomsection\label{rem:contact-meaning}
The assumption combines efficiency at the target
\citep{lei2013sets,sadinle2019least} with a quadratic Taylor bound
\citep[Section~9.1.2]{boyd2004convex}.
When expected prediction-set size depends only on miscoverage, write
$h(q)=V_D(C_\tau)$ for $q=m_D(C_\tau)$.
At $q=\alpha$, the assumption requires $h(\alpha)=j(\alpha)$.
Even for a threshold family $\cC$, $j$ permits mixtures and can lie below $h$.
Smoothness or positive estimated curvature alone does not establish this equality.
Conversely, this equality and the bound
$|h'(q)-h'(q')|\le L|q-q'|$ for $q,q'\in
[\alpha-\delta,\alpha+\delta]\subset(0,1)$, with $L\ge0$ and $\delta>0$, give
$h'(\alpha)=-\lambda$ and a local quadratic Taylor bound. Bounded size
extends it globally with $c_+=\max\{L/2,(B+\lambda)/\delta^2\}$
(Appendix~\ref{app:contact}). The example in Section~\ref{sec:framework}
has $c_+=1/4$. A uniform additive gap $\Delta\ge0$ in the assumption adds
$T\Delta$ to the bound in Theorem~\ref{thm:rank}. A gap measured only at $\alpha$
is insufficient.
With the Lipschitz bound above, the target gap
$\Delta_\alpha:=h(\alpha)-j(\alpha)$ adds at most $2T\Delta_\alpha$ to
Theorem~\ref{thm:rank} (Proposition~\ref{prop:ext-target-gap}). Mixing two fixed
levels gives the same bounds with the mixture's gap at $\alpha$, under
Proposition~\ref{prop:ext-mixture}'s local smoothness conditions. This gap is
zero when the implemented population maps, mixed to miscoverage $\alpha$,
attain the benchmark. Levels $0$ and $1$ use the literal full and empty sets.

\textbf{Covariate-dependent example.}
Let $X$ be uniform on $\{0,1\}$, with densities $f_0(y)=3/2-y$ and
$f_1(y)=3/2-y+\sin(2\pi y)/(4\pi)$ on $[0,1]$.
The score $s(x,y)=3/2-f_x(y)$ gives different interval lengths across
covariates and attains $j$ among all measurable randomized maps.
Lemma~\ref{lem:conditional-density} proves $4/27\le j''\le12$, verifying
quadratic upper and lower growth. Estimated scores require a separate efficiency
check. A uniform outcome has $j(q)=1-q$ and all three prices
below one (Appendix~\ref{app:flat-frontier}).

\textbf{Residual scores.}
In a correctly specified location model with symmetric, strictly unimodal
noise, thresholding absolute residuals gives sets where the conditional density
exceeds a common cutoff, attaining the frontier of all measurable maps. Scaled
residuals need not attain this frontier, but attain their own threshold family's
frontier when the density of the absolute noise is nonincreasing. In both cases,
positivity and local Lipschitz continuity of this density near its
$(1-\alpha)$-quantile give
Assumption~\ref{ass:contact} under the bounded-support conditions of
Proposition~\ref{prop:ext-residual}.

Algorithm~\ref{alg:rank} estimates thresholds from past scores and stores
$O(t)$ scores. Barrier itself stores only $G_t$ and $t$.
Define the insertion rank of $S_t$ among the first $t$ scores by
$I_t=1+\sum_{u<t}\one\{S_u<S_t\}$. Write
$\rhist_{t-1}=\sigma(I_u,\varpi_u:u<t)$,
where $\varpi_t\sim\mathrm{Unif}[0,1)$ are independent rounding seeds,
independent of the scores. The request $q_t$ uses this history, while the
threshold also uses the stored scores.
Under continuous i.i.d.\ scores, the fresh score has $t$ equally likely
ranks. Selecting $J_t=j$ covers $j$ of them. Randomizing between adjacent
ranks gives $\E[J_t\mid\rhist_{t-1}]=t(1-q_t)$, avoiding systematic
rounding toward larger sets.

\begin{theorem}[Empirical-rank implementation]\label{thm:rank}
Under Assumptions~\ref{ass:protocol}--\ref{ass:contact} with continuous
fixed-score cdf $F$, let $q_t\in[0,1]$ be $\rhist_{t-1}$-measurable for
each $t\ge1$ and implement it by Algorithm~\ref{alg:rank}. Then
$\Pp(E_t=1\mid\rhist_{t-1})=\E[m_D(C_t)\mid\rhist_{t-1}]=q_t$
and $\Var(m_D(C_t)\mid\rhist_{t-1})\le1/(4t)$.
For every integer $T\ge1$, with $\harm_T=\sum_{t=1}^T1/t$,
$\Reg_T(\mathcal A,D)\le\lambda\E G_T+
c_+\{\E\sum_{t=1}^T(q_t-\alpha)^2+\harm_T/4\}$.
\end{theorem}
\begin{remark}\label{rem:rank-subroutine}
The upper bound includes at most $c_+\harm_T/4=O(\log T)$ for threshold
estimation. Here $q_t$ uses rank history, whereas $p_t$ uses the full observed
history and averages over the current random seed. They need not agree with
each other or with the selected map's miscoverage $m_D(C_t)$.
For example, $q_2=1/2$ selects threshold $S_1$ without rounding, so
$p_2=m_D(C_2)=1-F(S_1)$. Averaging over $S_1$ gives the requested $1/2$.
Counting and Barrier remain feasible with tied scores. The exact rank law and variance
bound require continuity.
The full proof is in Appendix~\ref{app:rank-proof}.
\end{remark}
\textbf{Proof sketch.}
Lemma~\ref{lem:rank-moments} uses uniform insertion ranks
\citep{shafer-vovk} and order-statistic moments to give the requested miscoverage
law and conditional variance bound. The conditional mean squared deviation
of $m_D(C_t)$ from $\alpha$ is
$(q_t-\alpha)^2+\Var(m_D(C_t)\mid\rhist_{t-1})$.
Apply Assumption~\ref{ass:contact} and sum: the linear terms yield
$\lambda\E G_T$, and the conditional variances contribute at most
$c_+\sum_{t\le T}1/(4t)$.

\textbf{Controller upper bounds.}
Combining Theorem~\ref{thm:rank} with Lemmas~\ref{lem:counting}
and~\ref{lem:quadratic-upper} gives $\Reg_T=O(\log T)$ for Counting and
$\max_{t\le T}\Reg_t=O(T^{1/3})$ for one infinite run of Barrier.
Applying the martingale maximal inequality to
\eqref{eq:realized-cost-martingale} gives
$\E\max_{t\le T}R_t=O(\sqrt T)$ for the same run
(Appendix~\ref{app:stopping}). These upper bounds require neither
quadratic lower growth nor knowledge of $D,F,j,\lambda,c_+$.

\subsection{Lower bounds}\label{sec:matching-lower}

\begin{theorem}[Lower bounds]
\label{thm:lower-bounds}\label{thm:anytime-lower}
Suppose Assumption~\ref{ass:protocol} holds and
$\psi(q)\ge c_-(q-\alpha)^2$ for all $q\in[0,1]$, with $c_->0$.
There exist $c_0\ge0$ and $c_1,c_2,c_3>0$, depending only on
$\alpha,B,\lambda,c_-$, such that every integer $T\ge1$ and every
policy $\mathcal A$ feasible through $T$ satisfy
(i) $\Reg_T\ge c_1\log T-c_0$,
(ii) $\max_{1\le t\le T}\Reg_t\ge c_2T^{1/3}-1$, and
(iii) $\E_D\max_{1\le t\le T}R_t\ge c_3\sqrt T$.
For (iii), $\psi(1)>0$ suffices in place of quadratic lower growth,
with $c_3$ depending only on $\alpha,B,\lambda,\psi(1)$.
\end{theorem}
\begin{remark}
The three rates are unavoidable even with knowledge of $D,T$, without
score continuity or Assumption~\ref{ass:contact}. Together with the upper
bounds, they establish the optimal orders in Table~\ref{tab:trichotomy}.
For the continuous example, bound (ii) does not certify
finite-horizon near-optimality at the horizons considered here.
The full proof and constants are in
Appendix~\ref{app:lower-proof}.
\end{remark}
\textbf{Proof sketch.}
Compare with independent Bernoulli$(\alpha)$ miscoverage indicators.
Lemma~\ref{lem:survival} shows that all quotas hold with probability
$O(T^{-1/2})$, whereas a feasible policy assigns this event probability one.
Changing the miscoverage law therefore costs logarithmic relative entropy,
which the quadratic lower bound on $\psi$ converts into excess prediction-set size.
For (ii), generate both
miscoverage sequences using the same uniform random variables.
Lemma~\ref{lem:window-adjustment} shows that fluctuations in each window
require starting slack or reduced miscoverage probabilities. Summing over disjoint
windows gives the $T^{1/3}$ bound for policies using the full history.
For (iii), $\psi(1)>0$ forces square-root fluctuations in
\eqref{eq:realized-cost-martingale}, unless expected terminal cost already has that order.

\subsection{Optimal rates}\label{sec:algorithms-upper}

In each infimum, policies satisfy Definition~\ref{def:valid} through
$T$ and may know $D,T$.

\begin{theorem}[Optimal rates]\label{thm:optimal-prices}\label{thm:sine-upper}
Fix $D$, $\alpha\in(0,1)$, and a score with continuous cdf $F$.
Suppose Assumptions~\ref{ass:protocol}--\ref{ass:contact} hold and
$\psi(q)\ge c_-(q-\alpha)^2$ for every $q\in[0,1]$, with $c_->0$.
As $T\to\infty$, the optimal prices are
(i) $\inf_{\mathcal A\ \mathrm{feasible}}\Reg_T(\mathcal A,D)=\Theta(\log T)$,
(ii) $\inf_{\mathcal A\ \mathrm{feasible}}\max_{1\le t\le T}\Reg_t(\mathcal A,D)=\Theta(T^{1/3})$,
and (iii) $\inf_{\mathcal A\ \mathrm{feasible}}\E_D\max_{1\le t\le T}R_t=\Theta(\sqrt T)$.
The constants depend only on $\alpha,B,\lambda,c_+,c_-$.
Algorithm~\ref{alg:count} attains (i). One infinite run of
Algorithm~\ref{alg:sine} attains (ii)--(iii). With
Algorithm~\ref{alg:rank}, both satisfy every quota on every stream, and
Algorithm~\ref{alg:count} has $M_T=b_T$.
\end{theorem}
\begin{remark}\label{rem:temporal-criteria}
At a bounded stopping time, the martingale term in the realized-cost
decomposition has zero expectation. The slack and accumulated nonnegative
cost then give the same $T^{1/3}$ bound
(Proposition~\ref{prop:stopped-cost} and Theorem~\ref{thm:lower-bounds}).
The full proof is in
Appendix~\ref{app:optimal-proof}.
Appendix~\ref{app:fixed-buffer-prefix} extends the expected-prefix result
to a fixed initial miscoverage allowance under the same assumptions.
\end{remark}
\textbf{Proof sketch.}
The upper bounds in Section~\ref{sec:learning} and the uniform lower
bounds in Theorem~\ref{thm:lower-bounds} prove the three optimal orders.
Table~\ref{tab:trichotomy} summarizes the criteria and attaining policies.

\begingroup
\subsection{Implications beyond the oracle model}\label{sec:ext-implications}
Our controllers satisfy the quota on every stream. The lower bounds for
i.i.d.\ data with known $D,T$ remain worst-case bounds in broader models
containing $D$, with the same family and benchmark. Matching upper bounds
use stationarity.

\textbf{Existing methods.}
Adaptive conformal inference (ACI) \citep{gibbs21} with fixed step $\gamma$,
no projection, and the full-set convention in Proposition~\ref{prop:ext-aci}
satisfies $M_t\le b_t+k_\gamma$ on every stream, where
$k_\gamma=\lceil\alpha/\gamma+1-\alpha\rceil$.
Under the lower-bound assumptions, fixed $\gamma$ gives expected-prefix cost
$\Omega(T^{1/3})$. Allowances $o(T^{1/3})$ cannot improve this order
(Appendix~\ref{app:fixed-buffer-prefix}). With population thresholds, the
excess cost is asymptotically linear in $T$.
The quota guard returns the full set if another error could violate the quota,
ensuring feasibility under drift with immediate labels
(Section~\ref{sec:experiments}).

\textbf{Relaxed conditions.}
Delaying labels by at most a fixed $d$ rounds preserves all three rates.
Restarting over $N$ stationary periods fixed in advance, with refitting between
them and common bounds on the geometric constants, gives expected-prefix cost
$O(N^{2/3}T^{1/3})$ relative to the periods' own
benchmarks (Propositions~\ref{prop:ext-delay} and~\ref{prop:ext-restart}).
Remark~\ref{rem:ext-ties} covers ties.
\endgroup

\par\vspace{-0.15in}
\section{Experiments}\label{sec:experiments}
\vspace{-0.05in}
We examine evaluation time, threshold estimation, and coverage control
in Theorems~\ref{thm:decomposition}--\ref{thm:optimal-prices}.
The continuous model in Section~\ref{sec:framework} satisfies the quadratic
conditions. We use California Housing to evaluate learned scores.
Unless stated otherwise, \emph{Counting} and \emph{Barrier} mean
Algorithms~\ref{alg:count} and~\ref{alg:sine} with default settings.
\emph{Empirical} thresholds use Algorithm~\ref{alg:rank}, while \emph{known $F$}
uses population quantiles. This comparison keeps the controller fixed.
Main figures use $\alpha=0.1$. Other targets and variants are in Appendix~\ref{app:experiments}.

Section~5.1 compares the three cost criteria in Table~\ref{tab:trichotomy}
using a model where the assumptions hold exactly. Section~5.2
isolates the cost of not knowing $F$. Section~5.3 compares our controllers
with the finite-horizon optimum and tuned guarded ACI.
Section~5.4 measures the quota's cost for learned scores, where
we cannot verify Assumption~\ref{ass:contact}.

\begin{figure}[t]
\centering
\begin{minipage}[t]{.32\linewidth}\centering
\begin{tikzpicture}
\begin{axis}[width=\linewidth,height=3.25cm,title={(a) Counting: terminal},
 title style={font=\footnotesize},xlabel={$\log T$},ylabel={$\Reg_T$},
 label style={font=\footnotesize},tick label style={font=\footnotesize},
 grid=major,grid style={gray!15},axis lines=left,
 legend style={font=\footnotesize,draw=none,at={(.5,0)},anchor=north,yshift=-30pt}]
\addplot[black,thick,mark=triangle*,mark size=1pt] table[x=logT,y=optimum]{plot_data_12.dat};
\addlegendentry{Optimum}
\addplot[teal!80!black,dashed,mark=o,mark size=1.2pt] table[x=logT,y=known_f]{plot_data_12.dat};
\addlegendentry{Known $F$}
\addplot[blue!70!black,thick,mark=*,mark size=1pt] table[x=logT,y=empirical]{plot_data_12.dat};
\addlegendentry{Empirical}
\end{axis}
\end{tikzpicture}
\end{minipage}\hfill
\begin{minipage}[t]{.32\linewidth}\centering
\begin{tikzpicture}
\begin{axis}[width=\linewidth,height=3.25cm,title={(b) Barrier: prefixes},
 title style={font=\footnotesize},xlabel={$T$},ylabel={$\max_{t\le T}\Reg_t$},
 label style={font=\footnotesize},tick label style={font=\footnotesize},xmode=log,ymode=log,
 grid=major,grid style={gray!15},axis lines=left,
 legend style={font=\footnotesize,draw=none,at={(.5,0)},anchor=north,yshift=-30pt}]
\addplot[teal!80!black,dashed,mark=o,mark size=1.2pt] table[x=T,y=expected_known_f]{plot_data_11.dat};\addlegendentry{Known $F$}
\addplot[blue!70!black,thick,mark=*,mark size=1pt] table[x=T,y=expected_empirical]{plot_data_11.dat};\addlegendentry{Empirical}
\addplot[orange!80!black,dotted,mark=triangle*,mark size=1pt,error bars/.cd,y dir=both,y explicit]
 table[x=T,y=guard,y error=guard95]{plot_data_11.dat};\addlegendentry{Quota guard}
\addplot[gray,dashed,forget plot] table[x=T,y=cube_guide]{plot_data_11.dat};
\end{axis}
\end{tikzpicture}
\end{minipage}\hfill
\begin{minipage}[t]{.32\linewidth}\centering
\begin{tikzpicture}
\begin{axis}[width=\linewidth,height=3.25cm,title={(c) Barrier: worst prefix},
 title style={font=\footnotesize},xlabel={$T$},ylabel={$\E\max_{t\le T}R_t$},
 label style={font=\footnotesize},tick label style={font=\footnotesize},xmode=log,ymode=log,
 grid=major,grid style={gray!15},axis lines=left,
 legend style={font=\footnotesize,draw=none,at={(.5,0)},anchor=north,yshift=-30pt}]
\addplot[teal!80!black,dashed,mark=o,mark size=1.2pt,error bars/.cd,y dir=both,y explicit]
 table[x=T,y=retrospective_known_f,y error=retrospective_known_f95]{plot_data_11.dat};\addlegendentry{Known $F$}
\addplot[blue!70!black,thick,mark=*,mark size=1pt,error bars/.cd,y dir=both,y explicit]
 table[x=T,y=retrospective_empirical,y error=retrospective_empirical95]{plot_data_11.dat};\addlegendentry{Empirical}
\addplot[orange!80!black,dotted,mark=triangle*,mark size=1pt,error bars/.cd,y dir=both,y explicit]
 table[x=T,y=guard,y error=guard95]{plot_data_11.dat};\addlegendentry{Quota guard}
\addplot[gray,dashed,forget plot] table[x=T,y=sqrt_guide]{plot_data_11.dat};
\end{axis}
\end{tikzpicture}
\end{minipage}
\caption{Three criteria for excess prediction-set size in the continuous model
($\alpha=0.1$). (a) Counting's terminal cost versus the numerical optimum under
all prefix quotas with known $F,T$. (b) Largest expected cost over prefixes.
(c) Average of each run's largest realized cost. Known $F$ uses population
quantiles. Empirical uses Algorithm~\ref{alg:rank}. Panels (b)--(c) compare
default Barrier with the quota guard. Panel (a) and Barrier in (b) use numerical
expectations. Other curves use $512$ streams with pointwise $\pm1.96$
standard-error bars. Gray guides show $T^{1/3}$ in (b) and $\sqrt T$ in (c),
with arbitrary vertical scales.}
\label{fig:three-criteria}
\end{figure}
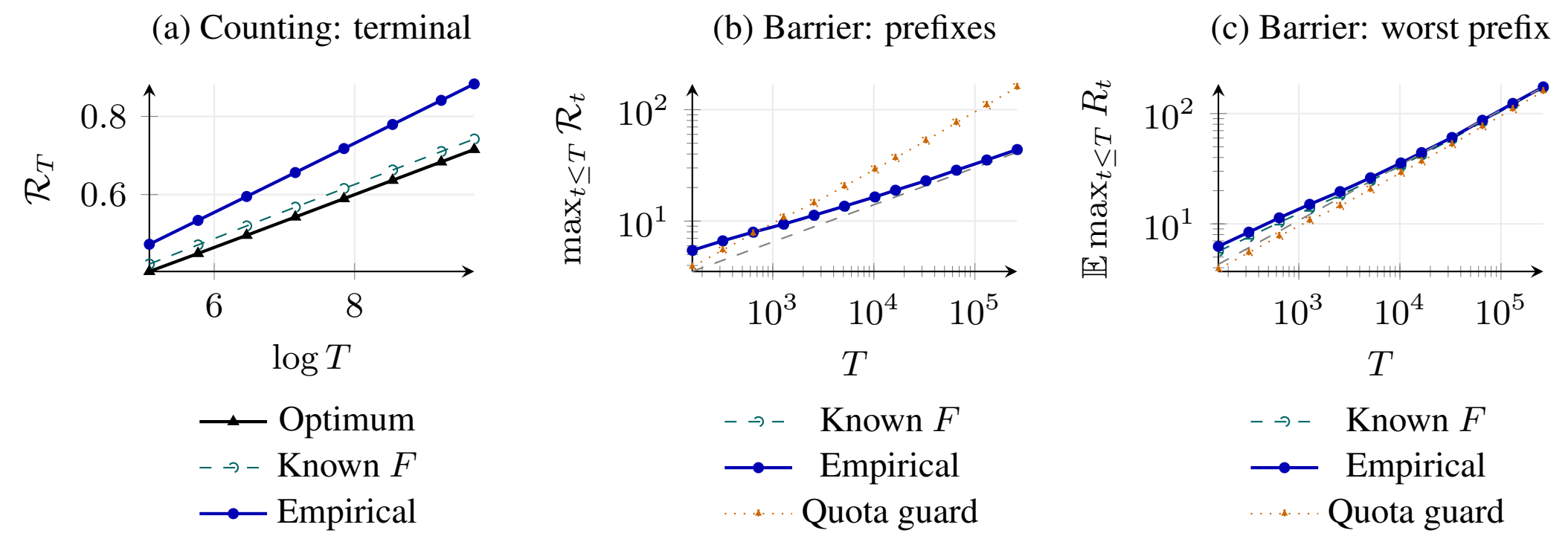

\subsection{Evaluation time changes the cost}
\vspace{-0.05in}
In Figure~\ref{fig:three-criteria}a, Counting is rerun for each announced
horizon. It stays close to the numerical terminal optimum,
but its low terminal cost conceals much higher costs at earlier prefixes
(Figure~\ref{fig:curvature-validation}a). Terminal efficiency need not
persist throughout a run
(Corollary~\ref{cor:terminal-prefix-tradeoff}).

Panels (b)--(c) use one horizon-free Barrier controller, with
$L_t=4+\lceil t^{1/3}\rceil$. Panel (b) takes the largest expected cost
over all prefixes, computed from the controller's state probabilities.
Panel (c) instead averages the largest realized excess size
$R_t=W_t-tj(\alpha)$ on each of $512$ independent score streams.
The quota guard proposes a population threshold with miscoverage $\alpha$
and switches to the full set whenever another miscoverage could exceed $b_t$.
At $T=262{,}144$, the quota guard has higher expected-prefix cost than
Barrier but lower estimated retrospective cost.
Normalized curves still drift at these horizons (Appendix~\ref{app:additional-rates}).

\subsection{The cost of estimating thresholds}

The known-$F$ and empirical curves have the same miscoverage-count law
and expected slack (Theorem~\ref{thm:rank}). In this quadratic
model, their expected terminal costs differ by one quarter of the accumulated
order-statistic variance, at most $\harm_T/16$.
At $T=16{,}380$, estimating thresholds adds $0.141$ for Counting and
$0.144$ for Barrier, below the bound $0.643$.
These numerical expectations show a small estimation cost. The near
overlap of the curves does not reflect Monte Carlo uncertainty.
The expected-size bound does not cover differences between retrospective
maxima (Appendix~\ref{app:rank-cost}).

\subsection{Finite-horizon control and baselines}

Optimal asymptotic rates need not give the smallest finite-horizon cost.
At $T=16{,}380$, the numerical expected-prefix optimum lies in
$[5.6547,5.6598]$, whereas the default Barrier costs $18.926$, about
$3.34$ times the feasible upper bound in excess cost (Figure~\ref{fig:followup}a).
Theorem~\ref{thm:decomposition} explains most of this cost as unused quota:
its slack contribution is $18.043$.
The near-optimal policy builds slack early and uses it near
the final round. A separately validated width variant costs $11.484$.
Its known-$F$ counterpart still reserves slack for continued operation
(Figure~\ref{fig:policy-diagnostics}).
The remaining gap cannot be attributed to threshold estimation alone.

Guarded adaptive conformal inference (ACI) issues the full set when a miscoverage would
violate $b_t$. Its internal update uses the original proposal's miscoverage indicator.
Figure~\ref{fig:followup}b compares horizon-selected and decaying
steps with the two Barrier widths. At the largest horizon, both ACI schedules have lower cost than
the default Barrier, while the width variant has lower cost than decaying
ACI. The finite-horizon ranking therefore depends on the settings.
Horizon-selected steps use independent validation for each horizon.
The decaying multiplier and Barrier width are each selected once and
frozen across test horizons (Appendix~\ref{app:aci-horizons}).
ACI with horizon-selected steps is retuned at each horizon. Barrier
uses one horizon-free run.

\subsection{Learned scores and online performance}

\begin{figure}[!ht]
\centering
\begin{minipage}[t]{.47\linewidth}\vspace{0pt}\centering
\begin{tikzpicture}
\begin{axis}[width=\linewidth,height=3.15cm,title={(a) Miscoverage and length},
 title style={font=\footnotesize},xlabel={Miscoverage probability},ylabel={Mean length},
 label style={font=\footnotesize},tick label style={font=\footnotesize},xmin=.02,xmax=.3,
 xtick={.02,.1,.2,.3},xticklabels={0.02,0.1,0.2,0.3},
 grid=major,grid style={gray!15},axis lines=left,
 legend style={font=\footnotesize,draw=none,at={(.5,0)},anchor=north,yshift=-30pt,legend columns=2,/tikz/column sep=6pt}]
\addplot[blue!70!black,thick] table[x=risk,y=linear]{plot_data_05.dat};\addlegendentry{Ridge}
\addplot[orange!80!black,dashed,thick] table[x=risk,y=trees]{plot_data_05.dat};\addlegendentry{Extra Trees}
\end{axis}
\end{tikzpicture}
\end{minipage}\hfill
\begin{minipage}[t]{.47\linewidth}\vspace{0pt}\centering
\begin{tikzpicture}
\begin{axis}[width=.90\linewidth,height=3.15cm,title={(b) Barrier minus guarded ACI},
 title style={font=\footnotesize},xlabel={Frozen split},ylabel={Length difference},
 label style={font=\footnotesize},tick label style={font=\footnotesize},xmin=.5,xmax=5.5,
 xtick={1,2,3,4,5},ymin=-.010,ymax=.0005,scaled y ticks=false,
 yticklabel style={/pgf/number format/fixed,/pgf/number format/precision=3},
 grid=major,grid style={gray!15},axis lines=left,
 legend style={font=\footnotesize,draw=none,at={(.5,0)},anchor=north,yshift=-30pt,legend columns=2,/tikz/column sep=6pt}]
\addplot[gray,dashed,forget plot] coordinates {(.5,0)(5.5,0)};
\addplot[blue!70!black,only marks,mark=*,error bars/.cd,y dir=both,y explicit]
 table[x expr=\thisrow{split}-.12,y=linear,y error=linear_err]{plot_data_04.dat};\addlegendentry{Ridge}
\addplot[orange!80!black,only marks,mark=triangle*,error bars/.cd,y dir=both,y explicit]
 table[x expr=\thisrow{split}+.12,y=trees,y error=trees_err]{plot_data_04.dat};\addlegendentry{Extra Trees}
\end{axis}
\end{tikzpicture}
\end{minipage}
\caption{California Housing with frozen predictors. (a) Empirical curves
averaged over five test splits. (b) Default Barrier (Algorithms~\ref{alg:sine}
and~\ref{alg:rank}) minus original guarded ACI at $\alpha=0.1$. Negative values mean shorter
intervals for Barrier. Each split uses 32 paired test permutations and
every test row once per run. Bars are $\pm1.96$ ordering SEs, excluding
training and population uncertainty.}
\label{fig:real-followup}
\end{figure}
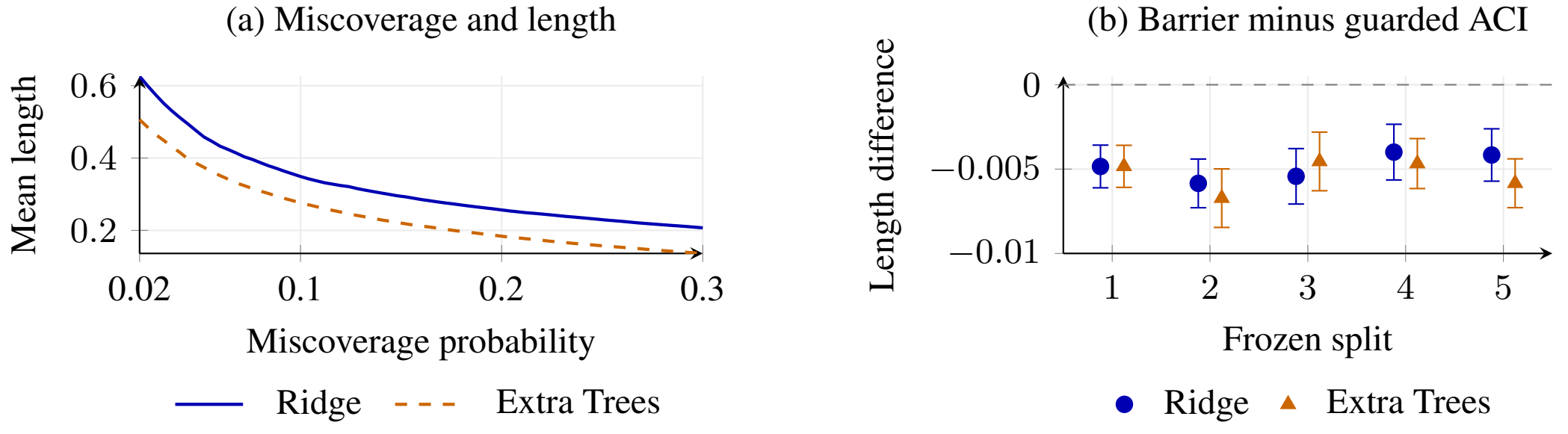

California Housing uses five frozen splits, each with separate training,
validation, and test observations, and two predictors: Ridge and Extra Trees.
Normalized responses lie in $[0,1]$, giving $B=1$ for interval length.
The size curves in Figure~\ref{fig:real-followup}a bend upward near the target,
but mixing thresholds can reduce size at the same miscoverage. At $\alpha=0.1$,
empirical gaps above the convex envelope range from $0$ to $0.00293$ for Ridge and $0.000033$ to
$0.00183$ for Extra Trees (Table~\ref{tab:contact-gaps}). These diagnostics
establish neither target efficiency nor global quadratic lower growth, and cannot replace
the uniform additive gap $\Delta$ that adds $T\Delta$ to the upper bound
(Appendices~\ref{app:contact} and~\ref{app:real-geometry}).

Controller settings are selected using $16$ validation permutations and frozen before
the $32$ paired test permutations. Each run starts with no past scores and
uses every test observation once. At $\alpha=0.1$, all splits select the
default Barrier width. Figure~\ref{fig:real-followup}b compares it with
original guarded ACI. Barrier reduces mean interval length by $1.33\%$ for Ridge and $1.83\%$ for
Extra Trees. With the same empirical-rank implementation for ACI,
these reductions are $1.68\%$ and $2.21\%$.
Table~\ref{tab:matched-rank-aci} reports absolute lengths, paired differences,
miscoverage, and full/empty frequencies. Every tested guarded prefix satisfies $M_t\le b_t$,
and empty outputs remain rare. Uncertainty intervals quantify ordering and controller
randomness on fixed test sets, excluding training and population uncertainty. Appendix~\ref{app:guard-ablation} isolates the guard's added length using
paired ACI runs with identical proposals.
Unguarded ACI exceeds the quota in every run, by up to $12$ errors.
The guard removes every violation and increases mean interval length by
$2.9\%$ (Ridge) and $4.1\%$ (Extra Trees). For the fixed score's threshold
family, simultaneous $95\%$ upper bounds on the target gap are $0.158$--$0.175$.
These conservative bounds cannot certify the $T^{-2/3}$ gap needed for a
$T^{1/3}$ cost contribution (Appendix~\ref{app:ext-gap}).

\FloatBarrier

\section{Discussion}\label{sec:conclusion}

We characterize the efficiency cost of enforcing hard cumulative coverage in online conformal prediction. Under suitable geometric conditions, the optimal excess prediction-set size scales as $\Theta(\log T)$, $\Theta(T^{1/3})$, or $\Theta(\sqrt{T})$, depending on when performance is evaluated. These rates remain achievable using empirical ranks without knowing the score distribution.

\textbf{Limitations and future work.} Our efficiency guarantees rely on a stationary frontier and uniform insertion ranks, assumptions that may fail under distribution shift or when data are reused for fitting and calibration. The application examples motivate the tradeoff, but their feedback availability and frontier conditions need validation. Extending the theory to drifting distributions, alternative comparators, calibration error, and more realistic deployment constraints is an important direction for future work. Appendix~\ref{app:extensions-applications} gives extensions and their proofs.

\FloatBarrier
\label{m:end-main}

\clearpage
\subsection*{Acknowledgments}
This work is supported by NSF grants \#2436217, \#2425687, and \#2601681.

\subsection*{AI use statement}
The core research ideas and initial proof sketches were developed by the
author. Generative AI tools assisted in developing these sketches into
detailed proofs, refining mathematical arguments, designing and implementing
experiments, checking numerical results, and revising the manuscript.
The author is responsible for all final claims, proofs, code, and
experimental results.

\subsection*{Ethics statement}
The experiments use public benchmark data and do not evaluate
suitability for a particular deployment.

\bibliographystyle{iclr2027_conference}
\bibliography{references}
\appendix

\section{Proof of Theorem~\ref{thm:decomposition}}
\label{app:decomposition}\label{app:conformal}
The full-set and empty-set maps attain pairs of miscoverage and expected size $(0,B)$ and $(1,0)$,
and their mixtures attain every $q\in[0,1]$. Thus $0\le j\le B$ and $j(1)=0$.
Combining nearly optimal mixtures interpolates miscoverage and expected size.
Letting their approximation errors vanish proves convexity of $j$.
For $q<q'\le1$, combining a mixture of miscoverage $q$
with the empty-set map with weight $(q'-q)/(1-q)$ raises its miscoverage to
$q'$ without increasing size. Thus $j$ is nonincreasing and has a finite
supporting slope $-\lambda\le0$ at every interior target $\alpha$.

\begin{lemma}[One-round identity]\label{lem:one-round-identity}
Under Assumption~\ref{ass:protocol}, at every round $t$,
\[
 \E[\ell_{C_t}(X_t)\mid\history_{t-1}]
 =j(\alpha)-\lambda(p_t-\alpha)+\psi(p_t)+\epsilon_t.
\]
\end{lemma}
\begin{proof}
Conditional on $\history_{t-1}$, the selected map is independent of the
fresh pair and its law is an admissible mixture in the definition of $j$.
Averaging over the current seed gives $p_t$ as the mixture's miscoverage probability.
Its conditional expected size is $j(p_t)+\epsilon_t$. Expand $j(p_t)$
using the definition of $\psi$.
\end{proof}

\begin{proof}[Proof of Theorem~\ref{thm:decomposition}]
Taking expectations in Lemma~\ref{lem:one-round-identity} and summing gives
\[
 \Reg_T=-\lambda\E\sum_{t\le T}(p_t-\alpha)
       +\E\sum_{t\le T}\psi(p_t)+\E\sum_{t\le T}\epsilon_t.
\]
The tower property gives $\E\sum_{t\le T}p_t=\E M_T$, so the first term
is $\lambda\E G_T$. Bounded sizes and probabilities ensure integrability
for finite $T$.
\end{proof}

\subsection{Shared probability tools}\label{app:proof-tools}
The following facts are used in the upper bounds, lower bounds, and
empirical-rank analysis. They do not require either controller.

\paragraph{Relative entropy and squared deviations.}
For probability laws $P,Q$, write
$\KL(P\Vert Q)=\E_P\log(dP/dQ)$ when $P$ is absolutely continuous
with respect to $Q$, and $+\infty$ otherwise.
For $q,q'\in[0,1]$, its Bernoulli case is
$\kl(q\Vert q')=\KL(\operatorname{Bernoulli}(q)\Vert\operatorname{Bernoulli}(q'))$,
agreeing with Section~\ref{sec:controllers} when $q'=\alpha$.
Both denote relative entropy; lowercase $\kl$ takes probabilities, while
uppercase $\KL$ takes probability laws. In particular,
$\kl(0\Vert0)=\kl(1\Vert1)=0$.
For $q\in[0,1]$ and $\alpha\in(0,1)$,
\begin{align}
 \kl(q\Vert\alpha)&\le\frac{(q-\alpha)^2}{\alpha(1-\alpha)},
 \label{eq:binary-kl-upper}\\
 \kl(q\Vert\alpha)&\ge2(q-\alpha)^2.
 \label{eq:pinsker-binary}
\end{align}
The first follows from $\log x\le x-1$. The second follows by integrating
$\partial_q^2\kl(q\Vert\alpha)=1/[q(1-q)]\ge4$ twice from $\alpha$.
Both extend to the endpoints by continuity.

\paragraph{Realized excess size.}
Recall $R_t=W_t-tj(\alpha)$. With $p_t,\epsilon_t$ from
Theorem~\ref{thm:decomposition}, define
\begin{align*}
 D_t&=\psi(p_t)+\epsilon_t\ge0,\qquad
 \mathsf Z_t=\sum_{u\le t}\xi_u,\\
 \xi_t&=\ell_{C_t}(X_t)-\E[\ell_{C_t}(X_t)\mid\history_{t-1}]
             +\lambda(E_t-p_t).
\end{align*}
Then $\E[\xi_t\mid\history_{t-1}]=0$ and $|\xi_t|\le B+\lambda$.
Thus $\mathsf Z_t$ is a martingale with respect to $(\history_t)$.
Expanding $D_t$ and summing yields the pathwise identity
\begin{equation}\label{eq:realized-cost-martingale}
 R_t=\lambda G_t+\sum_{u\le t}D_u+\mathsf Z_t.
\end{equation}
This identity holds for every policy under Assumption~\ref{ass:protocol},
whether or not it is feasible. Taking expectations recovers
Theorem~\ref{thm:decomposition}.

\section{Proof of Theorem~\ref{thm:rank} and upper bounds}\label{app:rank-proof}
\label{app:rank-moments}
We first prove the empirical-rank law and size bound in
Theorem~\ref{thm:rank}, then establish the controller upper bounds stated
in Section~\ref{sec:learning}. The proof uses rank symmetry,
order-statistic moments, and the stated assumptions.

\begin{lemma}[Rank law and moments]\label{lem:rank-moments}
For continuous i.i.d.\ scores and $\rhist_{t-1}$-measurable requests,
Algorithm~\ref{alg:rank} satisfies
$\Pp(E_t=1\mid\rhist_{t-1})=\E[m_D(C_t)\mid\rhist_{t-1}]=q_t$
and $\Var(m_D(C_t)\mid\rhist_{t-1})\le1/(4t)$.
\end{lemma}
\begin{proof}
\textbf{Conditional miscoverage law.}
For any $n$, the permutation ordering $n$ i.i.d.\ continuous scores is
uniform over the $n!$ permutations and independent of their ordered values.
Its insertion ranks form a bijection with
$\prod_{i=1}^n\{1,\ldots,i\}$. Thus the $I_t$ are independent uniforms on
$\{1,\ldots,t\}$, and the sorted first $t-1$ scores are independent of
$\rhist_{t-1}$, including its independent rounding coins.

Write $r_t=\lfloor t(1-q_t)\rfloor$ and
$\theta_t=t(1-q_t)-r_t$. Algorithm~\ref{alg:rank} selects
$J_t=r_t+\one\{\varpi_t<\theta_t\}$.
For an interior rank, the fresh score exceeds the $J_t$-th past score
exactly when $I_t>J_t$. The same relation holds for the literal empty-set and
full-set maps at $J_t=0,t$, so $E_t=\one\{I_t>J_t\}$.
Past miscoverage indicators are therefore rank-history measurable. The new rank is
independent of $\rhist_{t-1}$ and $\varpi_t$, giving miscoverage probability $1-J_t/t$
conditional on $\rhist_{t-1}$ and $J_t$. Since
$\E[J_t\mid\rhist_{t-1}]=t(1-q_t)$, averaging over $\varpi_t$ gives
\begin{equation}\label{eq:rank-history-law}
 \Pp(E_t=1\mid\rhist_{t-1})=q_t.
\end{equation}
\textbf{Conditional moments.}
The probability integral transform gives independent uniforms $U_i=F(S_i)$.
Continuity suffices: $F$ need not be strictly increasing, as the transform
preserves ordering almost surely.
Let $U_{(j)}$ be the $j$-th order statistic of the first $t-1$ uniforms,
with $U_{(0)}=0$ and $U_{(t)}=1$. Then
$m_D(C_t)=1-U_{(J_t)}$, including the literal endpoints.
For $1\le j\le t-1$, the density of $U_{(j)}$ is
\[
 \frac{(t-1)!}{(j-1)!(t-1-j)!}\,u^{j-1}(1-u)^{t-1-j},\qquad 0<u<1.
\]
Integrating gives
\begin{equation}\label{eq:order-moments}
 \E U_{(j)}=\frac jt,\qquad
 \E U_{(j)}^2=\frac{j(j+1)}{t(t+1)},\qquad
 \Var(U_{(j)})=\frac{j(t-j)}{t^2(t+1)}.
\end{equation}
The formulas also hold at the endpoints, including $t=1$.

Condition on $\rhist_{t-1}$ and abbreviate $q=q_t$, $\theta=\theta_t$,
$a=t(1-q)$. The sorted values retain their unconditional law and are
independent of the new coin. Therefore
\begin{equation}\label{eq:rank-mean}
 \E[J_t\mid\rhist_{t-1}]=a,\quad
 \Var(J_t\mid\rhist_{t-1})=\theta(1-\theta),\quad
 \E[m_D(C_t)\mid\rhist_{t-1}]=q.
\end{equation}
The law of total variance gives
\begin{align}
 \Var(m_D(C_t)\mid\rhist_{t-1})
 &=\frac{\E[J_t(t-J_t)\mid\rhist_{t-1}]}{t^2(t+1)}
          +\frac{\theta(1-\theta)}{t^2}\nonumber\\
 &=\frac{q(1-q)}{t+1}+\frac{\theta(1-\theta)}{t(t+1)}
 \le\frac1{4t}.\label{eq:rank-variance}
\end{align}
The second equality uses
$\E[J_t(t-J_t)\mid\rhist_{t-1}]=a(t-a)-\theta(1-\theta)$.
The last inequality uses $q(1-q),\theta(1-\theta)\le1/4$.
The rounding variance vanishes
when $t(1-q)$ is an integer.
The mean in \eqref{eq:rank-mean} removes the cross term
in $(m_D(C_t)-\alpha)^2$. Thus
\begin{equation}\label{eq:rank-energy}
 \E\sum_{t=1}^T(m_D(C_t)-\alpha)^2
 \le\E\sum_{t=1}^T(q_t-\alpha)^2+\frac{\harm_T}{4}.
\end{equation}
\end{proof}

\begin{proof}[Proof of Theorem~\ref{thm:rank}]
Lemma~\ref{lem:rank-moments} gives the miscoverage law and moments.
Conditional on the full past and the new coin, the selected map is fixed
and the fresh pair has law $D$. Assumption~\ref{ass:contact} gives
\[
 \E[\ell_{C_t}(X_t)\mid\history_{t-1},\varpi_t]
 \le j(\alpha)-\lambda(m_D(C_t)-\alpha)
                         +c_+(m_D(C_t)-\alpha)^2.
\]
This includes the literal endpoints
$J_t=0,t$ and the first round.
The tower property gives
$\E\sum_{t\le T}m_D(C_t)=\E M_T=\alpha T-\E G_T$.
Sum and apply \eqref{eq:rank-energy} to prove the bound.
\end{proof}

\paragraph{From controller requests to prediction-set size.}
Theorem~\ref{thm:rank} and \eqref{eq:pinsker-binary} give
\begin{equation}\label{eq:rank-general-transfer}
 \Reg_T\le\lambda\E G_T+
 c_+\left\{\frac{\mathcal K_T}{2}+\frac{\harm_T}{4}\right\},
 \qquad\mathcal K_T=\E\sum_{t=1}^T\kl(q_t\Vert\alpha),
 \quad\harm_T=\sum_{t=1}^T\frac1t.
\end{equation}
Counting and Barrier request $q_t$ using only time and past miscoverage
indicators (and the announced $T$ for Counting). Since past indicators are
rank-history measurable under Algorithm~\ref{alg:rank}, both requests meet
the measurability condition of Theorem~\ref{thm:rank}. Its conditional law
and the tower property give
\[
 \Pp(E_t=1\mid E_1,\ldots,E_{t-1})
 =\E[q_t\mid E_1,\ldots,E_{t-1}]=q_t.
\]
Thus the binary-path calculations below apply to the implemented
controllers. The additional term $c_+\harm_T/4=O(\log T)$ accounts for
threshold estimation.

\subsection{Fixed-horizon upper bound}\label{app:counting-proof}\label{app:upper}
Let $\mathcal A_T$ be the binary paths satisfying every quota and ending
with $b_T=\lfloor\alpha T\rfloor$ miscoverages, and let $P_\alpha$ be the law
of $T$ independent Bernoulli$(\alpha)$ variables. The path with entries
$b_t-b_{t-1}$ shows that $\mathcal A_T$ is nonempty.

\begin{lemma}\label{lem:counting}
Algorithm~\ref{alg:count} with literal endpoint actions satisfies every
prefix quota and $M_T=b_T$ on every stream. If its miscoverage indicators have the
requested conditional probabilities given their binary history, their law
is $Q=P_\alpha(\cdot\mid\mathcal A_T)$ and
\begin{equation}\label{eq:entropy-id}
 \mathcal K_T:=\E_Q\sum_{t=1}^T\kl(q_t\Vert\alpha)
 =\log\frac1{P_\alpha(\mathcal A_T)}
 \le\log T+\log(T+1)+\frac1{\alpha(1-\alpha)T}.
\end{equation}
Moreover, $\E_Q\sum_{t=1}^T(q_t-\alpha)^2\le\mathcal K_T/2$.
\end{lemma}
\begin{proof}
The recursion for $H_t(m)$ counts valid continuations. From a state with
positive count, an outcome with zero count has probability zero and is
prevented by the literal full-set or empty-set map. Otherwise either outcome is
valid. Induction proves feasibility and the terminal total on every stream.

Under the requested conditional probabilities, the continuation-count
ratios telescope to $1/H_0(0)$: each numerator is the next state's count,
and the final count is one. Thus the miscoverage law is uniform on
$\mathcal A_T$, which is also the conditional law $Q$:
all paths with $b_T$ ones have the same $P_\alpha$ probability.
The relative-entropy chain rule gives
$\mathcal K_T=\KL(Q\Vert P_\alpha)$.

To bound $\mathcal K_T$, condition on $N_T=b_T$, where $N_T$ is the total
under $P_\alpha$. Every cyclic orbit of such words has a quota-feasible
rotation: subtract $b_T/T$ from each entry and rotate after a largest
partial sum. The new centered partial sums are nonpositive differences
from this maximum, including after wrapping since the centered total is
zero. Thus the
integer partial sums are at most $\lfloor\alpha t\rfloor$.
Each orbit has at most $T$ words, hence
$P_\alpha(\mathcal A_T\mid N_T=b_T)\ge1/T$.
Set $u=b_T/T$. The count $b_T$ is a mode of Binomial$(T,u)$, including
$b_T=0$, and therefore has probability at least $1/(T+1)$. Comparing
the same count under parameters $u$ and $\alpha$ gives
\[
 P_\alpha(\mathcal A_T)
 \ge\frac{\exp\{-T\kl(u\Vert\alpha)\}}{T(T+1)}.
\]
Since $|u-\alpha|<1/T$, the upper relative-entropy bound
\eqref{eq:binary-kl-upper} proves \eqref{eq:entropy-id}.
The lower relative-entropy bound \eqref{eq:pinsker-binary} bounds squared
request deviations.
\end{proof}

The binary-history law established above
verifies the hypothesis of Lemma~\ref{lem:counting} for empirical ranks.
Applying \eqref{eq:rank-general-transfer} then yields
\[
 \Reg_T\le\lambda(\alpha T-b_T)
       +c_+\left(\frac{\mathcal K_T}{2}+\frac{\harm_T}{4}\right)
 =O(\log T).
\]
The continuation table uses $O(T+\alpha T^2)$ arithmetic operations and storage
entries. This does not bound bit complexity for its integer counts.
The implementation uses rolling log counts with
\texttt{np.logaddexp}, stores double-precision request arrays, and uses
finite-support masks to set forced requests to exactly zero or one.
At the largest Counting horizon $T=16{,}380$, for
$\alpha=(0.05,0.1,0.2)$, median construction times over three fresh
processes per target were $(0.114,0.170,0.289)$ seconds.
The request arrays occupied $(51.2,102.4,204.8)$ MiB. The largest process
peak resident set size across repeats was $(98.6,158.7,277.1)$ MiB, including Python,
NumPy, and temporary arrays.
These measurements used an Apple M5 Pro with 48 GiB RAM, macOS 26.4.1
(arm64), Python 3.12.14, and NumPy 2.3.5. Timing excludes interpreter
startup and file I/O.
An independent integer recurrence checked 200 small cases through
$T=40$, with maximum request error $1.34\times10^{-15}$ and all forced
endpoints exact.

\subsection{Expected-prefix upper bound}
\label{app:anytime}\label{app:sine-spec}
To apply \eqref{eq:rank-general-transfer}, we need to control both the miscoverage
slack $G_T$ and the cumulative squared request deviations
$\E\sum_{t\le T}(q_t-\alpha)^2$. Keeping $G_t$ inside $(0,L_t)$ controls
the first term. We bound the second through the cumulative relative
entropy $\mathcal K_T$, as for Counting.
We first analyze an auxiliary controller with a sine potential: its
addition formula makes the relative-entropy bound telescope across rounds.
We then transfer the bound to Algorithm~\ref{alg:sine}, which uses the
quadratic function $\Phi_t$ and does not evaluate sine.

Write $\beta=1-\alpha$. Replace $\Phi_t$ in
Algorithm~\ref{alg:sine} by the sine potential
\begin{align}
 L_t&=4+\lceil t^{1/3}\rceil,\qquad \omega_t=\pi/L_t,
 \qquad
 \kappa_t=\log\frac{\alpha\sin(\beta\omega_t)}
                         {\beta\sin(\alpha\omega_t)},\label{eq:sine-params}\\
 \phi_t(z)&=e^{\kappa_tz}\sin(\omega_tz)\one\{0<z<L_t\}.
 \label{eq:sine-potential}
\end{align}
The sine factor is positive inside $(0,L_t)$ and zero at its boundaries.
The exponential factor is chosen to make \eqref{eq:sine-eigen-main}
hold even when $\alpha\ne\beta$.
For $z=G_{t-1}$, the next slack is $z+\alpha$ if $E_t=0$ and
$z-\beta$ if $E_t=1$. Weighting these two destinations gives the request
\begin{equation}\label{eq:sine-request}
 Z_t(z)=\beta\phi_t(z+\alpha)+\alpha\phi_t(z-\beta),
 \qquad q_t=\frac{\alpha\phi_t(z-\beta)}{Z_t(z)}.
\end{equation}

\paragraph{Feasibility.}
At $t=1$, $q_1=0$ and the full-set map gives $G_1=\alpha\in(0,L_1)$.
Since $\phi_1(G_0)=0$, the logarithmic calculation starts at the next round.
Suppose $0<z=G_{t-1}<L_{t-1}\le L_t$. The next states $z+\alpha$
and $z-\beta$ are separated by one, and at least one lies in $(0,L_t)$:
if $z-\beta\le0$, then $0<z+\alpha\le1<L_t$. If $z+\alpha\ge L_t$,
then $0<L_t-1\le z-\beta<L_t$. Thus $Z_t(z)>0$. A destination outside
the interval has zero weight and is excluded by the corresponding
full-set or empty-set map.
Induction gives
\begin{equation}\label{eq:sine-safe}
 0<G_t<L_t\quad\text{on every stream and every coin realization},
\end{equation}
which implies $M_t\le\lfloor\alpha t\rfloor$.

\paragraph{Sine-potential bound.}\label{app:sine-proof}
We first bound the normalizing factor $Z_t(z)$ in
\eqref{eq:sine-request}. This will bound each round's relative entropy
by a change in log potential plus a term of order $L_t^{-2}$.
Let $\bar\phi_t(z)=e^{\kappa_tz}\sin(\omega_tz)$ without truncation,
and set $\chi(x)=\log(\sin x/x)$, with $\chi(0)=0$.
For $0\le\omega\le\pi/4$, write
$\kappa(\omega)=\chi(\beta\omega)-\chi(\alpha\omega)$, so
$\kappa_t=\kappa(\omega_t)$. With $\kappa=\kappa(\omega)$,
\[
 \beta e^{\alpha\kappa}\sin(\alpha\omega)
 =\alpha e^{-\beta\kappa}\sin(\beta\omega).
\]
This identity is the reason for the choice of $\kappa_t$:
expanding the two shifted sines cancels their cosine terms and gives
\begin{align}
 \beta\bar\phi_t(z+\alpha)+\alpha\bar\phi_t(z-\beta)
 &=\rho_t\bar\phi_t(z),\label{eq:sine-eigen-main}\\
 \log\rho_t&=\chi(\omega_t)-\alpha\chi(\alpha\omega_t)
                             -\beta\chi(\beta\omega_t).\nonumber
\end{align}
Thus the weighted average of the untruncated potential at the two next
states is a scalar multiple $\rho_t$ of its current value.
For $0\le x\le\pi/4$,
\begin{equation}\label{eq:sinc-bounds}
 -x\le\chi'(x)\le0,\quad -x^2/2\le\chi(x)\le0,
 \quad |\kappa(\omega)|\le\omega^2/2,
 \quad |\kappa'(\omega)|\le\omega.
\end{equation}
Indeed,
$-\chi'(x)=\int_0^x u\sin u\,du/(x\sin x)
\le x^2/(3\sin x)\le\pi x/6\le x$.
Integration proves the bound on $\chi$. Its difference formula and
$\alpha^2+\beta^2\le1$ prove the bounds on $\kappa$.
All angles in use satisfy these estimates because $L_t\ge5$.
Monotonicity and nonpositivity of $\chi$ imply
\begin{equation}\label{eq:rho-bound}
 0<\rho_t\le1,\qquad -\log\rho_t\le\omega_t^2/2.
\end{equation}
For $0<z<L_t$, a forbidden destination lies in $(-1,0]$ or
$[L_t,L_t+1)$, where $\bar\phi_t$ is nonpositive. Truncating to zero
therefore gives
\begin{equation}\label{eq:sine-normalizer}
 Z_t(z)\ge\rho_t\phi_t(z)>0.
\end{equation}

We now compare the controller's miscoverage law with independent
Bernoulli$(\alpha)$ indicators. The sine request also uses only past
miscoverage indicators, so the same rank-history and tower-property
argument gives $\Pp(E_t=1\mid E_1,\ldots,E_{t-1})=q_t$.
For $t\ge2$, every allowed move satisfies
\begin{align*}
 \log\frac{\Pp(E_t\mid E_1,\ldots,E_{t-1})}{P_\alpha(E_t)}
 &=\log\phi_t(G_t)-\log Z_t(G_{t-1})\\
 &\le\log\phi_t(G_t)-\log\phi_t(G_{t-1})-\log\rho_t.
\end{align*}
The factors $\alpha$ or $\beta$ cancel, and \eqref{eq:sine-normalizer}
gives the inequality. Zero-weight moves cannot occur, so the logarithms are finite.
The conditional mean of the log likelihood ratio is
$\kl(q_t\Vert\alpha)$. If the potential were fixed, its successive
log increments would telescope exactly. Since $L_t$ grows, we must also
bound the change from $\phi_t$ to $\phi_{t+1}$ at the same slack value.
For $0<z<L_t\le L_{t+1}$, monotonicity of
$\sin x/x$ on $(0,\pi)$ and \eqref{eq:sinc-bounds} give
\begin{align*}
 \log\frac{\sin(\omega_tz)}{\sin(\omega_{t+1}z)}
 &\le\log\frac{L_{t+1}}{L_t},\\
 |\kappa_t-\kappa_{t+1}|z
 &\le\frac{\omega_t^2-\omega_{t+1}^2}{2}\frac\pi{\omega_t}
 \le\pi(\omega_t-\omega_{t+1}).
\end{align*}
Hence
\begin{equation}\label{eq:potential-expansion}
 \log\phi_t(z)-\log\phi_{t+1}(z)
 \le\log\frac{L_{t+1}}{L_t}+\pi(\omega_t-\omega_{t+1}).
\end{equation}
For $T\ge2$, along every allowed path,
\begin{align*}
 \sum_{t=2}^T[\log\phi_t(G_t)-\log\phi_t(G_{t-1})]
 &=\log\phi_T(G_T)-\log\phi_2(\alpha)\\
 &\quad+\sum_{t=2}^{T-1}
       [\log\phi_t(G_t)-\log\phi_{t+1}(G_t)]\\
 &\le\frac{\pi^2}{2L_T}-\log\phi_2(\alpha)
        +\log\frac{L_T}{L_2}+\pi(\omega_2-\omega_T).
\end{align*}
For the terminal term, $\sin(\omega_TG_T)\le1$ and
$|\kappa_T|G_T\le\pi^2/(2L_T)$. The intermediate terms are bounded by
\eqref{eq:potential-expansion}.
The first round costs $\kl(0\Vert\alpha)=\log(1/\beta)$.
Set $C_\alpha=\log[3/(\alpha\beta)]+\pi^2$.
Using $\sin(\pi\alpha/6)\ge\alpha/3$ and \eqref{eq:rho-bound}
therefore gives
\begin{equation}\label{eq:sine-entropy}
 \mathcal K_T:=\E\sum_{t=1}^T\kl(q_t\Vert\alpha)
 \le C_\alpha+\log L_T+\frac{\pi^2}{2}\sum_{t=2}^T L_t^{-2}.
\end{equation}
This also holds for $T=1$. Since $\sum_{t=1}^T L_t^{-2}\le3T^{1/3}$,
Equation~\eqref{eq:rank-general-transfer} implies
\begin{equation}\label{eq:sine-regret}
 \Reg_T\le\lambda L_T+c_+\left(\frac{\mathcal K_T}{2}
                          +\frac{\harm_T}{4}\right)=O(T^{1/3}).
\end{equation}
The bound holds at every horizon for the same run, so
$\max_{t\le T}\Reg_t=O(T^{1/3})$.

\paragraph{Quadratic-potential bound.}\label{app:quadratic-proof}
We transfer the preceding bound to
$\Phi_t(z)=z(L_t-z)\one\{0<z<L_t\}$, the potential used in
Algorithm~\ref{alg:sine}. Write $q_t^{\Phi}$ and
$q_t^\phi$ for the requests obtained from $\Phi_t$ and $\phi_t$, respectively,
at the same miscoverage history. Their potentials need not have similar
values: only the relative weights of the two next slack values determine
the request. We bound the change in these relative weights and show that replacing
$\phi_t$ by $\Phi_t$ adds at most $9/(8L_t^2)$ per round to the
relative-entropy bound.

\begin{lemma}[Quadratic barrier]\label{lem:quadratic-upper}
The quadratic controller has $0<G_t<L_t$ on every stream. With
empirical ranks under the assumptions of Theorem~\ref{thm:rank},
\begin{equation}\label{eq:quadratic-entropy}
 \mathcal K_T^{\Phi}:=\E\sum_{t=1}^T\kl(q_t^{\Phi}\Vert\alpha)
 \le C_\alpha+\log L_T+
 \left(\frac{\pi^2}{2}+\frac98\right)\sum_{t=2}^T L_t^{-2}.
\end{equation}
Consequently $\max_{t\le T}\Reg_t=O(T^{1/3})$.
\end{lemma}
\begin{proof}
The supports of $\Phi_t$ and $\phi_t$ agree, so the feasibility proof applies.
Let $Q^{\Phi},Q^\phi$ be the laws of their miscoverage indicators through $T$.
Both requests are functions of past miscoverage indicators. Theorem~\ref{thm:rank}
and the tower property therefore realize their requested probabilities
conditional on this binary history. The two laws allow exactly the same
binary paths.

For $L=L_t$, $\kappa=\kappa_t$, and $u=z/L\in(0,1)$, set
$r(z)=z(L-z)/[e^{\kappa z}\sin(\pi z/L)]$. Then
using $\cot x:=\cos x/\sin x$ and
$\frac{d}{dx}\log\sin x=\cot x$ for $0<x<\pi$ gives
\[
 (\log r)'(z)=\frac1L\left\{\frac1u-\frac1{1-u}
                                  -\pi\cot(\pi u)\right\}-\kappa.
\]
The expression in braces has absolute value at most $2$. On
$0<u\le1/2$, the function $1/u-\pi\cot(\pi u)$ increases from $0$
to $2$, since its derivative is
$\pi^2/\sin^2(\pi u)-1/u^2\ge0$, while $1\le1/(1-u)\le2$.
The other half follows by antisymmetry. Equation~\eqref{eq:sinc-bounds}
and $L\ge5$ give $|(\log r)'|\le2/L+\pi^2/(2L^2)<3/L$.

When both next slack values lie inside $(0,L_t)$, they are one unit
apart. The derivative bound therefore controls the change in log odds
at the same miscoverage history:
\[
 \left|\log\frac{q_t^{\Phi}}{1-q_t^{\Phi}}
          -\log\frac{q_t^\phi}{1-q_t^\phi}\right|
 =|\log r(G_{t-1}-\beta)-\log r(G_{t-1}+\alpha)|
 \le3/L_t.
\]
In log-odds coordinates, Bernoulli relative entropy is a first-order
Taylor remainder of $a(x)=\log(1+e^x)$. Since $a''\le1/4$,
$\kl(q_t^{\Phi}\Vert q_t^\phi)\le9/(8L_t^2)$.
If a destination is forbidden, both requests are the same endpoint and
this relative entropy is zero. Both also request zero at $t=1$. The chain rule gives
\[
 \KL(Q^{\Phi}\Vert Q^\phi)\le\frac98\sum_{t=2}^T L_t^{-2}.
\]
To compare $Q^{\Phi}$ with $P_\alpha$, we also need the sine controller's bound
under $Q^{\Phi}$, rather than under $Q^\phi$. The calculation preceding
\eqref{eq:sine-entropy} holds for every allowed path, before taking any
expectation. Hence, for
every path $e$ in this common support,
\[
 \log\frac{Q^\phi(e)}{P_\alpha(e)}
 \le C_\alpha+\log L_T+\frac{\pi^2}{2}\sum_{t=2}^T L_t^{-2}.
\]
Averaging under $Q^{\Phi}$ and using
\[
 \KL(Q^{\Phi}\Vert P_\alpha)
 =\KL(Q^{\Phi}\Vert Q^\phi)
   +\E_{Q^{\Phi}}\log\frac{Q^\phi(e)}{P_\alpha(e)}
\]
proves \eqref{eq:quadratic-entropy}. All ratios are defined on the common
support. Finally, \eqref{eq:pinsker-binary} bounds the cumulative squared
request deviations by $\mathcal K_t^{\Phi}/2$. Together with $G_t<L_t$,
Equation~\eqref{eq:rank-general-transfer} gives \eqref{eq:sine-regret} with
$\mathcal K_t^{\Phi}$ in place of $\mathcal K_t$, and therefore the
expected-prefix bound for Algorithm~\ref{alg:sine}.
\end{proof}

\subsection{Retrospective upper bound and adaptive stopping}\label{app:stopping}
Use the realized-cost identity \eqref{eq:realized-cost-martingale}
and the variables $D_t,\xi_t,\mathsf Z_t$ defined in
Appendix~\ref{app:proof-tools}.
For Barrier and the sine variant in Appendix~\ref{app:sine-spec},
$G_t<L_t\le L_T$ and
$\sum_{u\le t}D_u\le\sum_{u\le T}D_u$. Martingale differences are uncorrelated, so
$\E\mathsf Z_T^2=\sum_{t\le T}\E\xi_t^2\le T(B+\lambda)^2$.
The $L^2$ martingale maximal inequality therefore gives
$\E\max_{t\le T}|\mathsf Z_t|\le2(\E\mathsf Z_T^2)^{1/2}$.
Taking expectations in \eqref{eq:realized-cost-martingale} at $T$ gives
$\E\sum_{u\le T}D_u=\Reg_T-\lambda\E G_T$. Hence
\begin{align*}
 \E\max_{1\le t\le T}R_t
 &\le\lambda L_T+\E\sum_{u\le T}D_u+\E\max_{t\le T}|\mathsf Z_t|\\
 &\le\lambda L_T+\Reg_T-\lambda\E G_T+2(B+\lambda)\sqrt T
 =O(\sqrt T),
\end{align*}
by the expected-prefix bound, proving the retrospective upper bound
for the same run.
With the deterministic benchmark $tj(\alpha)$, this criterion includes
oracle size fluctuations.
The oracle interval in Section~\ref{sec:framework} has constant size,
so the square-root lower bound also holds without these fluctuations.

\begin{proposition}\label{prop:stopped-cost}
Under the assumptions of Theorem~\ref{thm:rank}, Barrier and the sine variant each satisfy, for
every stopping time $1\le\tau\le T$ of $(\history_t)$,
\begin{equation}\label{eq:stopped-cost}
 \E[W_\tau-\tau j(\alpha)]
 \le\lambda L_T+\Reg_T-\lambda\E G_T=O(T^{1/3}).
\end{equation}
\end{proposition}
\begin{proof}
Since $\tau\le T$ and the increments are bounded, optional stopping gives
$\E\mathsf Z_\tau=0$.
In \eqref{eq:realized-cost-martingale}, use $G_\tau<L_T$ and
$\sum_{u\le\tau}D_u\le\sum_{u\le T}D_u$, then take expectations.
\end{proof}

\section{Proof of Theorem~\ref{thm:lower-bounds}}
\label{app:lower-proof}\label{app:lower}\label{app:binaryreduction}
\label{app:anytime-lower}\label{app:general-pathwise}
Throughout this section, write $v=\alpha(1-\alpha)$ and
$x_+=\max\{x,0\}$ for the positive part of a real number $x$.
The proof uses the common identities in Appendix~\ref{app:proof-tools}
and the two lemmas below. It applies to arbitrary feasible policies,
including policies that use the full history and know $D,T$.

\begin{lemma}[Bernoulli survival bound]\label{lem:survival}
Let $B_t$ be independent Bernoulli$(\alpha)$ variables with joint law
$P_\alpha$, let $\mathsf S_t=\sum_{u\le t}(B_u-\alpha)$,
and set $K_\alpha=1+5/v$. For every integer $T\ge1$,
\begin{equation}\label{eq:survival}
 P_\alpha(\mathsf S_t\le0\text{ for all }t\le T)
 \le\min\{1,K_\alpha/\sqrt T\}.
\end{equation}
\end{lemma}
\begin{proof}
For $a\ge1$, stop at
$\tau=\inf\{t\ge1:\mathsf S_t>0\text{ or }\mathsf S_t\le-a\}$.
The stopped process lies in $[-a-1,1]$, so
$v\E(\tau\wedge n)=\E\mathsf S_{\tau\wedge n}^2\le(a+1)^2$.
Stop $\mathsf S_t^2-vt$ at bounded $\tau\wedge n$ for this identity.
The stopped mean is also zero.
Monotone convergence gives $\E\tau<\infty$. Bounded
convergence of the stopped positions gives
$\E\mathsf S_\tau=0$ and $\E\mathsf S_\tau^2=v\E\tau$.
If $p_-$ is the lower-exit probability, then
$0\le1-(a+1)p_-$, whence $p_-\le1/a$ and
\[
 v\E\tau\le1+p_-(a+1)^2\le5a.
\]
A path staying nonpositive through $T$ either exits through the lower
boundary or has $\tau>T$. Its probability is at most
$1/a+5a/(vT)$. Set $a=\sqrt T$ to obtain \eqref{eq:survival}.
\end{proof}

\begin{lemma}[Squared deviations on a time window]\label{lem:window-adjustment}
For a feasible policy through $T$, let
$\bar q_t=\Pp(E_t=1\mid E_1,\ldots,E_{t-1})$ and $a=v/4$.
For integers $s\ge0$, $n\ge1$ with $s+n\le T$,
\begin{equation}\label{eq:window-main}
 \E\sum_{t=s+1}^{s+n}(\bar q_t-\alpha)^2
 \ge\frac{(a\sqrt n-\E G_s)_+^2}{n}.
\end{equation}
\end{lemma}
\begin{proof}
For $N\sim\operatorname{Binomial}(n,\alpha)$, $n\ge1$, we use
\begin{equation}\label{eq:binomial-positive}
 \E(N-n\alpha)_+\ge a\sqrt n.
\end{equation}
Indeed, $X=N-n\alpha$ satisfies
$\E X^2=nv$ and
$\E X^4=3n^2v^2+nv(1-6v)\le n^2v(1+3v)$.
H\"older's inequality and $\E X=0$ give
\[
 \E X_+=\frac12\E|X|
 \ge\frac{(\E X^2)^{3/2}}{2(\E X^4)^{1/2}}
 \ge\frac{v\sqrt n}{2\sqrt{1+3v}}\ge a\sqrt n.
\]

Realize the policy's miscoverage law using independent $U_t\sim\mathrm{Unif}[0,1]$:
$E_t=\one\{U_t\le\bar q_t(E_1,\ldots,E_{t-1})\}$, and couple it with
$B_t=\one\{U_t\le\alpha\}$.
This preserves the policy's miscoverage law without assuming independent miscoverage indicators.
For a window $s+1,\ldots,s+n$ with $s+n\le T$, let
\[
 N=\sum_{t=s+1}^{s+n}B_t,\qquad
 D_{s,n}=\sum_{t=s+1}^{s+n}\one\{\bar q_t<U_t\le\alpha\}.
\]
The coupling and feasibility give
$N-(M_{s+n}-M_s)\le D_{s,n}$ and
$M_{s+n}-M_s\le\alpha n+G_s$. Since $D_{s,n}+G_s\ge0$,
\[
 a\sqrt n\le\E(N-\alpha n)_+
 \le\E G_s+\E D_{s,n}
 =\E G_s+\E\sum_{t=s+1}^{s+n}(\alpha-\bar q_t)_+.
\]
Cauchy--Schwarz gives
\[
 \E\sum_{t=s+1}^{s+n}(\alpha-\bar q_t)_+
 \le\left\{n\E\sum_{t=s+1}^{s+n}(\bar q_t-\alpha)^2\right\}^{1/2}.
\]
Rearrange and square when $a\sqrt n>\E G_s$. Otherwise the lower bound
in \eqref{eq:window-main} is zero.
\end{proof}

\begin{proof}[Proof of Theorem~\ref{thm:lower-bounds}]
Let $Q$ be the miscoverage-path law of any feasible policy and put
$\bar q_t=Q(E_t=1\mid E_1,\ldots,E_{t-1})$.
Conditioning on past miscoverages averages over the remaining history and
current randomization, leaving an admissible mixture independent of the fresh pair.
Thus, for every $s\le T$,
\begin{equation}\label{eq:reducedcost}
 \Reg_s\ge\lambda\E G_s+\E_Q\sum_{t=1}^s\psi(\bar q_t).
\end{equation}
This includes policies using the full history and knowing $D,T$.
Feasibility gives $G_s\ge0$. Since $j(1)=0$ and $j(\alpha)\ge0$,
$\psi(1)>0$ also implies $\lambda>0$.

\textbf{(i) Terminal lower bound.}
Let $\mathcal B_T$ be the binary paths satisfying every prefix quota.
Because counts are integer valued, this is precisely the event in
\eqref{eq:survival}. Feasibility gives $Q(\mathcal B_T)=1$.
The reference law $P_\alpha$ assigns positive mass to every binary path, so
\begin{align}\label{eq:KLlower}
 \E_Q\sum_{t=1}^T\kl(\bar q_t\Vert\alpha)
 &=\KL(Q\Vert P_\alpha)\nonumber\\
 &=\KL\bigl(Q\Vert P_\alpha(\cdot\mid\mathcal B_T)\bigr)
       +\log\frac1{P_\alpha(\mathcal B_T)}
 \ge\log\frac1{P_\alpha(\mathcal B_T)}.
\end{align}
Conditioning $P_\alpha$ on $\mathcal B_T$ divides feasible path probabilities
by $P_\alpha(\mathcal B_T)$. Relative entropy is nonnegative.
Apply the quadratic lower bound on $\psi$,
\eqref{eq:binary-kl-upper}, and $G_T\ge\alpha T-b_T$ in
\eqref{eq:reducedcost}. Together with Lemma~\ref{lem:survival}, these give
\begin{equation}\label{eq:quadratic-lower}
 \Reg_T\ge\lambda(\alpha T-b_T)+c_-v
       \left(\tfrac12\log T-\log K_\alpha\right)_+,
\end{equation}
which proves (i).

\textbf{(ii) Expected-prefix lower bound.}
Set $a=v/4$, $R=\max_{t\le T}\Reg_t\ge0$ and choose
\[
 n=\left\lceil\left(1+\frac{2R}{a\lambda}\right)^2\right\rceil,
 \qquad d=c_-a^2/4.
\]
Equation~\eqref{eq:reducedcost} gives
$\E G_s\le R/\lambda\le a\sqrt n/2$. The same holds at $s=0$ because
$G_0=0$. Lemma~\ref{lem:window-adjustment} gives at least $a^2/4$
in squared deviations per complete window. Summing disjoint windows gives
\[
 R\ge\Reg_T\ge d\lfloor T/n\rfloor,\qquad T<n(1+R/d).
\]
Expected squared deviations add across windows. Independence is unnecessary.
These inequalities also hold when $n>T$. With
$A=2\max\{1,2/(a\lambda)\}^2$, we have $n\le A(1+R)^2$, so
$T\le A(1+1/d)(1+R)^3$.
Thus (ii) holds with $c_2=[A(1+1/d)]^{-1/3}>0$.
This establishes the exponent, without certifying finite-horizon near-optimality.

\textbf{(iii) Retrospective-maximum lower bound.}
For this part assume only $\delta:=\psi(1)>0$.
Use $D_t,\xi_t,\mathsf Z_t$ from \eqref{eq:realized-cost-martingale}.
Since $j(1)=0$ and costs are nonnegative,
$\delta=\lambda(1-\alpha)-j(\alpha)\le\lambda$. Expanding $\xi_t$ gives
\[
 \xi_t=\ell_{C_t}(X_t)+\lambda E_t-
                \{j(\alpha)+\lambda\alpha+D_t\}.
\]
On a miscoverage, $\xi_t\ge\delta-D_t$. As $D_t$ is predictable,
\[
 \E[\xi_t^2\mid\history_{t-1}]
 \ge\frac{\delta^2}{4}p_t\one\{D_t\le\delta/2\}.
\]
Using $p_t\le1$, $\sum_t\E p_t=\alpha T-\E G_T$, and
$\Reg_T=\lambda\E G_T+\E\sum_tD_t$, we obtain
\begin{align*}
 \E\mathsf Z_T^2
 &\ge\frac{\delta^2}{4}\left\{\alpha T-\E G_T
                   -\frac2\delta\E\sum_{t\le T}D_t\right\}\\
 &\ge\frac{\delta^2}{4}
                   \left\{\alpha T-\frac2\delta\Reg_T\right\}.
\end{align*}
Martingale differences are orthogonal, so their second moments add.
Rounds with $D_t>\delta/2$ contribute at most
$(2/\delta)\E\sum_tD_t$ to the discarded miscoverage probabilities. The last
inequality uses $\delta\le\lambda$ and $G_T\ge0$.
If $\Reg_T\ge\alpha\delta T/4$, the expected maximum is at least
$(\alpha\delta/4)\sqrt T$. Otherwise,
$\E\mathsf Z_T^2\ge\alpha\delta^2T/8$.

Put $K=B+\lambda$. The bounded martingale differences satisfy
$|\xi_t|\le K$ and $\E\mathsf Z_{t-1}^2\le K^2(t-1)$.
Expanding the fourth power conditionally and using
$4K^3|x|\le2K^2x^2+2K^4$ gives
\begin{align*}
 \E\mathsf Z_t^4
 &\le\E\mathsf Z_{t-1}^4+6K^2\E\mathsf Z_{t-1}^2
                         +4K^3\E|\mathsf Z_{t-1}|+K^4\\
 &\le\E\mathsf Z_{t-1}^4+\{8(t-1)+3\}K^4.
\end{align*}
Summing yields $\E\mathsf Z_T^4\le4K^4T^2$. Hence
\[
 \E(\mathsf Z_T)_+=\frac12\E|\mathsf Z_T|
 \ge\frac{(\E\mathsf Z_T^2)^{3/2}}{2(\E\mathsf Z_T^4)^{1/2}}
 \ge\frac{\alpha^{3/2}\delta^3}{64\sqrt2K^2}\sqrt T.
\]
The decomposition gives $R_T\ge\mathsf Z_T$.
The first prediction is full under the all-stream quota, and mixing the
full-set and empty-set maps gives $j(\alpha)\le B(1-\alpha)$. Thus $R_1\ge0$ and
$\max_{1\le t\le T}R_t\ge(\mathsf Z_T)_+$ pathwise. We conclude that
\begin{equation}\label{eq:general-retrospective-constant}
 \E\max_{1\le t\le T}R_t\ge k\sqrt T,\qquad
 k=\min\left\{\frac{\alpha\delta}{4},
       \frac{\alpha^{3/2}\delta^3}{64\sqrt2(B+\lambda)^2}\right\}>0.
\end{equation}
Under the theorem's quadratic assumption,
$\delta\ge c_-(1-\alpha)^2>0$, completing (iii).
\end{proof}

\subsection{Terminal and expected-prefix tradeoff}
\label{app:tradeoff-proof}
\begin{corollary}[Terminal and expected-prefix tradeoff]\label{cor:terminal-prefix-tradeoff}
Under the assumptions of Theorem~\ref{thm:lower-bounds}, there is a
constant $c_4>0$, depending only on $\alpha,\lambda,c_-$, such that
every policy feasible through $T$ satisfies
\[
 (1+\Reg_T)\left(1+\max_{1\le t\le T}\Reg_t\right)^2\ge c_4T.
\]
Thus $\Reg_T=O(\log T)$ requires
$\max_{t\le T}\Reg_t=\Omega(\sqrt{T/\log T})$, while
$\max_{t\le T}\Reg_t=O(T^{1/3})$ requires $\Reg_T=\Omega(T^{1/3})$.
\end{corollary}
\begin{proof}
Write $P=\max_{1\le t\le T}\Reg_t\ge0$ and use the constants
$a=\alpha(1-\alpha)/4$, $d=c_-a^2/4$, and
$A=2\max\{1,2/(a\lambda)\}^2$ from part~(ii).
With $n=\lceil(1+2P/(a\lambda))^2\rceil$, the same window argument gives
$\Reg_T\ge d\lfloor T/n\rfloor$ and $n\le A(1+P)^2$.
Consequently
\[
 T<n(1+\Reg_T/d)
 \le A(1+P)^2(1+\Reg_T/d)
 \le A(1+1/d)(1+P)^2(1+\Reg_T).
\]
This proves the corollary with $c_4=[A(1+1/d)]^{-1}=c_2^3$.
\end{proof}
All constants are independent of the policy and $T$, so the two
asymptotic consequences also hold for policies chosen separately at each
horizon. This lower bound does not assert attainability of the entire
terminal--prefix tradeoff curve.

\section{Proof of Theorem~\ref{thm:optimal-prices}}
\label{app:optimal-proof}
The theorem combines the constructive upper bounds in
Appendix~\ref{app:rank-proof} with the lower bounds in
Appendix~\ref{app:lower-proof}.

\begin{proof}
The assumptions of Theorem~\ref{thm:optimal-prices} include those of
both Theorem~\ref{thm:rank} and Theorem~\ref{thm:lower-bounds}.
Counting implemented by empirical ranks is feasible on every stream,
ends with $M_T=b_T$, and has $\Reg_T=O(\log T)$
(Appendix~\ref{app:counting-proof}).
One infinite run of Barrier implemented by empirical ranks is feasible
on every stream and satisfies $\max_{t\le T}\Reg_t=O(T^{1/3})$
and $\E\max_{t\le T}R_t=O(\sqrt T)$
(Appendices~\ref{app:anytime} and~\ref{app:stopping}).
These give the upper bounds in the three infima.
Theorem~\ref{thm:lower-bounds} supplies matching lower bounds uniformly
over every policy feasible through $T$, even if it knows $D,T$.
Taking the respective infima proves all three $\Theta$ rates, with
constants depending only on $\alpha,B,\lambda,c_+,c_-$.
The implemented controllers use only $\alpha$, time, past scores and
miscoverage indicators, and fresh random seeds (plus $T$ for Counting).
\end{proof}

\paragraph{Adaptive-stopping consequence.}\label{app:stopping-optimal}
Let $\mathcal T_T$ consist of integer-valued stopping times of
$(\history_t)$ with $1\le\tau\le T$.
Under the assumptions of Theorem~\ref{thm:optimal-prices}, the optimal
price over feasible policies, which may know $D,T$, is
\begin{equation}\label{eq:stopping-minimax}
 \inf_{\mathcal A\ \mathrm{feasible\ through}\ T}
 \sup_{\tau\in\mathcal T_T}\E_D[W_\tau-\tau j(\alpha)]
 =\Theta(T^{1/3}).
\end{equation}
The lower bound follows from Theorem~\ref{thm:lower-bounds}(ii), since
deterministic times belong to $\mathcal T_T$.
Proposition~\ref{prop:stopped-cost} gives the uniform upper bound for
Algorithm~\ref{alg:sine}.

\section{Details of the geometric assumptions}\label{app:contact}
\subsection{Target efficiency and smoothness}
Target efficiency and local smoothness imply
Assumption~\ref{ass:contact}. A Taylor bound controls excess size near
$\alpha$, and bounded size extends this bound to all miscoverage levels.

Suppose threshold size is parametrized by miscoverage as
$h:[0,1]\to[0,B]$, including the full-set and empty-set maps at $q=0$ and $q=1$.
Since thresholds belong to $\cC$, $h(q)\ge j(q)$. If
$h(\alpha)=j(\alpha)$ and $h'$ is $L$-Lipschitz on
$[\alpha-\delta,\alpha+\delta]\subset(0,1)$ for some $\delta>0$, the
supporting line for $j$ gives
\[
 h(\alpha+u)-h(\alpha)
 \ge j(\alpha+u)-j(\alpha)\ge-\lambda u.
\]
Divide by $u>0$ and $u<0$ and take limits. Differentiability of $h$ at
$\alpha$ gives $h'(\alpha)=-\lambda$.
For $|q-\alpha|\le\delta$, the remainder is therefore
$h(q)-j(\alpha)+\lambda(q-\alpha)
=\int_\alpha^q[h'(u)-h'(\alpha)]\,du$.
The Lipschitz bound on the integrand yields
\[
 0\le h(q)-j(\alpha)+\lambda(q-\alpha)
 \le\frac L2(q-\alpha)^2\qquad(|q-\alpha|\le\delta).
\]
Outside this interval the excess above the supporting line is at most $B+\lambda$, because
$h(q)\le B$, $j(\alpha)\ge0$, and $|q-\alpha|\le1$.
There $(q-\alpha)^2\ge\delta^2$, so
Assumption~\ref{ass:contact} holds with
$c_+=\max\{L/2,(B+\lambda)/\delta^2\}$.
Conversely, the assumption at $q=\alpha$ requires $h(\alpha)=j(\alpha)$.
Smoothness alone does not imply this equality.

If the upper bound in Assumption~\ref{ass:contact} is relaxed to
$c_+(q-\alpha)^2+\Delta$ for a constant $\Delta\ge0$ and every threshold,
the proof of
Theorem~\ref{thm:rank} adds $T\Delta$ to the expected-size bound.
A fixed positive gap can dominate the sublinear control cost.
Without target efficiency, quota feasibility, the rank miscoverage law and variance bound, and
Theorem~\ref{thm:decomposition} remain valid under their respective
assumptions. The matching upper rates need not hold.
Taking $\cC$ to be a fixed score's threshold family restricts both the
comparator and admissible policies. Its frontier gives the lowest expected
size attainable by mixing thresholds, which can lie below the
threshold-size curve. If its value at $\alpha$ is attained
by mixing two thresholds with miscoverages $q_-<\alpha<q_+$, the frontier
is linear on $[q_-,q_+]$. Then $\psi$ vanishes on this interval, violating
strict quadratic lower growth. Restricting the comparison family alone therefore
does not preserve the three matching rates. The empirical gaps in
Table~\ref{tab:contact-gaps} concern this restricted comparison. They
neither establish efficiency at the target nor estimate a uniform $\Delta$.

\subsection{The continuous example}
We verify the frontier in Section~\ref{sec:framework} against all
measurable sets and their randomized mixtures.

The model has quantile function $Q$, cdf $F$, and density $f$ given by
\[
 Q(u)=\frac{3u+u^2}{4},\qquad
 F(y)=\frac{\sqrt{9+16y}-3}{2},\qquad
 f(y)=\frac4{\sqrt{9+16y}},\qquad 0\le y\le1.
\]
For a measurable set $A$ of length $v$, the sets
$A\setminus[0,v]$ and $[0,v]\setminus A$ have equal length.
Since $f$ decreases, the former has no greater probability than the latter.
Thus $\int_A f\le F(v)$: a set with miscoverage $q$ has length at least
$Q(1-q)$, attained by $[0,Q(1-q)]$.
For a randomized set $A$, apply this bound to each realization before
averaging. Convexity of $Q$ then gives
$\E\mu(A)\ge\E Q(1-m_D(A))\ge Q(1-\E m_D(A))$.
Thus thresholds attain $j$ as defined in Section~\ref{sec:setup}, giving
\[
 j(q)=1-\frac54q+\frac14q^2,\qquad
 \lambda=\frac54-\frac\alpha2,\qquad
 \psi(q)=\frac14(q-\alpha)^2.
\]
This verifies both quadratic conditions with $c_+=c_-=1/4$, including
the full-set and empty-set maps.

\subsection{The covariate-dependent example}\label{app:conditional-density}
The quadratic conditions also hold when optimal sets depend on the
covariate. The optimal prediction-set rule uses the same density cutoff
for both covariate values, giving different interval lengths.

Let $X$ be uniform on $\{0,1\}$ and let $Y\mid X=x$ have density
\begin{equation}\label{eq:conditional-densities}
 f_0(y)=\tfrac32-y,\qquad
 f_1(y)=\tfrac32-y+\frac{\sin(2\pi y)}{4\pi},\qquad 0\le y\le1.
\end{equation}
Both integrate to one and satisfy $1/2\le-f_x'\le3/2$.
Use $s(x,y)=3/2-f_x(y)$ and length cost, with $\cC$ containing all measurable
prediction-set rules. The frontier $j$ also allows randomization over these rules.

\begin{lemma}[Conditional-density thresholds]\label{lem:conditional-density}
For~\eqref{eq:conditional-densities}, the score thresholds attain the
randomized population frontier at every miscoverage. For $k\in[1/2,3/2]$, let
$r_x(k)$ be the unique solution of $f_x(r_x)=k$, let $F_x$ be the
conditional cdf, and set
\[
 q(k)=1-\tfrac12\{F_0(r_0(k))+F_1(r_1(k))\},\qquad
 A(k)=\tfrac12\sum_{x=0}^1\frac1{-f_x'(r_x(k))}.
\]
As $k$ increases from $1/2$ to $3/2$, $q(k)$ increases from $0$ to $1$.
The frontier and its derivatives are
\begin{equation}\label{eq:conditional-frontier}
 j(q(k))=\tfrac12\{r_0(k)+r_1(k)\},\qquad
 j'(q(k))=-\frac1k,\qquad
 j''(q(k))=\frac1{k^3A(k)}\in[4/27,12].
\end{equation}
Consequently Assumption~\ref{ass:contact} holds with $c_+=6$, and
$\psi(q)\ge(2/27)(q-\alpha)^2$ for every $q\in[0,1]$ and every
$\alpha\in(0,1)$. The score cdf is continuous.
\end{lemma}
\begin{proof}
Write $C^{(k)}:=C_{3/2-k}$ for the score-threshold map, so
$C^{(k)}(x)=\{y:f_x(y)\ge k\}=[0,r_x(k)]$.
For any $C\in\cC$, the integrand below is nonnegative at every
covariate and outcome, so
\[
 \E_X\int_0^1\bigl(\one\{y\in C^{(k)}(X)\}-\one\{y\in C(X)\}\bigr)(f_X(y)-k)\,dy\ge0.
\]
Average this inequality over $C\sim\pi$, with $\pi\in\Delta(\cC)$.
If $m_D(\pi)=m_D(C^{(k)})$, the density
terms cancel, leaving $V_D(\pi)\ge V_D(C^{(k)})$. Hence $C^{(k)}$ attains
the frontier even when randomization is allowed. The full-set and
empty-set maps attain the endpoints. Singleton boundaries have zero length
and probability.

Implicit differentiation gives $r_x'(k)=1/f_x'(r_x(k))$,
and $f_x(r_x(k))=k$ makes
$\frac{d}{dk}F_x(r_x(k))=k/f_x'(r_x(k))$.
Hence $q'(k)=kA(k)>0$ and
$\frac{d}{dk}V_D(C^{(k)})=-A(k)$. Dividing gives $j'=-1/k$, and
differentiating again gives $j''=1/(k^3A(k))$.
Since $2/3\le A(k)\le2$ and $1/2\le k\le3/2$, the stated bounds follow.
At $k=1/2$ both intervals are full, and at $k=3/2$ both have zero
length and probability. Thus $q(k)$ covers the full miscoverage range.
Write $k(q)$ for its inverse.
Taylor's formula around $\alpha$, with $\lambda=1/k(\alpha)$,
then bounds the remainder between
$(2/27)(q-\alpha)^2$ and $6(q-\alpha)^2$, including the endpoints
by continuity. These are the quadratic conditions in
Theorem~\ref{thm:optimal-prices}.
Finally, each $s(x,\cdot)$ is strictly increasing with derivative in
$[1/2,3/2]$ and the conditional response density is continuous. The
mixture score distribution is therefore continuous.
\end{proof}

\subsection{The flat-frontier example}\label{app:flat-frontier}
A linear frontier allows full-set and empty-set maps to use
each new quota unit immediately, keeping all three prices bounded.

Let $Y\sim\mathrm{Unif}[0,1]$, take length cost, and allow all measurable
sets and randomization over them. Coverage equals expected length, so $j(q)=1-q$,
$\lambda=1$, and $\psi\equiv0$. Issue the empty set when
$b_t-b_{t-1}=1$ and the full set otherwise. The miscoverage indicators are deterministic
on every stream, giving
\[
 M_t=b_t,\qquad W_t=t-b_t,\qquad
 R_t=\alpha t-\lfloor\alpha t\rfloor\in[0,1).
\]
All three criteria are bounded by one. Assumption~\ref{ass:contact} holds
with $c_+=0$, but positive quadratic lower growth fails.

\section{Experimental details}
\label{app:experiments}

\subsection{Continuous model and cost estimation}
\label{app:curvature-validation}
This model has explicit formulas for prediction-set size and miscoverage,
so we can separate coverage control from threshold estimation.

Draw $U\sim\mathrm{Unif}[0,1]$, set $Y=(3U+U^2)/4$, and use score
$s(y)=y$, intervals $[0,\tau]$, and interval length as set size.
The empty and full sets have size zero and one, respectively.
The density is $f(y)=4/\sqrt{9+16y}$, and
\[
j(q)=1-\tfrac54q+\tfrac14q^2,\qquad
 \lambda=\tfrac54-\tfrac\alpha2,\qquad
 \psi(q)=\tfrac14(q-\alpha)^2.
\]
Appendix~\ref{app:contact} verifies Assumption~\ref{ass:contact}.
We use targets $\alpha\in\{0.05,0.1,0.2\}$. Counting is recomputed at horizons
$160,320,640,1280,2560,5120,10240,16380$. These multiples of $20$
remove integer-rounding slack. Barrier and the sine variant use
$L_t=4+\lceil t^{1/3}\rceil$.
\emph{Barrier} is Algorithm~\ref{alg:sine}, with quadratic potential
$\Phi_t(z)=z(L_t-z)$. The \emph{sine variant} replaces only this potential
by~\eqref{eq:sine-potential} for a secondary comparison.
Both use Algorithm~\ref{alg:rank} unless the curve is labeled known $F$.
The \emph{width variant} uses the quadratic potential with the separately
validated schedule in Appendix~\ref{app:aci-horizons}. Learners receive
only past scores and miscoverage indicators. The known law is used for the oracle comparators,
evaluation, and the validation criteria below.

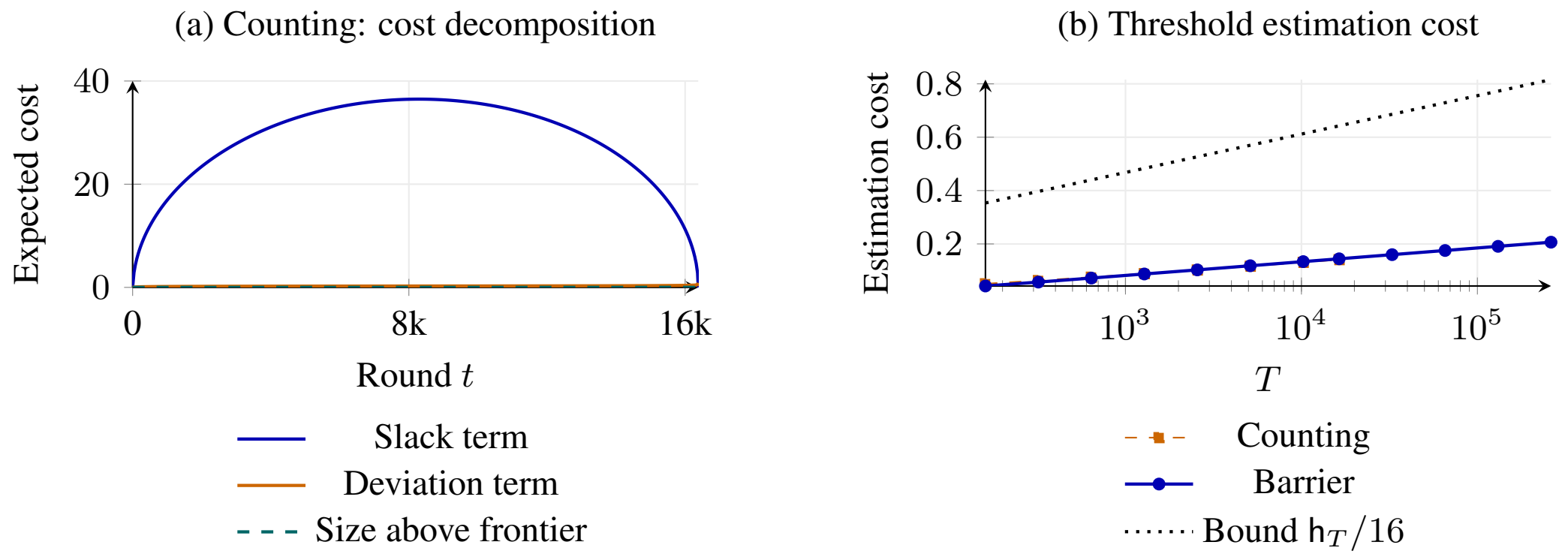
\begin{figure}[t]
\centering
\begin{minipage}[t]{.47\linewidth}\centering\vspace{0pt}
\begin{tikzpicture}
\begin{axis}[width=\linewidth,height=3.4cm,
 title={(a) Counting: cost decomposition},title style={font=\footnotesize},
 xlabel={Round $t$},ylabel={Expected cost},
 label style={font=\footnotesize},tick label style={font=\footnotesize},
 xmin=0,xmax=16380,ymin=0,ymax=40,scaled x ticks=false,xtick={0,8000,16000},
 xticklabels={0,8k,16k},ytick={0,20,40},
 grid=major,grid style={gray!15},legend style={font=\footnotesize,draw=none,
 at={(.5,0)},anchor=north,yshift=-30pt},axis lines=left]
\addplot[blue!70!black,thick] table[x=t,y=surplus]{plot_data_08.dat};
\addlegendentry{Slack term}
\addplot[orange!80!black,thick] table[x=t,y=curvature]{plot_data_08.dat};
\addlegendentry{Deviation term}
\addplot[teal!80!black,dashed,thick] table[x=t,y=selection]{plot_data_08.dat};
\addlegendentry{Size above frontier}
\end{axis}\end{tikzpicture}\end{minipage}\hfill
\begin{minipage}[t]{.47\linewidth}\centering\vspace{0pt}
\begin{tikzpicture}
\begin{axis}[width=\linewidth,height=3.4cm,
 title={(b) Threshold estimation cost},title style={font=\footnotesize},
 xlabel={$T$},ylabel={Estimation cost},xmode=log,
 label style={font=\footnotesize},tick label style={font=\footnotesize},
 grid=major,grid style={gray!15},legend style={font=\footnotesize,draw=none,
 at={(.5,0)},anchor=north,yshift=-30pt},axis lines=left]
\addplot[orange!80!black,dashed,mark=square*,mark size=1.2pt] table[x=T,y=rank_gap]{plot_data_12.dat};
\addlegendentry{Counting}
\addplot[blue!70!black,thick,mark=*,mark size=1.2pt] table[x=T,y=terminal_rank_gap]{plot_data_11.dat};
\addlegendentry{Barrier}
\addplot[black,dotted,thick] table[x=T,y=harmonic_bound]{plot_data_11.dat};
\addlegendentry{Bound $\harm_T/16$}
\end{axis}\end{tikzpicture}\end{minipage}
\caption{Direct checks of Theorems~\ref{thm:decomposition} and~\ref{thm:rank}
in the continuous model, $\alpha=0.1$.
(a) Counting with announced horizon $16{,}380$. Slack explains
its interior peak. The terminal terms are $0$, $0.8745$, and $0.0086$.
(b) Difference in expected terminal cost when each controller replaces
known-$F$ thresholds by Algorithm~\ref{alg:rank}. Counting uses each plotted
horizon as its endpoint. All values are numerical expectations, not
retrospective maxima. The bound is not fitted to these results.}
\label{fig:curvature-validation}
\end{figure}

\paragraph{Numerical expectations.}
We track the probability of each controller state from one round to the
next and compute expected sizes using uniform order-statistic moments.
For empirical ranks, let
$v_t:=\Var(m_D(C_t)\mid\rhist_{t-1})$ be the variance of the selected
map's miscoverage. Conditional on $\rhist_{t-1}$, this miscoverage has mean
$q_t$ for adjacent randomized ranks, and
\[
v_t=\frac{q_t(1-q_t)}{t+1}
       +\frac{\theta_t(1-\theta_t)}{t(t+1)},\qquad
 \theta_t=t(1-q_t)-\lfloor t(1-q_t)\rfloor.
\]
For a deterministic order index $J_t\in\{0,\ldots,t\}$, with indices
$0,t$ denoting the empty-set and full-set maps, $q_t=1-J_t/t$ and
$v_t=q_t(1-q_t)/(t+1)$. For population thresholds, use $v_t=0$,
since $m_D(C_t)=q_t$.
Thus
\[
 \Reg_T=\lambda\E G_T+\tfrac14\sum_{t\le T}
                  \E\{(q_t-\alpha)^2+v_t\}.
\]
The prefix objective maximizes this expectation over every integer prefix.
Figure~\ref{fig:curvature-validation}a fixes Counting's announced horizon
at $16{,}380$. Its term for size above the frontier satisfies
$\E[\epsilon_t\mid\rhist_{t-1}]=\theta_t(1-\theta_t)/[2t(t+1)]$.
Subtracting it from $\tfrac14\{(q_t-\alpha)^2+v_t\}$ gives the
deviation term in Theorem~\ref{thm:decomposition}.

\paragraph{Isolating threshold estimation.}\label{app:rank-cost}
For Counting and Barrier, empirical and known-$F$ thresholds have the
same law for $M_t$, expected slack, and squared request deviations.
Their difference in expected terminal cost is exactly
$\tfrac14\sum_{t\le T}\E v_t$, between zero and
$\harm_T/16$ by Theorem~\ref{thm:rank}. At $T=16{,}380$, this gives
$0.141045$ for Counting and $0.144278$ for Barrier, against the upper
bound $0.642566$. This comparison concerns expected terminal costs, not
retrospective maxima. At $T=262{,}144$, Barrier's estimation cost is
$0.206571$, or $0.47\%$ of its total expected cost.

\paragraph{Score-stream estimates.}
An independent implementation checks Counting on $512$ score streams
per target through $16{,}380$, using raw scores and independent rank coins.
Let $\tau_s$ be the selected threshold at round $s$. Empty-set and full-set maps
have miscoverage one and zero, respectively. At a fixed prefix $t$, we use
\[
 \widehat R_t=\lambda G_t+\tfrac14\sum_{s\le t}
                     (m_D(C_s)-\alpha)^2,\qquad
m_D(C_s)=1-F(\tau_s)
\]
for finite thresholds, with the stated endpoint conventions.
Its expectation equals that of $R_t=W_t-tj(\alpha)$ because
$R_t-\widehat R_t=\lambda\sum_{s\le t}(E_s-m_D(C_s))$ has mean zero.
Only evaluation uses $F$. At each fixed prefix, the estimator is unbiased,
but its variance need not be lower than that of $R_t$.\ Maxima of $\widehat R_t$ do not estimate maxima of $R_t$.
Pointwise Monte Carlo intervals are mean $\pm1.96$ standard errors (SEs)
over independent streams. Methods and checkpoints on the same streams
are correlated.

\paragraph{ACI and the quota guard.}\label{app:baseline-provenance}
Let $a_t$ be ACI's virtual target, $\gamma_t$ its step size, and
$E_t^{\rm virtual}$ its proposal's miscoverage indicator before the guard.
Starting from $a_1=\alpha$, ACI updates
$a_{t+1}=a_t+\gamma_t(\alpha-E_t^{\rm virtual})$ without projection
\citep{gibbs21}. For $0<a_t<1$, its proposal uses order index
$\lceil(t-1)(1-a_t)\rceil$ among past scores, with inclusive membership.
An empty history or $a_t\le0$ gives the full-set map, and $a_t\ge1$ gives the empty-set map.
The guard enforces the same prefix quota by issuing the full set when
$M_{t-1}+1>b_t$, and issuing the proposal otherwise.
After observing the outcome, we can determine the proposal's miscoverage indicator even
when the guard replaced it. The count $M_t$ includes only miscoverages of
the issued sets.
The population quota guard instead proposes the known-$F$ threshold at $\alpha$.

\subsection{The three temporal criteria}\label{app:additional-rates}
Expected-prefix cost takes the maximum of the expected cumulative excess size over
prefixes. Retrospective cost instead takes the largest realized excess
size on each stream and then averages across streams. We compute these
two criteria separately for the same controllers.

\begin{figure}[t]
\centering
\begin{minipage}[t]{.48\linewidth}\centering
\begin{tikzpicture}
\begin{axis}[width=\linewidth,height=3.45cm,xmode=log,xlabel={$T$},
 ylabel={$\max_{t\le T}\Reg_t/T^{1/3}$},label style={font=\small},tick label style={font=\small},
 grid=major,grid style={gray!15},axis lines=left,
 legend style={font=\scriptsize,draw=none,at={(.5,0)},anchor=north,yshift=-30pt}]
\addplot[blue!70!black,thick,mark=*] table[x=T,y=quad_cube]{plot_data_09.dat};\addlegendentry{Barrier}
\addplot[teal!80!black,dashed,mark=o] table[x=T,y=sine_cube]{plot_data_09.dat};\addlegendentry{Sine variant}
\end{axis}
\end{tikzpicture}
\end{minipage}\hfill
\begin{minipage}[t]{.48\linewidth}\centering
\begin{tikzpicture}
\begin{axis}[width=\linewidth,height=3.45cm,xmode=log,xlabel={$T$},
 ylabel={$\E\max_{t\le T}R_t/\sqrt T$},label style={font=\small},tick label style={font=\small},
 grid=major,grid style={gray!15},axis lines=left,
 legend style={font=\scriptsize,draw=none,at={(.5,0)},anchor=north,yshift=-30pt}]
\addplot[blue!70!black,thick,mark=*,error bars/.cd,y dir=both,y explicit]
 table[x=T,y=quad_sqrt,y error=sqrt95]{plot_data_09.dat};\addlegendentry{Barrier}
\addplot[teal!80!black,dashed,mark=o] table[x=T,y=sine_sqrt]{plot_data_09.dat};\addlegendentry{Sine variant}
\end{axis}
\end{tikzpicture}
\end{minipage}
\caption{Default Barrier and the sine variant, both using Algorithm~\ref{alg:rank}, at $\alpha=0.1$. The left panel uses numerical
expectations at every prefix. The right panel averages raw per-stream maxima.
Even after cube-root normalization, the curves still drift at these
horizons. Bars are pointwise approximate $95\%$ Monte Carlo intervals.}
\label{fig:normalized-rates}
\end{figure}
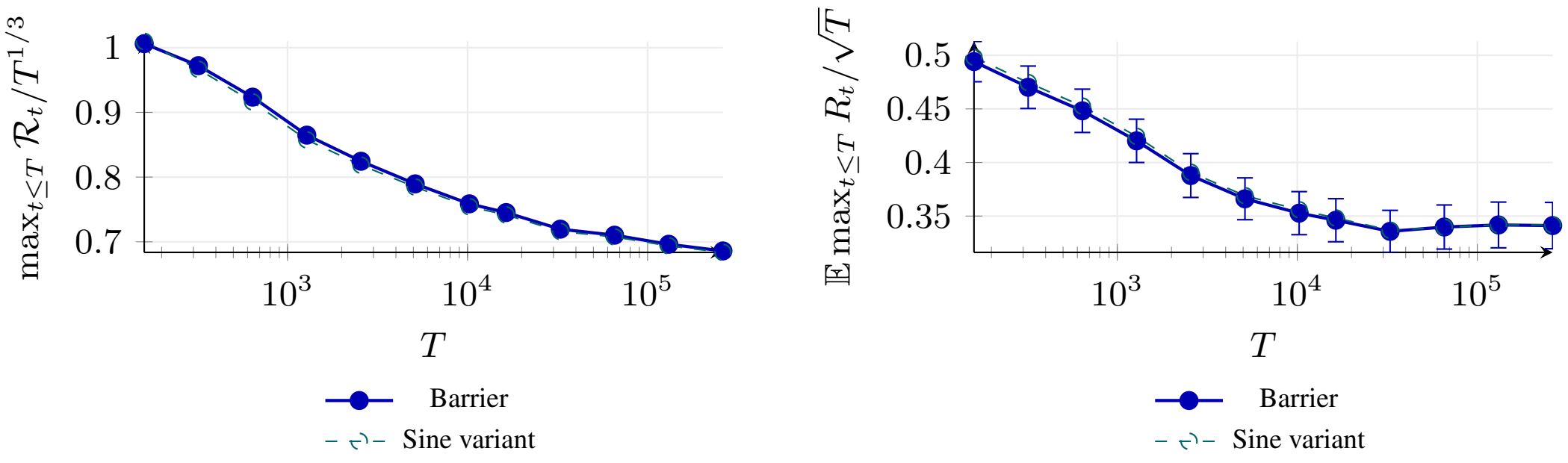

Figure~\ref{fig:three-criteria} uses the model and targets above.
The longer study extends Barrier, the sine variant, and the population quota guard to
$262{,}144$ rounds. Checkpoints are the Counting horizons above, followed
by $32{,}768$, $65{,}536$, $131{,}072$, and $262{,}144$.
The $512$ independent score streams per target are shared by Barrier
and the sine variant with population and empirical thresholds, and by
the population quota guard. Rank coins are independent.
Figure~\ref{fig:three-criteria} shows only the default Barrier.
Figure~\ref{fig:normalized-rates} also compares the sine variant. We chose this
extension after inspecting shorter runs.

Propagating controller-state probabilities computes $\max_{t\le T}\Reg_t$.
The retrospective estimate averages $\max_{1\le t\le T}R_t$ over simulated
score streams, using every prefix without clipping at zero. For the population
quota guard, an ordinary round has prediction-set size $j(\alpha)$ and a
forced full round has size one. Its realized excess cost is therefore
nondecreasing, so its terminal mean
estimates both prefix criteria. All five methods satisfy
every tested quota. Figure~\ref{fig:normalized-rates} divides these
costs by the theoretical orders. These normalized curves still vary with
$T$, so the simulations do not establish asymptotic scaling.

\begin{table}[!t]
\centering\small\setlength{\tabcolsep}{5pt}
\caption{Both prefix criteria at all targets, $T=262{,}144$. Expected-prefix maxima for Barrier and the sine variant are computed by state propagation. Quota-guard costs and retrospective maxima are raw trajectory estimates, mean $\pm$ Monte Carlo SE over $512$ streams. Barrier and the sine variant use Algorithm~\ref{alg:rank} at the default width. The quota guard knows the distribution.}
\label{tab:additional-rates}
\begin{tabular}{@{}llrr@{}}
\toprule
$\alpha$ & Method & $\max_{t\le T}\Reg_t$ & $\E\max_{t\le T}R_t$\\\midrule
0.05 & Barrier & 42.67 & $130.45\pm3.70$\\
0.05 & Sine variant & 42.56 & $131.35\pm3.68$\\
0.05 & Quota guard & $114.74\pm3.56$ & $114.74\pm3.56$\\
0.10 & Barrier & 43.90 & $174.65\pm5.64$\\
0.10 & Sine variant & 43.76 & $174.94\pm5.63$\\
0.10 & Quota guard & $159.55\pm5.51$ & $159.55\pm5.51$\\
0.20 & Barrier & 47.89 & $212.77\pm6.45$\\
0.20 & Sine variant & 47.75 & $212.64\pm6.49$\\
0.20 & Quota guard & $201.18\pm6.63$ & $201.18\pm6.63$\\
\bottomrule
\end{tabular}
\end{table}

\subsection{Numerical oracle benchmarks}
\label{app:prefix-optimum}
These numerical benchmarks know the distribution and horizon and may
choose any population threshold. Using double precision, we compute the
terminal optimum and lower and upper bounds on the expected-prefix optimum.
These computations compare performance at finite horizons. They are not
used in the analytic rate proofs in Appendices~\ref{app:rank-proof}--\ref{app:optimal-proof}.
The reported intervals do not account for
floating-point roundoff.

\paragraph{Terminal optimum.}\label{app:oracle-benchmark}
Let $\mathsf V_t(m)$ be the minimum expected remaining excess size after $t$
rounds and $m$ miscoverages, with $\mathsf V_T(m)=0$ at feasible terminal states.
The allowable requests are $\mathcal Q_t(m)=[0,1]$ if
$m+1\le\lfloor\alpha t\rfloor$, and $\mathcal Q_t(m)=\{0\}$ otherwise.
The terminal optimum in Figure~\ref{fig:three-criteria}a is $\mathsf V_0(0)$ from
\[
\mathsf V_{t-1}(m)=\min_{q\in\mathcal Q_t(m)}
 \{j(q)-j(\alpha)+(1-q)\mathsf V_t(m)+q\mathsf V_t(m+1)\}.
\]
The recursion adds this round's excess size to the average remaining
cost after the two possible outcomes. Starting from the terminal
condition, it solves earlier rounds in reverse order.
Infeasible branches are omitted. Write
$[x]_0^1=\min\{1,\max\{0,x\}\}$ for clipping to $[0,1]$.
When a miscoverage is permitted, the minimizer is
$[5/2-2\{\mathsf V_t(m+1)-\mathsf V_t(m)\}]_0^1$. Otherwise $q=0$.
We recompute it at each Counting horizon and target, without requiring
the final miscoverage count to fill the quota.

\paragraph{Prefix optimum.}
At $\alpha=0.1$, we bracket
$U_T^*=\inf_{\mathcal A\ \mathrm{feasible}}\max_{t\le T}\Reg_t$.
A weighted average of expected prefix costs cannot exceed their maximum,
so minimizing such an average gives a lower bound. The expected-prefix
cost of any feasible policy gives an upper bound.
For nonnegative weights $w_t$ summing to one,
$L(w):=\inf_{\mathcal A\ \mathrm{feasible}}\sum_t w_t\Reg_t\le U_T^*$.
Write $\beta_s=\sum_{t=s}^T w_t$. The same recursion computes $L(w)$
with stage cost $\beta_s\{j(q)-j(\alpha)\}$. When $\beta_s>0$ and
a miscoverage is permitted, the minimizer is
$[5/2-2\{\mathsf V_s(m+1)-\mathsf V_s(m)\}/\beta_s]_0^1$.
Take $q=0$ when miscoverage is forbidden. Zero-weight stages have
linear objectives. Convexity lets us average actions at the same round
and miscoverage count without increasing cost, so the recursion covers all
feasible history-dependent policies.

Each recursion returns a feasible policy $\mathcal A^{(i)}$. Let
$r_t^{(i)}:=\Reg_t(\mathcal A^{(i)},D)$ be its expected excess size at
prefix $t$, and let $z_i$ denote the probability of selecting this policy.
A linear program minimizes
$\max_t\sum_i z_i r_t^{(i)}$ over nonnegative $z_i$ summing to one.
Selecting one policy before deployment preserves every quota and gives
an upper bound. The linear program also identifies time weights for the
next recursion. Where $L$ is differentiable, we refine positive time weights using
$\partial L/\partial w_t=\Reg_t$.
Every integer prefix is a constraint. At $T=16{,}380$, the lower
and upper values are $5.654720$ and $5.659844$, a $0.091\%$ gap.

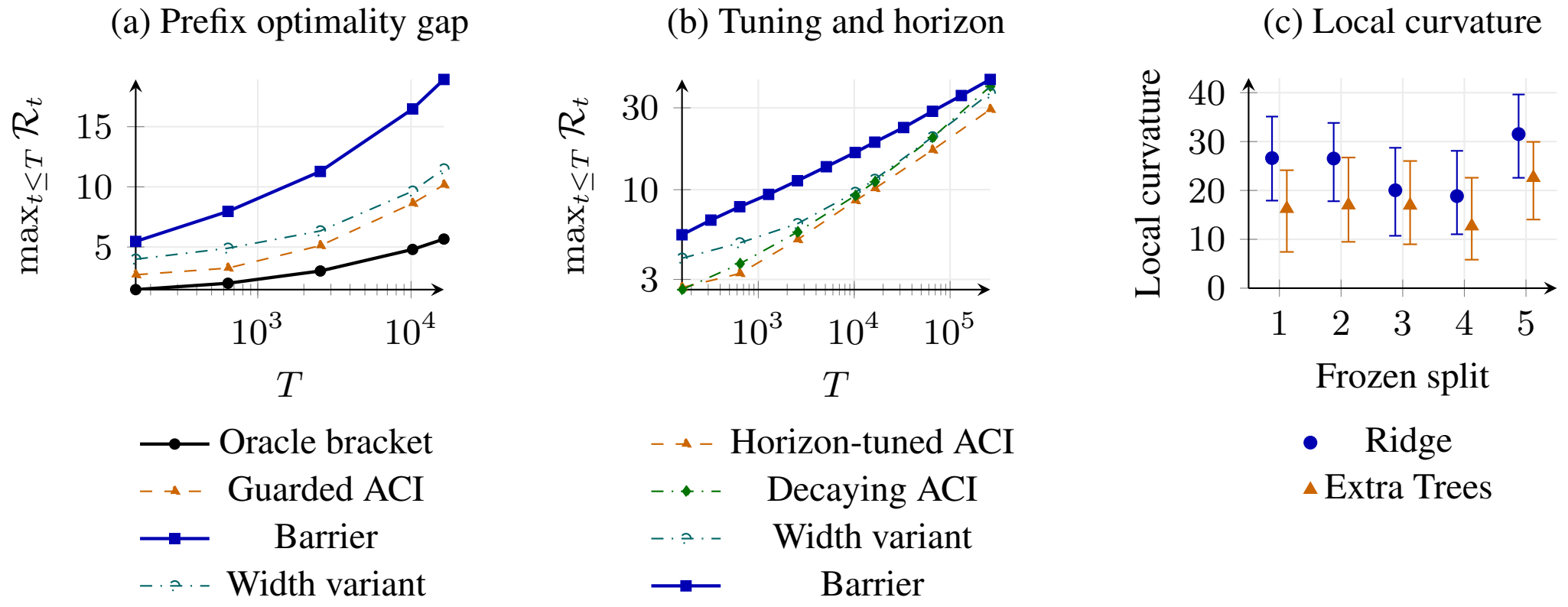
\begin{figure}[t]
\centering
\begin{minipage}[t]{.32\linewidth}\vspace{0pt}\centering
\begin{tikzpicture}
\begin{axis}[width=.95\linewidth,height=3.4cm,title={(a) Prefix optimality gap},
 title style={font=\footnotesize},xlabel={$T$},ylabel={$\max_{t\le T}\Reg_t$},
 label style={font=\footnotesize},tick label style={font=\footnotesize},xmode=log,
 grid=major,grid style={gray!15},axis lines=left,
 legend style={font=\footnotesize,draw=none,at={(.5,0)},anchor=north,yshift=-30pt}]
\addplot[black,thick,mark=*,mark size=1pt] table[x=T,y=upper]{plot_data_10.dat};
\addlegendentry{Oracle bracket}
\addplot[black,densely dashed,forget plot] table[x=T,y=lower]{plot_data_10.dat};
\addplot[orange!80!black,dashed,mark=triangle*,mark size=1.3pt] table[x=T,y=guarded_aci]{plot_data_10.dat};
\addlegendentry{Guarded ACI}
\addplot[blue!70!black,thick,mark=square*,mark size=1pt] table[x=T,y=barrier_default]{plot_data_10.dat};
\addlegendentry{Barrier}
\addplot[teal!80!black,dashdotted,mark=o,mark size=1.2pt] table[x=T,y=barrier_tuned]{plot_data_10.dat};
\addlegendentry{Width variant}
\end{axis}
\end{tikzpicture}
\end{minipage}\hfill
\begin{minipage}[t]{.32\linewidth}\vspace{0pt}\centering
\begin{tikzpicture}
\begin{axis}[width=.95\linewidth,height=3.4cm,title={(b) Tuning and horizon},
 title style={font=\footnotesize},xlabel={$T$},ylabel={$\max_{t\le T}\Reg_t$},
 label style={font=\footnotesize},tick label style={font=\footnotesize},xmode=log,ymode=log,
 ytick={3,10,30,100},yticklabels={$3$,$10$,$30$,$100$},
 grid=major,grid style={gray!15},axis lines=left,
 legend style={font=\footnotesize,draw=none,at={(.5,0)},anchor=north,yshift=-30pt}]
\addplot[orange!80!black,dashed,mark=triangle*,mark size=1.3pt] table[x=T,y=selected]{plot_data_02.dat};
\addlegendentry{Horizon-tuned ACI}
\addplot[green!45!black,dashdotted,mark=diamond*,mark size=1.4pt] table[x=T,y=decay]{plot_data_01.dat};
\addlegendentry{Decaying ACI}
\addplot[teal!80!black,dashdotted,mark=o,mark size=1.2pt] table[x=T,y=quadratic]{plot_data_02.dat};
\addlegendentry{Width variant}
\addplot[blue!70!black,thick,mark=square*,mark size=1pt] table[x=T,y=expected_empirical]{plot_data_11.dat};
\addlegendentry{Barrier}

\end{axis}
\end{tikzpicture}
\end{minipage}\hfill
\begin{minipage}[t]{.32\linewidth}\vspace{0pt}\centering
\begin{tikzpicture}
\begin{axis}[width=.95\linewidth,height=3.4cm,title={(c) Local curvature},
 title style={font=\footnotesize},xlabel={Frozen split},ylabel={Local curvature},
 label style={font=\footnotesize},tick label style={font=\footnotesize},
 xmin=.5,xmax=5.5,xtick={1,2,3,4,5},ymin=0,ymax=43,
 grid=major,grid style={gray!15},axis lines=left,
 legend style={font=\footnotesize,draw=none,at={(.5,0)},anchor=north,yshift=-30pt}]
\addplot[blue!70!black,only marks,mark=*,mark size=1.5pt,error bars/.cd,y dir=both,y explicit]
 table[x expr=\thisrow{split}-.12,y=linear,y error minus=linear_lo,y error plus=linear_hi]{plot_data_03.dat};
\addlegendentry{Ridge}
\addplot[orange!80!black,only marks,mark=triangle*,mark size=1.7pt,error bars/.cd,y dir=both,y explicit]
 table[x expr=\thisrow{split}+.12,y=trees,y error minus=trees_lo,y error plus=trees_hi]{plot_data_03.dat};
\addlegendentry{Extra Trees}
\end{axis}
\end{tikzpicture}
\end{minipage}
\caption{Finite-horizon comparisons at $\alpha=0.1$.
Barrier uses Algorithm~\ref{alg:sine}'s default width. Width variant uses
$L_t=\max\{5,4+\lceil t^{1/3}/4\rceil\}$. Both use Algorithm~\ref{alg:rank}.
(a) Expected-prefix costs and numerical optimum bracket. (b) All ACI
curves enforce the same quota. Fixed steps are selected by horizon.
The decay multiplier and variant width are selected once. Decaying ACI
uses 2,048 streams (SEs in Table~\ref{tab:decaying-aci}). Other costs are
numerical expectations. (c) Local fits on California Housing, bandwidth
$0.04$, with conditional $95\%$ paired-row bootstrap intervals (400 resamples).
Positive fitted curvature does not establish population target efficiency.}
\label{fig:followup}
\end{figure}

An independent linear program on a $0.05$ probability
grid at $T\in\{10,20,40,80\}$ lies within the bracket plus
$T(0.05)^2/16$. Mixing adjacent grid probabilities preserves transitions
and adds at most $(0.05)^2/16$ per round. Recovered policies agree with
their values from the recursion within $6\times10^{-11}$.

\subsubsection{Cost decomposition and strictly positive slack}
\label{app:policy-diagnostics}
We decompose expected cumulative excess size into the slack and deviation terms,
then measure the extra cost of requiring strictly positive slack.

\begin{figure}[t]
\centering
\begin{minipage}[t]{.32\linewidth}\centering
\begin{tikzpicture}
\begin{axis}[width=.92\linewidth,height=3.5cm,title={(a) Slack cost},
 title style={font=\footnotesize},xlabel={Round $t$},ylabel={$\lambda\E G_t$},
 label style={font=\footnotesize},tick label style={font=\footnotesize},
 xmin=0,xmax=16380,ymin=0,scaled x ticks=false,xtick={0,8000,16000},xticklabels={0,8k,16k},
 grid=major,grid style={gray!15},axis lines=left,
 legend style={font=\footnotesize,draw=none,at={(.5,0)},anchor=north,yshift=-30pt}]
\addplot[black,thick] table[x=t,y=mix_slack]{plot_data_07.dat};
\addlegendentry{Feasible mixture}
\addplot[blue!70!black,dashed,thick] table[x=t,y=barrier_slack]{plot_data_07.dat};
\addlegendentry{Width variant}
\end{axis}\end{tikzpicture}\end{minipage}\hfill
\begin{minipage}[t]{.32\linewidth}\centering
\begin{tikzpicture}
\begin{axis}[width=.92\linewidth,height=3.5cm,title={(b) Deviations from $\alpha$},
 title style={font=\footnotesize},xlabel={Round $t$},ylabel={$\frac14\E\sum_{s\le t}(q_s-\alpha)^2$},
 label style={font=\footnotesize},tick label style={font=\footnotesize},
 xmin=0,xmax=16380,ymin=0,scaled x ticks=false,xtick={0,8000,16000},xticklabels={0,8k,16k},
 grid=major,grid style={gray!15},axis lines=left,
 legend style={font=\footnotesize,draw=none,at={(.5,0)},anchor=north,yshift=-30pt}]
\addplot[black,thick] table[x=t,y=mix_adjustment]{plot_data_07.dat};
\addlegendentry{Feasible mixture}
\addplot[blue!70!black,dashed,thick] table[x=t,y=barrier_adjustment]{plot_data_07.dat};
\addlegendentry{Width variant}
\end{axis}\end{tikzpicture}\end{minipage}\hfill
\begin{minipage}[t]{.32\linewidth}\centering
\begin{tikzpicture}
\begin{axis}[width=.92\linewidth,height=3.5cm,title={(c) Zero slack},
 title style={font=\footnotesize},xlabel={Round $t$},ylabel={Mean $\Pp(G_t=0)$},
 ytick={0,.05,.1},yticklabels={0,0.05,0.10},
 label style={font=\footnotesize},tick label style={font=\footnotesize},
 xmin=0,xmax=16380,ymin=0,scaled x ticks=false,xtick={0,8000,16000},xticklabels={0,8k,16k},
 grid=major,grid style={gray!15},axis lines=left,
 legend style={font=\footnotesize,draw=none,at={(.5,0)},anchor=north,yshift=-30pt}]
\addplot[black,thick,mark=*,mark size=1pt] table[x=t,y=mixture]{plot_data_06.dat};
\addlegendentry{Feasible mixture}
\addplot[blue!70!black,dashed,thick] table[x=t,y=barrier]{plot_data_06.dat};
\addlegendentry{Width variant}
\end{axis}\end{tikzpicture}\end{minipage}
\caption{Structure of the recovered near-optimal prefix policy at
$\alpha=0.1,T=16{,}380$, with population thresholds. (a)--(b) The sum
is expected excess size. For readability, plotted checkpoints are quota
increments. Numerical maxima use all integer prefixes. (c) Mean zero-slack
probability over eligible prefixes in successive 1,000-round blocks
(the final block has 380 rounds). Terminal probability is one for the
mixture and zero for the width variant, which uses $L_t=\max\{5,4+\lceil t^{1/3}/4\rceil\}$. These are propagated expectations.}
\label{fig:policy-diagnostics}
\end{figure}
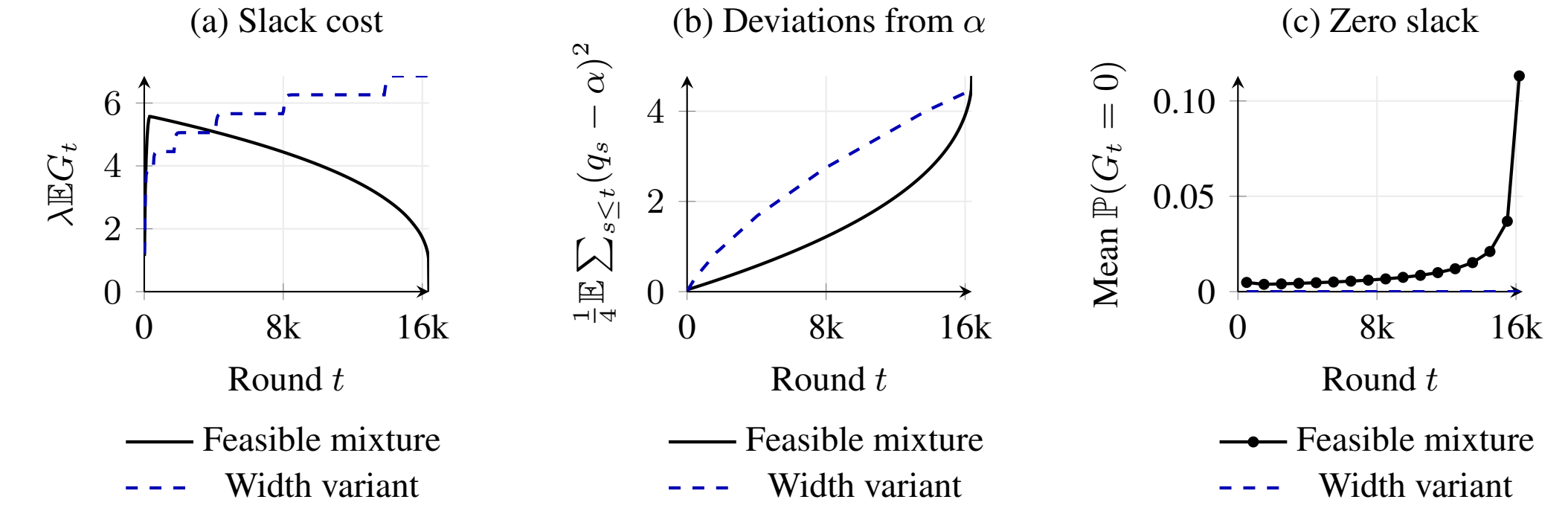

Figure~\ref{fig:policy-diagnostics} propagates the $25$ positive-weight
policies in the saved mixture at $\alpha=0.1,T=16{,}380$ and compares
them with the width variant at $L_t=\max\{5,4+\lceil t^{1/3}/4\rceil\}$.
Both use population thresholds. Theorem~\ref{thm:decomposition} gives
\[
 \Reg_t=\lambda\E G_t+\tfrac14\E\sum_{s\le t}(q_s-\alpha)^2.
\]
Both terms are computed at every integer prefix and averaged over the
initial mixture index. Their independently computed sum agrees with
cost to below $10^{-8}$.
This checks the numerical computation. The identity itself follows
from Theorem~\ref{thm:decomposition}.

Let $U_T^{>,*}$ be the expected-prefix optimum with $G_t>0$ at every round.
At $\alpha=0.1$,
this requires $M_t\le\lfloor(t-1)/10\rfloor$ and a full first prediction.
Deleting that round leaves an original-quota policy of length $T-1$.
Prepending a full prediction gives the converse. In the known-law
i.i.d.\ model the discarded observation supplies only independent
randomness, already permitted by the comparator. Nonnegative expected
excess at each feasible prefix then gives
\begin{equation}\label{eq:strict-prefix-identity}
U_T^{>,*}=B-j(\alpha)+U_{T-1}^*.
\end{equation}
At this target, requiring positive slack is equivalent to one initial
full prediction followed by the original problem with one fewer round.
For $T-1$, truncating and renormalizing the saved time weights gives
a lower bound, and truncating the original mixture gives an upper bound.
Consequently,
\[
U_T^{>,*}\in[5.777220,5.782344],\qquad
U_T^{>,*}-U_T^*\in[0.117376,0.127624].
\]
The restriction therefore explains at most
$0.127624/(11.336362-5.659844)=2.25\%$ of the width variant's gap.
Optimizing over all policies with strictly positive slack does not separate the
remaining effects of horizon knowledge, width schedule, and potential shape.

\subsection{Guarded ACI and controller selection}
\label{app:aci-horizons}
We compare step-size and width choices in the continuous model at
$\alpha=0.1$, with the quota guard above. Fixed ACI steps are selected at
each horizon from $\{0.001,0.005,0.02,0.08\}$ using $64$ independent
validation rank streams. The criterion is the maximum, over every prefix,
of the validation-stream mean of
\[
 \lambda G_t+\tfrac14\sum_{s\le t}\{(q_s-\alpha)^2+v_s\}.
\]
Uniform insertion ranks and independent rounding coins reproduce the
rank-history law. Choices are frozen before independent evaluation.
Maximizing after averaging distinguishes this criterion from retrospective
cost. Selecting at each horizon does not give one horizon-free controller.

The \emph{width variant}, $L_t=\max\{5,4+\lceil t^{1/3}/4\rceil\}$ in
Figure~\ref{fig:followup}, was selected at $T=16{,}380$ in a separate
$64$-stream validation study. For offsets $b\in\{0,4\}$ and multipliers
$c\in\{1/8,1/4,1/2,1,2\}$, candidates were
$L_t=\max\{5,b+\lceil ct^{1/3}\rceil\}$. Both the terminal and expected-prefix
validation objectives selected $(b,c)=(4,1/4)$. We chose the grid
after inspecting the default Barrier. The selected width is kept at all
horizons and evaluated on $256$ independent paired test streams.

To compute fixed-step ACI expectations, we track the joint distribution of
the proposal miscoverage count $N_t:=\sum_{u\le t}E_u^{\rm virtual}$ and
the issued-set miscoverage count $M_t$.
The guard gives
\[
N_t-M_t=\max\{N_{t-1}-M_{t-1},N_t-b_t\},\qquad N_0=M_0=0.
\]
The empirical order index depends on $(t,N_{t-1})$, while $M_{t-1}$
determines whether the guard issues the full set. Order-statistic moments
then give the expected issued size. States with probability below
$10^{-22}$ are omitted in the longer runs.
The recorded lost mass $\delta$ gives cost error at most $2T\delta$,
below $7.5\times10^{-11}$ in every reported calculation.

\paragraph{Decaying step.}\label{app:decaying-aci}
For the same unprojected ACI update, let $c$ be the step multiplier in
$\gamma_t=ct^{-1/3}$
\citep{angelopoulos2024decaying}. At the single validation horizon
$16{,}380$, $256$ independent rank streams select $c=0.4$ from
$\{0.025,0.05,0.1,0.2,0.4,0.8,1.6\}$ by the same expected-prefix
criterion. We freeze it for $2{,}048$ new paired test streams through
$262{,}144$, without retuning at checkpoints.
The prefix cost is estimated by maximizing the test-stream mean over
every integer prefix. SEs use $400$ bootstrap resamples of $64$
independent batch-mean curves, with $32$ streams per batch. They measure
simulation variability without removing the upward bias from maximization.
This miscoverage-level update is an empirical comparator, not
an application of a rate theorem for threshold updates.

\begin{table}[!t]
\centering\small
\caption{One frozen decaying-step guarded ACI schedule, $\gamma_t=0.4t^{-1/3}$, selected on 256 validation streams at $T=16{,}380$. Decaying ACI uses 2,048 new test streams. Parentheses give batch-bootstrap Monte Carlo SEs of the expected-prefix cost estimate. Fixed-step ACI and the width variant use numerical expectations. The fixed step is selected separately by horizon.}\label{tab:decaying-aci}
\begin{tabular}{@{}rrrr@{}}
\toprule
$T$ & Decaying ACI & Selected fixed-step ACI & Width variant\\\midrule
160 & 2.617 (0.024) & 2.691 & 3.986 \\
640 & 3.699 (0.023) & 3.246 & 4.898 \\
2,560 & 5.664 (0.028) & 5.114 & 6.335 \\
10,240 & 9.228 (0.039) & 8.627 & 9.639 \\
16,380 & 11.116 (0.039) & 10.158 & 11.484 \\
65,536 & 20.100 (0.040) & 17.008 & 20.308 \\
262,144 & 39.977 (0.050) & 29.362 & 36.525 \\
\bottomrule
\end{tabular}
\end{table}

\subsection{California Housing}
\label{app:real-geometry}
We examine the relation between interval length and miscoverage, then
compare online interval lengths under the quota using the same frozen predictors.

The \href{https://scikit-learn.org/stable/datasets/real_world.html\#california-housing-dataset}{California Housing dataset}
has $20{,}640$ observations and eight features.
Seeds $71000,\ldots,71004$ give five frozen splits, each with $10{,}320$
training, $5{,}160$ validation, and $5{,}160$ test observations, disjoint
within a split. The two predictors are standardized Ridge with penalty
one and Extra Trees with $128$ trees and minimum leaf size five.
Standardization and fitting use training data only. The response is the
recorded capped house value divided by six, with imposed domain $[0,1]$.
Predictions are clipped to this domain. Models stay fixed during
diagnostics, controller selection, and online runs.

\paragraph{Empirical curves.}
For held-out response $Y_i$, prediction $c_i$, and score
$S_i=|Y_i-c_i|$, threshold $\tau\ge0$
issues $[c_i-\tau,c_i+\tau]\cap[0,1]$. On $n$ held-out rows we compute
\[
 \widehat V(C_\tau)=\frac1n\sum_{i=1}^n
 \{\min(\tau,c_i)+\min(\tau,1-c_i)\},\qquad
 \widehat m(C_\tau)=\frac1n\sum_{i=1}^n\one\{S_i>\tau\}.
\]
We retain every distinct observed score as a threshold, include the empty-set map,
remove larger-size points at identical miscoverage, and take the lower
convex envelope $\widehat j$. Write $\widehat h$ for linear interpolation of the
miscoverage--length points. The gap
$\widehat h(\alpha)-\widehat j(\alpha)$ measures empirical length
above the frontier. The full-set map remains available to the guard even when
dominated under the empirical law. This envelope compares only thresholds
of the frozen score and their mixtures, not all measurable prediction maps.

Curvature is the second derivative of a least-squares quadratic fitted
to $\widehat h$ at $81$ equally spaced levels around $\alpha$.
For nominal bandwidth $b\in\{0.02,0.04,0.06\}$, the fitting bandwidth
is $\min\{b,0.8\alpha\}$, with $b=0.04$ primary. We evaluate
$\alpha\in\{0.05,0.1,0.2\}$ on validation and test separately.
Four hundred paired-row bootstrap resamples recompute the curves and
envelopes. The percentile ranges condition on the fitted model, treating
held-out rows as sampling units. They omit training uncertainty and
geographic dependence. A positive local fit does not
establish population target efficiency or global quadratic growth, and the
empirical envelope itself is polygonal.

Table~\ref{tab:contact-gaps} reports the target gaps and bootstrap ranges.
At $\alpha=0.1$, gaps range from $0$ to $0.00293$ for Ridge and from
$0.000033$ to $0.00183$ for Extra Trees. The ranges describe sampling
variation in the curves and envelopes, not a calibrated test of target
efficiency. Estimating the envelope can shift the bootstrap distribution:
for Extra Trees split 2, the point estimate lies below the percentile
range. Small estimates and positive lower endpoints do not establish
whether target efficiency holds. These pointwise gaps cannot replace the uniform
$\Delta$ in Appendix~\ref{app:contact}.

\begin{table}[t]
\centering\small
\caption{Target-gap diagnostics on the five frozen test splits at $\alpha=0.1$.
Entries are in units of $10^{-3}$ normalized length. Ranges are the
2.5th and 97.5th percentiles of 400 paired-row bootstrap resamples,
conditional on each fitted predictor. They do not establish target efficiency for the population frontier.}
\label{tab:contact-gaps}
\begin{tabular}{@{}r rr rr@{}}
\toprule
& \multicolumn{2}{c}{Ridge} & \multicolumn{2}{c}{Extra Trees}\\
\cmidrule(lr){2-3}\cmidrule(lr){4-5}
Split & Gap & Percentile range & Gap & Percentile range\\\midrule
1 & 0.000 & [0.000, 5.657] & 1.728 & [0.063, 3.996] \\
2 & 1.457 & [0.000, 3.261] & 0.033 & [0.052, 3.916] \\
3 & 2.932 & [0.030, 6.563] & 1.832 & [0.017, 7.021] \\
4 & 0.904 & [0.094, 4.198] & 0.214 & [0.170, 4.804] \\
5 & 2.902 & [0.035, 5.313] & 0.492 & [0.005, 2.328] \\
\bottomrule
\end{tabular}
\end{table}

\paragraph{Online evaluation.}
Each run permutes all $5{,}160$ observations without replacement and
starts with an empty score history. For each model, split, and target,
$16$ validation permutations select the ACI step from
$\{0.001,0.005,0.02,0.08\}$ and the quadratic width multiplier $c$ from
$\{1/8,1/4,1/2,1\}$ by mean issued length. The latter uses
$L_t=\max\{5,4+\lceil ct^{1/3}\rceil\}$.
Choices are frozen for $32$ paired test permutations. Test observations
are excluded from fitting and selection. Zero-radius intervals are
zero-length singletons, distinct from empty outputs. Score ties are included.

Figure~\ref{fig:real-followup}b averages paired length differences within
each split. Its SE is their sample standard deviation over $32$ permutations
divided by $\sqrt{32}$. Percentage reductions average the five split-specific
relative reductions equally. Uncertainty concerns ordering and controller
randomness on fixed test sets. Repeated permutations
and overlapping training splits are not independent population replications.
Every tested prefix for the guarded methods and Barrier satisfies the quota.

\paragraph{Matching the rank implementation.}
Alongside original guarded ACI, a variant passes
its clipped request $q_t=\min\{1,\max\{0,a_t\}\}$ through
Algorithm~\ref{alg:rank}. Its virtual target $a_t$ remains unprojected.
The update uses the unguarded proposal's miscoverage indicator even when the guard issues
the full set. Both variants use the same frozen pools and paired permutations,
with independent rank coins. We test matched ACI at the original selected
step and at its own validation choice from the same grid, frozen before
testing. At $\alpha=0.1$, both select
$\gamma=0.005$ in all ten model/split pairs, so these comparisons coincide.
The fixed-step comparison changes both the quantile implementation and
the resulting virtual targets. Requests are not held fixed.

Table~\ref{tab:matched-rank-aci} shows that matching ranks slightly increases
ACI's mean length. Barrier's reduction relative to matched ACI averages
$1.68\%$ for Ridge and $2.21\%$ for Extra Trees across splits.
All methods have similar miscoverage, and Barrier's empty-output frequency
is below $0.003\%$. For each length difference, the pooled conditional SE
is the square root of the sum of five squared within-split SEs,
divided by five. Intervals are the mean $\pm1.96$ SE. They concern
randomization on the frozen data, not population performance.
The default reproduction run generates both step-selection comparisons
at all three targets and saves all validation and test runs.

\begin{table}[!t]
\centering\small
\setlength{\tabcolsep}{4pt}
\caption{California Housing at $\alpha=0.1$. Both ACI variants are quota-guarded. ACI + ranks uses Algorithm~\ref{alg:rank}, and Barrier uses its default width. All ten model/split pairs select $\gamma=.005$ for both ACI variants, so the fixed-step and separately tuned comparisons coincide. Entries average five frozen splits with 32 paired test permutations each. Length is normalized. Miscoverage, full, and empty frequencies are percentages. Differences subtract ACI + ranks, with conditional 95\% ordering intervals excluding training and population uncertainty. Every tested prefix is feasible.}
\label{tab:matched-rank-aci}
\begin{tabular}{@{}llrrrrr@{}}
\toprule
Predictor & Method & Length & $M_T/T$ & Full & Empty & Difference [95\% interval]\\\midrule
Ridge & Original ACI & 0.3644 & 9.84 & 1.68 & 0.0000 & $-0.0013\;[-0.0016, -0.0010]$ \\
Ridge & ACI + ranks & 0.3657 & 9.84 & 1.94 & 0.0000 & --- \\
Ridge & Barrier & 0.3595 & 9.79 & 0.46 & 0.0013 & $-0.0062\;[-0.0068, -0.0055]$ \\
Extra Trees & Original ACI & 0.2910 & 9.84 & 1.69 & 0.0000 & $-0.0011\;[-0.0015, -0.0007]$ \\
Extra Trees & ACI + ranks & 0.2921 & 9.84 & 1.89 & 0.0000 & --- \\
Extra Trees & Barrier & 0.2856 & 9.79 & 0.52 & 0.0024 & $-0.0064\;[-0.0071, -0.0058]$ \\
\bottomrule
\end{tabular}
\end{table}

\paragraph{Paired quota-guard ablation.}\label{app:guard-ablation}
\begin{table}[t]
\centering\small
\setlength{\tabcolsep}{5pt}
\caption{Paired quota-guard ablation on California Housing ($\alpha=0.1$, $T=5{,}160$). Each entry averages five frozen splits and 32 test permutations per split. ACI pairs use identical proposals and the previously selected $\gamma=0.005$. The $M_T/T$ column reports terminal miscoverage. Violations count runs with any infeasible prefix. Max excess is the largest $(M_t-b_t)_+$ over all tested rounds and runs.}
\label{tab:guard-ablation}
\begin{tabular}{@{}llrrrr@{}}
\toprule
Predictor & Method & Length & $M_T/T$ (\%) & Violations (\%) & Max excess\\\midrule
Ridge & ACI & 0.3542 & 10.00 & 100.0 & 12\\
Ridge & ACI + guard & 0.3644 & 9.84 & 0.0 & 0\\
Ridge & Barrier & 0.3595 & 9.79 & 0.0 & 0\\
Extra Trees & ACI & 0.2795 & 10.00 & 100.0 & 12\\
Extra Trees & ACI + guard & 0.2910 & 9.84 & 0.0 & 0\\
Extra Trees & Barrier & 0.2856 & 9.79 & 0.0 & 0\\
\bottomrule
\end{tabular}
\end{table}

We reuse the same five frozen splits, both predictors, and all $32$ test
permutations at $\alpha=0.1$. Each ACI variant uses its previously selected
$\gamma=0.005$. No model, parameter, or test ordering is reselected.
Because the guarded method updates its virtual state with the proposal's
miscoverage indicator, removing the guard leaves the proposal sequence unchanged.
We therefore record two issued streams for identical proposals, feedback,
and, for matched ACI, rounding coins. Only one stream replaces proposals
by the full set when another miscoverage would violate the quota.
Their cumulative length difference equals the sum of full-set enlargements
on guarded rounds, including zero increments when the proposal is already full.

\begin{table}[t]
\centering\small
\setlength{\tabcolsep}{4.9pt}
\caption{Full paired guard ablation at $\alpha=0.1$. The two ACI variants each use $\gamma=0.005$, fixed by prior validation. Peak excess means $\max_t(M_t-b_t)_+$. Its mean and maximum are taken across the 160 runs per predictor. Full and empty frequencies are percentages of rounds.}
\label{tab:guard-ablation-details}
\begin{tabular}{@{}llrrrrrr@{}}
\toprule
Predictor & Method & Length & $M_T/T$ (\%) & Mean peak & Max peak & Full (\%) & Empty (\%)\\\midrule
Ridge & ACI & 0.3542 & 10.00 & 8.16 & 12 & 0.02 & 0.0000\\
Ridge & ACI + guard & 0.3644 & 9.84 & 0.00 & 0 & 1.68 & 0.0000\\
Ridge & ACI + ranks & 0.3544 & 10.00 & 8.06 & 12 & 0.09 & 0.0016\\
Ridge & ACI + ranks + guard & 0.3657 & 9.84 & 0.00 & 0 & 1.94 & 0.0000\\
Ridge & Barrier & 0.3595 & 9.79 & 0.00 & 0 & 0.46 & 0.0013\\
Extra Trees & ACI & 0.2795 & 10.00 & 8.41 & 12 & 0.02 & 0.0000\\
Extra Trees & ACI + guard & 0.2910 & 9.84 & 0.00 & 0 & 1.69 & 0.0000\\
Extra Trees & ACI + ranks & 0.2797 & 10.00 & 8.22 & 12 & 0.09 & 0.0028\\
Extra Trees & ACI + ranks + guard & 0.2921 & 9.84 & 0.00 & 0 & 1.89 & 0.0000\\
Extra Trees & Barrier & 0.2856 & 9.79 & 0.00 & 0 & 0.52 & 0.0024\\
\bottomrule
\end{tabular}
\end{table}

For both variants and predictors, every unguarded run violates some prefix,
despite terminal miscoverage near $0.1$. Peak excess is
$\max_{t\le T}(M_t-b_t)_+$: its mean is about eight miscoverages and its maximum
is twelve. The guarded streams and Barrier have zero violations.
For matched ACI, the guard adds $0.01127$ normalized length for Ridge
and $0.01236$ for Extra Trees, with paired conditional $95\%$ intervals
$[0.01071,0.01182]$ and $[0.01177,0.01296]$.
For original ACI, the guard adds $0.01017$ and $0.01150$ normalized length,
with paired conditional $95\%$ intervals $[0.00962,0.01072]$ and
$[0.01087,0.01212]$, respectively (Table~\ref{tab:guard-ablation}).
The paired intervals use the same within-split ordering SEs and pooling
rule as above. These differences isolate the guard for the fixed
implementation, not the infimum over feasible policies defining the price.

All $4{,}160$ candidate/repetition rows for the original guarded
implementations reproduce exactly.
Independent sorted-list replay checks complete first-permutation traces
for Ridge split 1 and Extra Trees split 5, including the proposal and
issued-set miscoverage indicators, every prefix, and both guarded and unguarded lengths.

\paragraph{Empty outputs.}
Counting uses empty sets to complete its final count, and Barrier can use
them to reduce large slack. Our size objective assigns them zero cost.
Their low observed frequency does not establish rates under a useful-output
restriction. Such a restriction must specify the allowed family or a
minimum useful size. Nonempty singletons can still have zero length.

\FloatBarrier

\section{Comparison with related control formulations}\label{app:control-comparison}
Hard prefix constraints and cost--buffer tradeoffs appear in control and
resource allocation. Table~\ref{tab:control-comparison} compares the
information available at each decision, the comparator, and when cost is
evaluated.

Constrained online convex optimization also studies regret--violation
tradeoffs \citep{guo2022hard,yi2021regret,sinha2024optimal}.
These works bound cumulative violations. The ``hard'' metric in
\citet{guo2022hard} sums positive violations, so conservative rounds cannot
cancel them. This differs from satisfying the quota at every realized
prefix. We measure the cost relative to the population coverage--efficiency frontier,
even when the distribution is known.

\begin{table}[H]
\centering\small
\setlength{\tabcolsep}{4pt}
\renewcommand{\arraystretch}{1.12}
\caption{Closest control formulations and their evaluated objectives.}
\label{tab:control-comparison}
\begin{tabular}{@{}L{0.22\linewidth}L{0.35\linewidth}L{\dimexpr0.43\linewidth-4\tabcolsep\relax}@{}}
\toprule
Work & Constraint and decision information & Comparator and time objective\\
\midrule
\citet{berry-gallager}, Propositions 4.2, 4.5
& Nonnegative queue. Current channel and buffer observed. Stationary
policies under its regularity assumptions.
& Long-run average power above the unconstrained power--rate frontier, with an
inverse-square gap as average delay grows.\\[3pt]
\citet{chenwang2025degeneracy}, Theorem 3.1
& Fixed initial inventory. Current reward and demand observed.
Distribution and horizon known.
& Expected terminal reward gap to a hindsight linear program.
Rates depend on the request distribution.\\[3pt]
\citet{chen2026replenishment}, Theorems 5.3, 5.8
& Inventory nonnegative at every prefix. Current reward, demand and
replenishment observed. Unknown distribution.
& Expected terminal reward gap to a prefix-feasible hindsight integer
program. Logarithmic and square-root regimes on different instance classes.\\[3pt]
\citet{mcmahan2024anytime}, Theorem 3
& Almost-sure prefix budgets. Action precedes realized cost. Policies
may use cumulative cost history.
& Optimal expected total reward among feasible policies over a finite
horizon. State augmentation supports planning and learning.\\
\bottomrule
\end{tabular}
\end{table}

Here the prediction-set rule is selected before the fresh observation,
and the comparator is $tj(\alpha)$. Changing the time criterion under the same geometric
conditions yields the three orders in
Theorem~\ref{thm:optimal-prices}. Lemma~\ref{lem:window-adjustment} bounds
squared probability adjustments on each window for every feasible policy,
including those using the full history and knowing $D,T$. Summing over
disjoint windows gives the expected-prefix lower bound without requiring
stationary or independent actions. Empirical ranks implement the upper
bounds without knowing the score distribution.

\section{A fixed additional miscoverage allowance}\label{app:fixed-buffer-prefix}
A fixed allowance permits additional miscoverages while retaining the same
comparator and expected-prefix cube-root order.

For an integer $k\ge0$, let $\mathcal A_k(T)$ be the policies satisfying
$M_t\le b_t+k$ for every $t\le T$, on every stream and
every realization of their randomization. Retain
$\Reg_t=\E W_t-tj(\alpha)$ and $G_t=\alpha t-M_t$.

\begin{proposition}[Expected-prefix cost with a fixed allowance]
\label{prop:fixed-buffer-prefix}
Under the assumptions of Theorem~\ref{thm:lower-bounds}, every
$\mathcal A\in\mathcal A_k(T)$ satisfies
\[
 \max_{1\le t\le T}\Reg_t(\mathcal A,D)
 \ge c_2T^{1/3}-1-2\lambda k,
\]
where $c_2$ is the coefficient in the proof of
Theorem~\ref{thm:lower-bounds}(ii), independent of $k$.
Under the assumptions of Theorem~\ref{thm:optimal-prices}, for each fixed
integer $k\ge0$,
\[
 \inf_{\mathcal A\in\mathcal A_k(T)}
 \max_{1\le t\le T}\Reg_t(\mathcal A,D)
 =\Theta(T^{1/3})\qquad(T\longrightarrow\infty).
\]
\end{proposition}
\begin{proof}
Write $\bar q_t=\Pp(E_t=1\mid E_1,\ldots,E_{t-1})$.
Equation~\eqref{eq:reducedcost} still holds, and feasibility gives
$G_s+k\ge0$. Thus, for $1\le s\le T$,
\[
 \Reg_s+\lambda k
 \ge\lambda\E(G_s+k)
       +c_-\E\sum_{t\le s}(\bar q_t-\alpha)^2.
\]
In particular, $\max_{t\le T}\Reg_t\ge-\lambda k$. Put
$R=\max_{t\le T}\Reg_t+2\lambda k\ge\lambda k\ge0$.
Then $\E(G_s+k)\le R/\lambda$ for $0\le s\le T$, including
$s=0$ since $G_0+k=k$, and
$c_-\E\sum_{t\le T}(\bar q_t-\alpha)^2\le R$.

For a window $s+1,\ldots,s+n$ contained in $1,\ldots,T$,
feasibility gives $M_{s+n}-M_s\le\alpha n+G_s+k$.
The proof of Lemma~\ref{lem:window-adjustment}, with
$a=\alpha(1-\alpha)/4$, gives
\[
 \E\sum_{t=s+1}^{s+n}(\bar q_t-\alpha)^2
 \ge\frac{(a\sqrt n-\E(G_s+k))_+^2}{n}.
\]
Choose $n=\lceil(1+2R/(a\lambda))^2\rceil$ and
$d=c_-a^2/4$. Since $\E(G_s+k)\le a\sqrt n/2$, each complete window
contributes at least $a^2/4$ in expected squared deviations. Summing disjoint
windows yields $R\ge d\lfloor T/n\rfloor$ and
$T<n(1+R/d)$. These inequalities also hold when $n>T$.
With $A=2\max\{1,2/(a\lambda)\}^2$, we have
$n\le A(1+R)^2$, and therefore
$T\le A(1+1/d)(1+R)^3$. Thus
$R\ge c_2T^{1/3}-1$, with the same
$c_2=[A(1+1/d)]^{-1/3}$ as before.
Substituting the definition of $R$ proves the lower bound.

The original Barrier policy belongs to $\mathcal A_k(T)$ and gives
$\max_{t\le T}\Reg_t=O(T^{1/3})$ under the upper-bound assumptions.
For fixed $k$, the additive $2\lambda k+1$ in the lower bound is
eventually dominated by $c_2T^{1/3}$, proving the asserted order.
\end{proof}

Relative to $tj(\alpha)$, additional miscoverages may make excess size negative. The
decomposition gives only $\Reg_t\ge-\lambda k$. This result fixes $k$
before $T\to\infty$. Its finite-horizon lower bound is uniform in $k$,
which also gives the following consequence.

\begin{corollary}[Initial allowances of order $o(T^{1/3})$]\label{cor:small-growing-buffer}
Under the assumptions of Theorem~\ref{thm:optimal-prices}, let $k_T$ be
nonnegative integers with $k_T=o(T^{1/3})$. If the allowance $k_T$ is
available from the first round and is held fixed throughout the run, then
\[
 \inf_{\mathcal A\in\mathcal A_{k_T}(T)}
 \max_{1\le t\le T}\Reg_t(\mathcal A,D)=\Theta(T^{1/3}).
\]
\end{corollary}
\begin{proof}
Substitute $k=k_T$ into the uniform lower bound of
Proposition~\ref{prop:fixed-buffer-prefix}. Its subtractive term
$1+2\lambda k_T=o(T^{1/3})$ cannot change the order.
The original Barrier is feasible for every nonnegative allowance and
supplies the upper bound, with constants independent of $k_T$.
\end{proof}

For example, $k_T=\lfloor\kappa T^\beta\rfloor$ with fixed $\kappa>0$
and $0\le\beta<1/3$ satisfies this condition. An initial allowance
$k_T$ differs from an allowance $k_t$ that changes during a run.
Larger initial allowances, time-varying allowances, and their terminal and
retrospective criteria require further analysis. In particular, a fixed
multiple of $\sqrt T$ does not guarantee that every Bernoulli path stays
within the enlarged quota.

\begingroup
\section{Extensions and applications}\label{app:extensions-applications}
Appendix~\ref{app:ext-implications} states what the oracle analysis implies
for existing online conformal methods.
Appendix~\ref{app:ext-gap} allows a gap above the frontier at the target,
bounds it using held-out data, and studies randomization between two thresholds.
Appendix~\ref{app:ext-extensions} treats delayed labels, restarts, and ties.
Appendix~\ref{app:ext-residual} gives conditions for residual scores, and
Appendix~\ref{app:ext-workcell} describes an operational example.

\subsection{What the oracle analysis implies for online conformal methods}
\label{app:ext-implications}
Stationarity plays different roles in feasibility and efficiency.

\textbf{Feasibility needs no distributional assumption.}
Definition~\ref{def:valid} is pathwise. Counting, Barrier, and the quota
guard satisfy $M_t\le b_t$ on every stream, including streams with drift,
and the guard can be placed around any online method.

\textbf{The lower bounds are proved in the most favorable case.}
Theorem~\ref{thm:lower-bounds} allows i.i.d.\ data and a policy that knows
$D$ and $T$. The same distribution establishes a worst-case lower bound
for any broader data model containing it, with the same prediction-set
family and benchmark. Let $\mathcal P$ be
any class of data-generating processes that contains one stationary law $D$
satisfying the hypotheses of Theorem~\ref{thm:lower-bounds}. A method that
issues sets from $\cC$ and is feasible on every stream is, in particular, a
feasible policy on $D$, so its cost on $D$ relative to $tj(\alpha)$ obeys
(i)--(iii). Thus methods designed for unknown or drifting distributions
still incur this cost on that i.i.d.\ distribution. This is not an
efficiency guarantee on every drifting stream, or a comparison to a
different, time-varying oracle.
Corollary~\ref{cor:terminal-prefix-tradeoff} adds that a method with
terminal cost of order $\log T$ incurs expected cost at least of order $\sqrt{T/\log T}$
at some earlier time.

\textbf{The matching upper bounds use stationarity} to obtain uniform
insertion ranks and a fixed frontier. Appendix~\ref{app:ext-extensions}
treats delayed labels and restarts at fixed times.

\textbf{Methods allowing additional miscoverages.}
Adaptive conformal inference bounds how far cumulative miscoverage can
exceed $\alpha t$, rather than requiring $M_t\le b_t$ \citep{gibbs21}.
The next result makes this allowance explicit and applies
Proposition~\ref{prop:fixed-buffer-prefix}.

\begin{proposition}[Fixed-step ACI]\label{prop:ext-aci}
Fix a step $\gamma\in(0,1]$. Let $a_1=\alpha$ and
$a_{t+1}=a_t+\gamma(\alpha-E_t)$, where $E_t$ is the miscoverage indicator
of the issued set. The full-set map is issued whenever $a_t\le0$, and the
issued map is otherwise arbitrary. Then, on every stream and for every
$t\ge1$,
\[
 M_t-\alpha t=\frac{\alpha-a_{t+1}}{\gamma}\le\frac\alpha\gamma+1-\alpha,
 \qquad\text{hence}\qquad
 M_t\le b_t+k_\gamma,\quad
 k_\gamma:=\Bigl\lceil\frac\alpha\gamma+1-\alpha\Bigr\rceil .
\]
Suppose the issued maps belong to $\cC$ and the assumptions of
Theorem~\ref{thm:lower-bounds} hold. Then, with the coefficient $c_2$ of
that theorem,
\begin{equation}\label{eq:ext-aci-lower}
 \max_{1\le t\le T}\Reg_t\ge c_2T^{1/3}-1-2\lambda k_\gamma .
\end{equation}
If, in addition, the score cdf $F$ is continuous and the issued map is the
population threshold $F^{-1}(1-a_t)$ when $0<a_t<1$ and the empty-set map
when $a_t\ge1$, then
\begin{equation}\label{eq:ext-aci-linear}
 \Reg_T\ge\frac{c_-\gamma^2\min\{\alpha,1-\alpha\}^2}{4}
          \Bigl\lfloor\frac T2\Bigr\rfloor
          -\lambda\Bigl(\frac\alpha\gamma+1-\alpha\Bigr).
\end{equation}
\end{proposition}
\begin{proof}
Summing the update gives $a_{t+1}=\alpha+\gamma(\alpha t-M_t)$, which is the
stated identity. We show $a_t\ge-\gamma(1-\alpha)$ by induction. It holds at
$t=1$. If $a_t\le0$, the full set gives $E_t=0$ and
$a_{t+1}=a_t+\gamma\alpha\ge a_t$. If $a_t>0$, then
$a_{t+1}\ge a_t-\gamma(1-\alpha)>-\gamma(1-\alpha)$. This proves the bound
on $M_t-\alpha t$. Since $M_t$ is an integer and
$\lfloor x+y\rfloor\le\lfloor x\rfloor+\lceil y\rceil$, we get
$M_t\le b_t+k_\gamma$. The policy therefore belongs to
$\mathcal A_{k_\gamma}(T)$, and Proposition~\ref{prop:fixed-buffer-prefix}
gives \eqref{eq:ext-aci-lower}.

For \eqref{eq:ext-aci-linear}, the request
$q_t=\min\{1,\max\{0,a_t\}\}$ is a function of $E_1,\ldots,E_{t-1}$ and, by
continuity of $F$, equals $\Pp(E_t=1\mid E_1,\ldots,E_{t-1})$.
Inequality~\eqref{eq:reducedcost} does not use feasibility. With quadratic
lower growth it gives
$\Reg_T\ge\lambda\E G_T+c_-\E\sum_{t\le T}(q_t-\alpha)^2$.
The identity gives
$G_T=(a_{T+1}-\alpha)/\gamma\ge-(\alpha/\gamma+1-\alpha)$.
Write $\underline\alpha=\min\{\alpha,1-\alpha\}$. If $a_t$ and $a_{t+1}$
both lie in $(0,1)$, then
$|q_{t+1}-q_t|=\gamma|\alpha-E_t|\ge\gamma\underline\alpha$, so one of
$|q_t-\alpha|$ and $|q_{t+1}-\alpha|$ is at least
$\gamma\underline\alpha/2$. Otherwise one of the two requests is $0$ or $1$,
at distance at least $\underline\alpha\ge\gamma\underline\alpha/2$ from
$\alpha$. Each of the $\lfloor T/2\rfloor$ disjoint pairs $(t,t+1)$ thus
contributes at least $\gamma^2\underline\alpha^2/4$ to
$\sum_{t\le T}(q_t-\alpha)^2$ on every path.
\end{proof}

\begin{remark}[Reading the two bounds]\label{rem:ext-aci}
Inequality~\eqref{eq:ext-aci-lower} uses only the allowance $k_\gamma$, so it
applies to every calibrator satisfying $M_t\le b_t+k_\gamma$, regardless of
its update. This is the form of the coverage guarantee of \citet{gibbs21}
needed here. The bound is informative when $\gamma T^{1/3}$ is large,
and Corollary~\ref{cor:small-growing-buffer} shows that an allowance
$o(T^{1/3})$ cannot improve the $T^{1/3}$ order. This lower bound does not
determine ACI's actual cost. For fixed $\gamma$, \eqref{eq:ext-aci-linear}
gives an asymptotic linear lower bound for population thresholds, whose
requested miscoverage probabilities keep fluctuating. Its loose coefficient
can leave the bound negative at the horizons studied here.
\citet[Theorem~3.1]{zaffran} expand ACI's long-run average interval length
to first order as $\gamma\to0$ under their exchangeable-score oracle assumptions.
Bound~\eqref{eq:ext-aci-linear} holds at every finite horizon for each
$\gamma\in(0,1]$ under Proposition~\ref{prop:ext-aci}'s assumptions;
its $\gamma^2$ dependence is not claimed to be sharp. A step that
decreases with the horizon reduces this term but increases the allowed
excess miscoverage.
At $\alpha=0.1$ and the selected $\gamma=0.005$ of
Appendix~\ref{app:real-geometry}, $k_\gamma=21$. The largest observed
excess in Table~\ref{tab:guard-ablation} is $12$.
\end{remark}

\subsection{Approximate target efficiency}\label{app:ext-gap}
Assumption~\ref{ass:contact} requires the threshold family to attain the
frontier at $\alpha$. We quantify the excess size at the target and bound
its contribution to cumulative cost.

Fix a score with continuous cdf $F$. For $q\in(0,1)$ let $h(q)=V_D(C_\tau)$
at $\tau=F^{-1}(1-q)$, and let $h(0)=B$ and $h(1)=0$ for the full-set and
empty-set maps. An observed score $S_u$ satisfies $F^{-1}(F(S_u))=S_u$
almost surely, so every map selected by Algorithm~\ref{alg:rank} has
$V_D(C_t)=h(m_D(C_t))$ almost surely. Thresholds belong to $\cC$, so
$h\ge j$. Define the \emph{target gap}
\[
 \Delta_\alpha:=h(\alpha)-j(\alpha)\ge0 .
\]

\begin{proposition}[A target gap suffices under local smoothness]
\label{prop:ext-target-gap}
Suppose Assumption~\ref{ass:protocol} holds, $F$ is continuous, and $h'$
exists and is $L$-Lipschitz on
$[\alpha-\delta,\alpha+\delta]\subset(0,1)$. Then, for every $q\in[0,1]$,
\begin{equation}\label{eq:ext-gap-bound}
 0\le h(q)-j(\alpha)+\lambda(q-\alpha)
 \le2\Delta_\alpha+c_+'(q-\alpha)^2,
 \qquad c_+':=\max\{L,(B+\lambda)/\delta^2\}.
\end{equation}
Consequently, if each $q_t$ is $\rhist_{t-1}$-measurable and implemented by
Algorithm~\ref{alg:rank}, then for every integer $T\ge1$,
\begin{equation}\label{eq:ext-gap-regret}
 \Reg_T\le\lambda\E G_T
 +c_+'\Bigl\{\E\sum_{t=1}^T(q_t-\alpha)^2+\frac{\harm_T}{4}\Bigr\}
 +2T\Delta_\alpha .
\end{equation}
In particular, Counting has $\Reg_T\le2T\Delta_\alpha+O(\log T)$, and one
infinite run of Barrier has
$\max_{t\le T}\Reg_t\le2T\Delta_\alpha+O(T^{1/3})$ and
$\E\max_{t\le T}R_t\le2T\Delta_\alpha+O(\sqrt T)$. The constants depend
only on $\alpha,B,\lambda,L,\delta$.
\end{proposition}
\begin{proof}
Put $g(q)=h(q)-j(\alpha)+\lambda(q-\alpha)$. Since $h\ge j$ and $j$ lies
above its supporting line, $g\ge0$, and $g(\alpha)=\Delta_\alpha$. On the
neighborhood, $g'=h'+\lambda$ is $L$-Lipschitz. For $|u|\le\delta$, Taylor's
bound on each side gives
$g(\alpha\pm u)\le g(\alpha)\pm g'(\alpha)u+Lu^2/2$. Adding the two
inequalities cancels the linear terms:
\[
 g(\alpha+u)\le g(\alpha+u)+g(\alpha-u)\le2\Delta_\alpha+Lu^2 .
\]
No relation between $h'(\alpha)$ and $-\lambda$ is needed. For
$|q-\alpha|>\delta$, we have $g(q)\le B+\lambda$ because $h\le B$,
$j(\alpha)\ge0$, and $|q-\alpha|\le1$, while $(q-\alpha)^2>\delta^2$.
This proves \eqref{eq:ext-gap-bound}.

Repeat the proof of Theorem~\ref{thm:rank} with \eqref{eq:ext-gap-bound}
in place of Assumption~\ref{ass:contact}. Conditional on the full past history
and the current random seed, the expected size is $h(m_D(C_t))$, which is at most
$j(\alpha)-\lambda(m_D(C_t)-\alpha)+c_+'(m_D(C_t)-\alpha)^2+2\Delta_\alpha$.
Summing and applying \eqref{eq:rank-energy} gives \eqref{eq:ext-gap-regret}.
The laws of the miscoverage indicators of Counting and Barrier, and the
bounds on $\mathcal K_T$ in Lemmas~\ref{lem:counting}
and~\ref{lem:quadratic-upper}, do not involve Assumption~\ref{ass:contact}.
The terminal and expected-prefix consequences follow as in
Appendix~\ref{app:rank-proof}. The retrospective bound follows from the
display in Appendix~\ref{app:stopping}, whose right-hand side contains
$\Reg_T-\lambda\E G_T$.
\end{proof}

When $\Delta_\alpha=0$, the proposition recovers the implication stated in
Section~\ref{sec:learning}. More generally, a gap measured only at $\alpha$
controls the added cost under the stated local smoothness condition.
For a sequence of problems whose constants $(B,\lambda,L,1/\delta)$ stay
bounded, such as a score fitted on a growing training sample, the three
upper-bound rates are preserved when $\Delta_\alpha$ is $O(\log T/T)$,
$O(T^{-2/3})$, and $O(T^{-1/2})$, respectively.

\begin{remark}[Linear cost from a nonzero target gap]
\label{rem:ext-linear}
Suppose $h$ is continuous at $\alpha$. Then Counting and Barrier,
implemented by Algorithm~\ref{alg:rank}, satisfy
$\Reg_T/T\to\Delta_\alpha$. Indeed, \eqref{eq:rank-energy} and the entropy
bounds give $\E\sum_{t\le T}(m_D(C_t)-\alpha)^2=O(T^{1/3})$. Since $h$ is
bounded, $T^{-1}\sum_{t\le T}\E h(m_D(C_t))\to h(\alpha)$. A positive target
gap therefore costs $\Delta_\alpha$ per round for any controller whose
selected miscoverage levels concentrate at $\alpha$.
Randomizing between two thresholds can reduce this gap
(Proposition~\ref{prop:ext-mixture}).
\end{remark}

\paragraph{A sufficient condition for learned density scores.}
Take $\cC$ to be all measurable maps, so that $j$ is the unrestricted
frontier. Let $f(y\mid x)$ be a conditional density of $Y$ given $X$ with
respect to $\mu$, and write $\nu=P_X\otimes\mu$, so that
$V_D(C)=\nu\{(x,y):y\in C(x)\}$.

\begin{lemma}[Density-level scores]\label{lem:ext-density}
Let $k>0$ satisfy $\Pp\{f(Y\mid X)\ge k\}=1-\alpha$. Let $\widehat f$ be a
fixed function with $\sup|\widehat f-f|\le\eta$, and let
$\widehat k\in[k-\eta,k+\eta]$ satisfy
$\Pp\{\widehat f(Y\mid X)\ge\widehat k\}=1-\alpha$. If
$\nu\{|f-k|\le z\}\le Az$ for $0\le z\le2\eta$, then
\[
 0\le\nu\{\widehat f\ge\widehat k\}-j(\alpha)\le\frac{4A\eta^2}{k}.
\]
Such a $\widehat k$ exists when
$u\mapsto\Pp\{\widehat f(Y\mid X)\ge u\}$ is continuous on
$[k-\eta,k+\eta]$.
\end{lemma}
\begin{proof}
Let $S=\{f\ge k\}$ and $\widehat S=\{\widehat f\ge\widehat k\}$.
For measurable $\varphi:\mathcal X\times\mathcal Y\to[0,1]$ with
$\int\varphi f\,d\nu=1-\alpha$, the inequality
$\int(\one_S-\varphi)(f-k)\,d\nu\ge0$ gives $\int\varphi\,d\nu\ge\nu(S)$.
A mixture $\pi$ with miscoverage $\alpha$ has this form with
$\varphi(x,y)=\pi\{C:y\in C(x)\}$, so $j(\alpha)=\nu(S)$.
On $\widehat S\setminus S$ we have $f<k$ and
$f\ge\widehat f-\eta\ge\widehat k-\eta\ge k-2\eta$. On
$S\setminus\widehat S$ we have $f\ge k$ and
$f<\widehat k+\eta\le k+2\eta$. Both sets have coverage $1-\alpha$, hence
$\int_{\widehat S\setminus S}f\,d\nu=\int_{S\setminus\widehat S}f\,d\nu$.
Subtracting $k$ times the corresponding measures yields
\[
 k\{\nu(\widehat S)-\nu(S)\}
 =\int_{S\triangle\widehat S}|f-k|\,d\nu
 \le2\eta\,\nu\{|f-k|\le2\eta\}\le4A\eta^2 .
\]
For existence, $\{\widehat f\ge k+\eta\}\subset S\subset
\{\widehat f\ge k-\eta\}$, so the coverage of $\{\widehat f\ge u\}$ crosses
$1-\alpha$ on $[k-\eta,k+\eta]$.
\end{proof}

If the score $-\widehat f$ has a continuous cdf $F$, then
$F^{-1}(1-\alpha)\le-\widehat k$. Its quantile set is contained in
$\widehat S$ with the same coverage, so $\Delta_\alpha\le4A\eta^2/k$:
a uniform density estimation error $\eta$ gives a target gap of order $\eta^2$.

\paragraph{Controlled study.}
Let $X$ be uniform on two groups, with $f(y\mid x)=1+\theta_x(1-2y)$ on
$[0,1]$ and $(\theta_0,\theta_1)=(0.55,0.65)$. From $n$ training
observations per group, estimate $\theta_x$ by $3(1-2\overline Y_x)$,
clipped to $[0.2,0.9]$, and use $s(x,y)=-\widehat f(y\mid x)$ as the
nonconformity score. Using the known population distribution, we choose
a common cutoff with exactly $90\%$ coverage,
which isolates score approximation from threshold estimation.
Table~\ref{tab:ext-density} reports the excess expected size relative to the optimal sets
defined by the true density
over $256$ independent fits per size (seed $557892026$). The sample means
are consistent with a $1/n$ gap. In this model that order also follows in
expectation over training: clipping does not increase estimation error and
\[
 \E\eta^2\le\sum_{x=0}^1\E(\widehat\theta_x-\theta_x)^2
 \le\frac{(3-\theta_0^2)+(3-\theta_1^2)}{n}=\frac{5.275}{n}.
\]
Here $\eta=\max_x|\widehat\theta_x-\theta_x|$ and
$\Var(Y\mid X=x)=1/12-\theta_x^2/36$. The linear densities satisfy the
condition on $\nu\{|f-k|\le z\}$ in Lemma~\ref{lem:ext-density} for all $z\ge0$ with
$A=(\theta_0^{-1}+\theta_1^{-1})/2$, so Lemma~\ref{lem:ext-density} gives
$\E\Delta_\alpha\le4A\cdot5.275/(kn)=O(1/n)$.
If the local smoothness constants also remain bounded uniformly over fitted
scores, the expected gap term preserves the expected-prefix rate for
$n\gtrsim T^{2/3}$, the terminal rate for $n\gtrsim T/\log T$, and the
retrospective rate for $n\gtrsim\sqrt T$. The gap estimate alone does not
establish this uniform smoothness.
A single interval $[0,c]$ used for both groups, with $90\%$ coverage, needs no training
and has gap $6.15\times10^{-4}$. The learned score's mean gap is about nine
times larger at $n=64$. Among the tested sizes it is larger at $n=256$ and
smaller at $n=1{,}024$. We did not determine the training size at which
the learned score first improves on this interval.

\begin{table}[H]
\centering\small
\caption{Target gap of a learned density score in the controlled study
($\alpha=0.1$). Means and Monte Carlo standard errors over $256$ fits.}
\label{tab:ext-density}
\begin{tabular}{@{}rrrr@{}}
\toprule
Training size $n$ per group & Mean $\Delta_\alpha$ & Standard error & $n\Delta_\alpha$\\
\midrule
64 & $5.43\times10^{-3}$ & $0.47\times10^{-3}$ & 0.35\\
256 & $1.27\times10^{-3}$ & $0.10\times10^{-3}$ & 0.32\\
1{,}024 & $2.90\times10^{-4}$ & $0.25\times10^{-4}$ & 0.30\\
4{,}096 & $7.56\times10^{-5}$ & $0.73\times10^{-5}$ & 0.31\\
16{,}384 & $2.08\times10^{-5}$ & $0.19\times10^{-5}$ & 0.34\\
\bottomrule
\end{tabular}
\end{table}

\paragraph{A confidence bound from held-out data.}\label{app:ext-certificate}
Take $\cC$ to be the threshold family of a score fixed before observing an
independent held-out sample $(X_i,Y_i)_{i\le n}$ from $D$. Write
$j_{\rm sc}$ for its frontier and
$\Delta_\alpha^{\rm sc}=h(\alpha)-j_{\rm sc}(\alpha)$. Let
\[
 \widehat m_\tau=\frac1n\sum_{i\le n}\one\{s(X_i,Y_i)>\tau\},\qquad
 \widehat v_\tau=\frac1n\sum_{i\le n}\mu(C_\tau(X_i)),
\]
and let $\widehat j_{\rm sc}$ be the lower convex envelope of the points
$(\widehat m_\tau,\widehat v_\tau)$ together with $(0,B)$ and $(1,0)$.

\begin{proposition}[Confidence bound from held-out data]\label{prop:ext-certificate}
Let $\varrho\in(0,1)$, $\epsilon_m=\sqrt{\log(4/\varrho)/(2n)}$, and
$\epsilon_v=B\{\sqrt{\pi/(2n)}+\epsilon_m\}$. With probability at least
$1-\varrho$, simultaneously for all thresholds,
$|\widehat m_\tau-m_D(C_\tau)|\le\epsilon_m$ and
$|\widehat v_\tau-V_D(C_\tau)|\le\epsilon_v$. On this event, if $F$ is
continuous,
\begin{equation}\label{eq:ext-certificate}
 \Delta_\alpha^{\rm sc}\le
 \sup_{\tau:\,|\widehat m_\tau-\alpha|\le\epsilon_m}\widehat v_\tau
 -\widehat j_{\rm sc}(\min\{1,\alpha+\epsilon_m\})+2\epsilon_v .
\end{equation}
\end{proposition}
\begin{proof}
The Dvoretzky--Kiefer--Wolfowitz inequality \citep{massart1990tight} extends
to cdfs with atoms by the quantile transform. It bounds
the probability that $\sup_\tau|\widehat m_\tau-m_D(C_\tau)|>\epsilon_m$ by
$2\exp(-2n\epsilon_m^2)=\varrho/2$. The size bound is trivial if $B=0$. Otherwise, draw reference labels
$Y_i'$ from $\mu/B$, independent of everything else. Then
$V_D(C_\tau)/B$ is the cdf of $s(X_i,Y_i')$ at $\tau$, and
$\widehat v_\tau/B$ is the conditional expectation, given the holdout
features, of the empirical cdf of these $n$ reference scores. Jensen's
inequality and the preceding probability bound give
$\E\sup_\tau|\widehat v_\tau-V_D(C_\tau)|
\le B\int_0^\infty2e^{-2nu^2}\,du=B\sqrt{\pi/(2n)}$.
Changing one holdout feature moves every $\widehat v_\tau$ by at most
$B/n$, so the bounded-differences inequality \citep{mcdiarmid1989method}
bounds the probability that the supremum exceeds $\epsilon_v$ by
$\exp(-2n\epsilon_m^2)=\varrho/4$.

On the event, the threshold $\tau^\star=F^{-1}(1-\alpha)$ has
$m_D(C_{\tau^\star})=\alpha$ and $V_D(C_{\tau^\star})=h(\alpha)$, so
$|\widehat m_{\tau^\star}-\alpha|\le\epsilon_m$ and $h(\alpha)$ is at most
the supremum in \eqref{eq:ext-certificate} plus $\epsilon_v$.
Let $\pi$ be a mixture of thresholds and full-set or empty-set maps with
$m_D(\pi)=\alpha$. Its empirical miscoverage is at most
$\alpha+\epsilon_m$, its empirical size is at most $V_D(\pi)+\epsilon_v$,
and its empirical point lies on or above $\widehat j_{\rm sc}$. The envelope
is convex with minimum value $0$ at $q=1$, hence nonincreasing. Therefore
$\widehat j_{\rm sc}(\min\{1,\alpha+\epsilon_m\})\le V_D(\pi)+\epsilon_v$.
Taking the infimum over $\pi$ bounds $j_{\rm sc}(\alpha)$ from below.
\end{proof}

For the ten Housing model--split pairs, $n=5{,}160$, $B=1$, and
$\widehat v_\tau$ uses exact interval lengths. Taking $\varrho=0.05/10$
gives $\epsilon_m=0.0255$ and $\epsilon_v=0.0429$. Under the assumptions of
Proposition~\ref{prop:ext-certificate}, the resulting simultaneous $95\%$ upper bounds
\eqref{eq:ext-certificate} range from $0.158$ to $0.175$, while the
empirical gaps in Table~\ref{tab:contact-gaps} are at most $0.00293$.
These bounds are too large to establish a gap of order $T^{-2/3}$, as needed
to preserve the expected-prefix rate. Even the additive $2\epsilon_v$ term
has order $n^{-1/2}$, so it reaches $T^{-2/3}$ only when $n$ is at least
of order $T^{4/3}$, up to factors depending on the confidence level. This requirement applies to
this confidence bound. Other ways of bounding the gap need not have the
same requirement. Each split's bound holds conditional on its own fitted
score. Averaging over the fitted scores and taking a union bound gives the
simultaneous statement without requiring independent splits. These bounds
concern the frontier of the fixed score's threshold family. Bounding the
gap relative to all measurable prediction-set rules would also require a
lower bound for their frontier.

\paragraph{Randomizing between two thresholds.}
Fix two miscoverage levels $0\le q_-<\alpha<q_+\le1$ before the online run,
for example using independent held-out data. Levels $0$ and $1$ use the
full-set and empty-set maps. Implement a request $q_t$ as follows. If
$q_t\in[q_-,q_+]$, draw an independent $\varpi'_t\sim\mathrm{Unif}[0,1)$ and
pass $q_+$ to Algorithm~\ref{alg:rank} when
$\varpi'_t<(q_t-q_-)/(q_+-q_-)$, and $q_-$ otherwise. If
$q_t\notin[q_-,q_+]$, pass $q_t$ itself. Define
\[
 \ell(q)=\frac{(q_+-q)h(q_-)+(q-q_-)h(q_+)}{q_+-q_-},\qquad
 \Delta_{\rm mix}:=\ell(\alpha)-j(\alpha)\ge0.
\]
For $q\in[q_-,q_+]$, $\ell(q)$ is the expected size obtained by mixing population thresholds
at the two fixed levels to achieve miscoverage $q$, and $\Delta_{\rm mix}$
is its gap above the frontier at $\alpha$. The gap is nonnegative because
this randomization is allowed in $\Delta(\cC)$.

\begin{proposition}[Randomization between two thresholds]\label{prop:ext-mixture}
Suppose Assumption~\ref{ass:protocol} holds and $F$ is continuous. For each
fixed level $a\in\{q_-,q_+\}\cap(0,1)$, suppose $h'$ exists and is $L$-Lipschitz
on $[a-\delta,a+\delta]\subset(0,1)$. Let $\rho=\min\{\alpha-q_-,q_+-\alpha\}$,
$c_{\rm mix}=(B+\lambda)/\rho^2$, and let $c_{\rm anc}$ be the largest value
of $\max\{L/2,(B+|h'(a)|)/\delta^2\}$ over these levels, or zero if
there are none. If the requests $q_t$ are functions of past miscoverage
indicators, then the implementation above satisfies
$\Pp(E_t=1\mid E_1,\ldots,E_{t-1})=q_t$ and, for every integer $T\ge1$,
\begin{equation}\label{eq:ext-mixture-regret}
 \Reg_T\le\lambda\E G_T+c_{\rm mix}\E\sum_{t=1}^T(q_t-\alpha)^2
 +c_{\rm anc}\frac{\harm_T}{4}+2T\Delta_{\rm mix}.
\end{equation}
Hence Counting has $\Reg_T\le2T\Delta_{\rm mix}+O(\log T)$, and one infinite
run of Barrier has $\max_{t\le T}\Reg_t\le2T\Delta_{\rm mix}+O(T^{1/3})$ and
$\E\max_{t\le T}R_t\le2T\Delta_{\rm mix}+O(\sqrt T)$.
\end{proposition}
\begin{proof}
Include the random variables $\varpi'_u$ in the rank history. The fresh insertion rank is
independent of this history, so Lemma~\ref{lem:rank-moments} applies to the
level sent to Algorithm~\ref{alg:rank}: given the history and $\varpi'_t$, the
miscoverage probability equals that level and $\Var(m_D(C_t))\le1/(4t)$.
Both statements are conditional on this history and $\varpi'_t$. Averaging over
$\varpi'_t$ gives conditional miscoverage probability $q_t$, and the tower
property gives the stated probability conditional on past miscoverage
indicators. The laws of the miscoverage
indicators of Counting and Barrier, and their entropy bounds, are therefore
unchanged.

Let $a$ be one of the fixed levels in $(0,1)$. For $|m-a|\le\delta$,
$h(m)\le h(a)+h'(a)(m-a)+L(m-a)^2/2$. For $|m-a|>\delta$, the difference
$h(m)-h(a)-h'(a)(m-a)$ is at most $B+|h'(a)|$. Since the selected
miscoverage has mean $a$ and variance at most $1/(4t)$, the expected size
given that $a$ is sent to Algorithm~\ref{alg:rank} is at most
$h(a)+c_{\rm anc}/(4t)$. Levels $0$ and $1$ use the full and empty sets
exactly. Hence, for $q_t\in[q_-,q_+]$, the
conditional expected size is at most $\ell(q_t)+c_{\rm anc}/(4t)$.

Put $g(q)=\ell(q)-j(\alpha)+\lambda(q-\alpha)$ on $[q_-,q_+]$. It is affine,
and nonnegative because $\ell(q)\ge j(q)$. For $|u|\le\rho$,
$g(\alpha+u)=2g(\alpha)-g(\alpha-u)\le2\Delta_{\rm mix}$. For other points
of $[q_-,q_+]$, $g\le B+\lambda\le c_{\rm mix}(q-\alpha)^2$. If
$q_t\notin[q_-,q_+]$, the conditional expected size minus
$j(\alpha)-\lambda(q_t-\alpha)$ is at most
$B+\lambda\le c_{\rm mix}(q_t-\alpha)^2$. In all cases,
\[
 \E[\ell_{C_t}(X_t)\mid E_1,\ldots,E_{t-1}]
 \le j(\alpha)-\lambda(q_t-\alpha)+c_{\rm mix}(q_t-\alpha)^2
     +\frac{c_{\rm anc}}{4t}+2\Delta_{\rm mix}.
\]
Sum over $t$ and use $\E\sum_{t\le T}q_t=\E M_T$ to obtain
\eqref{eq:ext-mixture-regret}. The controller bounds follow as in Proposition~\ref{prop:ext-target-gap}.
For the retrospective bound, include $\varpi'_t$ in the round-$t$ random seed and use the martingale argument in Appendix~\ref{app:stopping}.
\end{proof}

\begin{remark}[When a threshold or a mixture attains the frontier]\label{rem:ext-dichotomy}
Let $\cC$ be the threshold family of a fixed score. If
$h(\alpha)=j_{\rm sc}(\alpha)$, a single threshold attains this frontier at
the target, and Theorem~\ref{thm:optimal-prices} applies under its upper
smoothness and lower growth conditions. Otherwise, suppose
$j_{\rm sc}(\alpha)$ is attained by mixing the two population maps implemented
at levels $q_-<\alpha<q_+$ in Proposition~\ref{prop:ext-mixture}, with its
local smoothness conditions. At levels $0$ and $1$ these must be the literal
full and empty sets, not arbitrary zero- or unit-miscoverage thresholds.
Then $\Delta_{\rm mix}=0$, so Proposition~\ref{prop:ext-mixture} gives the
three upper rates without a term proportional to $T$. These rates need not be tight. In this case
$j_{\rm sc}$ is linear on $[q_-,q_+]$, and $\psi$ vanishes there.
The hypotheses of Theorem~\ref{thm:lower-bounds}(i)--(ii) fail on this
interval, while (iii) still holds when $\psi(1)>0$. Levels estimated from
held-out data generally give $\Delta_{\rm mix}>0$. Linearity of the frontier
alone does not ensure that the selected thresholds attain it or satisfy
the local smoothness conditions needed for an $O(\log T)$ threshold-estimation
cost. The bound therefore allows a nonzero mixture gap instead of requiring
one threshold to attain the frontier exactly at $\alpha$.
\end{remark}

\subsection{Delayed labels, restarts, and tied scores}
\label{app:ext-extensions}
Controllers can be combined across disjoint subsequences of rounds. If
each subsequence satisfies its own quota, then
$M_t\le\sum_i\lfloor\alpha n_i(t)\rfloor\le\lfloor\alpha t\rfloor$, where
$n_i(t)$ counts the rounds of subsequence $i$ among the first $t$. The two
propositions below combine this fact with the single-run bounds.

\begin{proposition}[Bounded feedback delay]\label{prop:ext-delay}
Fix an integer $d\ge0$ and suppose $Y_t$ is revealed before the prediction
of round $t+d+1$. Run $d+1$ copies of a feasible controller with
Algorithm~\ref{alg:rank}. Copy $i$ predicts on rounds
$t\equiv i\pmod{d+1}$ and uses only the observations of its own rounds.
Then $M_t\le b_t$ on every stream. Under the assumptions of
Theorem~\ref{thm:rank} and for $T\ge2(d+1)$, running each copy of Counting
for its assigned number of rounds gives $\Reg_T=O((d+1)\log T)$, and copies of Barrier
give $\max_{t\le T}\Reg_t=O((d+1)^{2/3}T^{1/3})$ and
$\E\max_{t\le T}R_t=O(\sqrt{(d+1)T})$. The lower bounds of
Theorem~\ref{thm:lower-bounds} remain valid under their assumptions, so the
three optimal cost rates are unchanged for fixed $d$.
\end{proposition}
\begin{proof}
Each copy predicts every $d+1$ rounds, so the label of its previous round
has been revealed. Each copy satisfies its local quota on every stream, and
combining the quotas gives $M_t\le b_t$. The rounds of one copy form an i.i.d.\ sample
from $D$, so the single-run bounds apply to its number of predictions, at most
$\lceil T/(d+1)\rceil$. Excess sizes add across copies:
$R_t=\sum_iR^{(i)}_{n_i(t)}$. Summing the logarithmic bounds gives the first
claim. Concavity gives
$\sum_in_i(t)^{1/3}\le(d+1)^{2/3}t^{1/3}$ and
$\sum_in_i(t)^{1/2}\le\sqrt{(d+1)t}$, and the additive constants are
absorbed because $d+1\le T$. The retrospective-cost bound also uses
$\max_tR_t\le\sum_i\max_nR^{(i)}_n$. A policy with delayed labels is a
policy of Assumption~\ref{ass:protocol} that ignores recent outcomes, so
Theorem~\ref{thm:lower-bounds} applies to it.
\end{proof}

\begin{proposition}[Restarts at fixed times]\label{prop:ext-restart}
Let $0=\tau_0<\tau_1<\cdots<\tau_{N}=T$ be fixed in advance, and call rounds
$\tau_{k-1}+1,\ldots,\tau_k$ period $k$. At each boundary, restart Barrier
with Algorithm~\ref{alg:rank}, an empty score history, zero slack, and a
score fixed before the period. Then $M_t\le b_t$ on every stream. Suppose
that, conditional on all information available at the start of period $k$,
including the data used to fit its score, the pairs in that period are i.i.d.\ from a distribution
$D_k$ satisfying the assumptions of Theorem~\ref{thm:rank}, with frontier
$j_k$ and with $\lambda$ and $c_+$ bounded by $\bar\lambda$ and $\bar c_+$ for
every $k$. Then there is $c<\infty$, depending only on
$\alpha,B,\bar\lambda,\bar c_+$, such that for every $t\le T$,
\[
 \E\!\left[W_t-\sum_{k\le N}n_k(t)\,j_k(\alpha)\right]
 \le c\,N^{2/3}t^{1/3},
\]
where $n_k(t)$ is the number of rounds of period $k$ among the first $t$.
The expectation includes the fitted scores and the frontiers $j_k$
determined by past observations.
\end{proposition}
\begin{proof}
Feasibility follows by combining the quotas. Conditional on the information
available at the restart, \eqref{eq:sine-regret} with the entropy bound
\eqref{eq:quadratic-entropy} gives an excess of at most $c_0+c_1n^{1/3}$
after $n$ rounds of a period, with $c_0,c_1$ depending only on
$\alpha,B,\bar\lambda,\bar c_+$. Let $N_t\le\min\{N,t\}$ be the number of
periods started by time $t$. Concavity gives
$\sum_kn_k(t)^{1/3}\le N_t^{2/3}t^{1/3}$, and
$N_t\le N_t^{2/3}t^{1/3}$. Apply the tower property to each period's
conditional bound, also averaging over its fitted score and population benchmark, and take
$c=c_0+c_1$.
\end{proof}

The benchmark in Proposition~\ref{prop:ext-restart} is the sum of the
periods' own population benchmarks, and the score may be refitted between
periods. Unknown change points and gradual drift are not covered.

\begin{remark}[Tied scores]\label{rem:ext-ties}
Counting, Barrier, and the guard satisfy the quota on every stream even
when scores tie.
Draw independent $\zeta_t\sim\mathrm{Unif}[0,1)$ and regard
$\widetilde X_t=(X_t,\zeta_t)$ as the fresh covariate. For a stored pair
$(\tau,u)$ define the map
\[
 \widetilde C_{\tau,u}(x,z)=\{y:s(x,y)<\tau\}
       \cup\{y:s(x,y)=\tau,\ z\le u\}.
\]
Select this map before observing $(X_t,\zeta_t)$, so all candidate labels
use the same random variable to break ties. Controller requests use only
past insertion ranks of $(S_t,\zeta_t)$ and independent rounding randomness.
The actual scores and tie-breaking values are not included in the history
used to choose these requests. Under the joint distribution $\widetilde D$
of $(X,\zeta,Y)$, the miscoverage probability and expected set size average
over the new tie-breaking variable. For i.i.d.\ observations, the randomized
probability integral transform
\[
 U_t=F(S_t^-)+\zeta_t\{F(S_t)-F(S_t^-)\}
\]
gives i.i.d.\ uniform variables with the same order as the pairs
$(S_t,\zeta_t)$: first compare scores, and use $\zeta_t$ when scores tie.
Thus Lemma~\ref{lem:rank-moments} holds for
$m_{\widetilde D}(\widetilde C_t)$. The size bounds require
Assumption~\ref{ass:contact} for this randomized prediction-set family
and its frontier. The variance bound concerns miscoverage averaged over
the new tie-breaking variable. If miscoverage is instead computed
conditional on that variable, its variance can be of constant order:
when all scores tie, the issued set can be either full or empty.
\end{remark}

\subsection{Residual and scaled-residual scores}\label{app:ext-residual}
\begin{proposition}[Residual scores]\label{prop:ext-residual}
Let $\mathcal Y=[0,B]$ with Lebesgue measure and
$Y=m(X)+\sigma(X)\varepsilon$, where $\varepsilon$ is independent of $X$
with a density $\varphi$ that is symmetric about zero and supported on
$[-\bar r,\bar r]$, $\sigma$ is positive and bounded, and
$[m(X)-\sigma(X)\bar r,\,m(X)+\sigma(X)\bar r]\subset[0,B]$ almost surely.
Let $f_{|\varepsilon|}=2\varphi$ on $[0,\bar r]$ be the density of
$|\varepsilon|$, $F_{|\varepsilon|}$ its cdf, and
$r_\alpha=F_{|\varepsilon|}^{-1}(1-\alpha)$.

\textup{(a)} \emph{Location model.} Let $\sigma\equiv1$, let $\varphi$ be
strictly decreasing on $[0,\bar r]$, and use $s(x,y)=|y-m(x)|$. Then the
threshold sets consist of outcomes whose conditional density is at least a
common cutoff, and $h(q)=2F_{|\varepsilon|}^{-1}(1-q)$ equals the frontier
over all measurable prediction-set rules for $q\in(0,1)$. If $f_{|\varepsilon|}$ is positive and
Lipschitz near $r_\alpha$, Assumption~\ref{ass:contact} holds with
$\lambda=2/f_{|\varepsilon|}(r_\alpha)$.

\textup{(b)} \emph{Location-scale model.} Use
$s(x,y)=|y-m(x)|/\sigma(x)$. Then
$h(q)=2\E[\sigma(X)]F_{|\varepsilon|}^{-1}(1-q)$ for $q\in(0,1)$. If
$f_{|\varepsilon|}$ is nonincreasing on $[0,\bar r]$, then $h$ is convex on
$[0,1]$, the frontier within this threshold family is $j_{\rm sc}=h$ on $(0,1]$, and
$\Delta_\alpha^{\rm sc}=0$. If $f_{|\varepsilon|}$ is also positive and
Lipschitz near $r_\alpha$, Assumption~\ref{ass:contact} holds relative to
$j_{\rm sc}$.
\end{proposition}
\begin{proof}
\textup{(a)} The conditional density of $Y$ at $y$ is $\varphi(y-m(x))$.
For $0<\tau<\bar r$, strict monotonicity gives
$\{y:|y-m(x)|\le\tau\}=\{y:\varphi(y-m(x))\ge\varphi(\tau)\}$, a level set
with cutoff $\varphi(\tau)>0$ for every $x$. The set has length $2\tau$ and
miscoverage $1-F_{|\varepsilon|}(\tau)$, which gives $h$. The first step of
the proof of Lemma~\ref{lem:ext-density} shows that such a level set
minimizes expected size among all randomized maps with the same
miscoverage, so $h=j$ on $(0,1)$. Finally
$h'(q)=-2/f_{|\varepsilon|}(F_{|\varepsilon|}^{-1}(1-q))$ is Lipschitz near
$\alpha$ under the stated condition, and Section~\ref{sec:learning} gives
Assumption~\ref{ass:contact}.

\textup{(b)} For $0\le\tau\le\bar r$, the threshold set is an interval of
length $2\sigma(x)\tau$ with miscoverage $1-F_{|\varepsilon|}(\tau)$,
which gives $h$. The nonincreasing density makes $F_{|\varepsilon|}$ concave,
so its quantile function is convex on $(0,1)$, as is $h$.
The endpoints preserve convexity: $h(1)=0$ is the limit of $h$ at $1$, and
$h(0)=B$ is at least its limit at $0$, which is at most
$2\E[\sigma(X)]\bar r\le B$. Every other zero-miscoverage threshold also
has size at least this limit. Thus mixing thresholds cannot lower the
convex curve at any level in $(0,1]$. The last claim follows as in
part~(a), with $\cC$ the threshold family.
\end{proof}

In part~(b), the endpoint density $\varphi(\tau)/\sigma(x)$ generally varies
with $x$, so $j_{\rm sc}$ need not equal the frontier over all measurable
maps. A nonmonotone density of $|\varepsilon|$ can make $h$ nonconvex
(Remark~\ref{rem:ext-dichotomy}). Neither model is verified for Housing,
so Section~\ref{sec:experiments} reports empirical gaps.

\subsection{Why the quota may be needed after every prediction}
\label{app:ext-workcell}
Inspection requires immediate correction and feedback before the next prediction.
The correction budget starts at zero and gains one unit before parts
$10,20,30,\ldots$, with carryover but no borrowing.
Propositions~\ref{prop:fixed-buffer-prefix} and~\ref{prop:ext-delay} cover
initial reserves and bounded delays. These assumptions need validation in deployment.
\endgroup

\end{document}